\newif\iftlp\tlptrue

\documentclass{new_tlp}

\newif\ifdraft\draftfalse
\newif\ifinlineref\inlinereftrue
\newif\iffinal\finaltrue
\newif\ifextended\extendedfalse
\newif\ifdotikz\dotikzfalse
\newif\ifshowotherappendix\showotherappendixtrue
\newif\ifmakeallproofsinline\makeallproofsinlinefalse
\newif\ifrevisionmarkers\revisionmarkersfalse
\newif\ifrevisionmarkerstwo\revisionmarkerstwofalse
\newif\ifcorrversion\corrversiontrue

\draftfalse

\dotikztrue
\ifdraft
\fi

\usepackage{times}
\usepackage[scaled]{helvet}
\usepackage{courier}

\usepackage{enumerate}
\usepackage{amsmath}
\usepackage{amssymb}
\usepackage{url}
\usepackage{xspace}
\usepackage{marvosym}
\usepackage{paralist}
\usepackage{graphicx}
\graphicspath{{figures/}{figures/experiments/}{experiments/dmcs_instances/}{experiments/revsel2/}{experiments/revsel1/analyze_1_8/}}
\usepackage{subfig}
\usepackage{multicol}
\usepackage{listings}
\usepackage{booktabs}
\usepackage{xcolor}
\usepackage[normalem]{ulem}
\usepackage{ifpdf}
\ifpdf
\usepackage{microtype}
\fi

\iftlp

\else
	\usepackage{amsthm}
\fi

\ifdraft
\usepackage{pdfsync}
\usepackage{srcltx}
\else
\fi

\usepackage[lined,algoruled]{algorithm2e}
\usepackage[outsourcebydefault]{mytheorem}

\newMyTheoremType{theorem}{Theorem}
\newMyTheoremType{claim}{Claim}
\newMyTheoremType{corollary}{Corollary}
\newMyTheoremType{proposition}{Proposition}
\newMyTheoremType{lemma}{Lemma}

\ifdraft
\newcommand{\comment}[1]{{\small\bf\color{blue} *** #1 ***}}
\else
\newcommand{\comment}[1]{}
\fi

\ifdotikz
\usepackage{tikz}
\pgfrealjobname{inlining}
\usetikzlibrary{shapes,arrows,backgrounds,chains,%
matrix,patterns,arrows,decorations.pathmorphing,decorations.pathreplacing,%
positioning,fit,calc,decorations.text,shadows%
}
\else
\long\def\beginpgfgraphicnamed#1#2\endpgfgraphicnamed{\includegraphics{#1}}
\fi

\newcommand{\nop}[1]{#1}
\renewcommand{\nop}[1]{} 

\ifmakeallproofsinline
\newcommand{\myinlineproof}[1]{#1}
\newcommand{\mylocatedproof}[1]{}
\else
\newcommand{\myinlineproof}[1]{}
\newcommand{\mylocatedproof}[1]{#1}
\fi

\nop{
\newcommand\myproofFOO{
}
\myinlineproof{%
  \begin{proof}[Proof of BAR~\ref{thm:BAZ}]
    \myproofFOO
  \end{proof}
}
  \mylocatedproof{%
    \begin{proof}[Proof of BAR~\ref{thm:BAZ}]
      \myproofFOO
    \end{proof}
  }
}

\ifrevisionmarkers
\definecolor{darkgreen}{rgb}{0,0.5,0}
\newcommand{\reva}[1]{{\color{darkgreen} #1}}
\newcommand{\revam}[1]{{\color{red} \sout{#1}}}
\newcommand{\revanop}[1]{}
\newcommand{\rev}[2]{{\color{blue}#2}}
\newcommand{\revmv}[1]{{\color{gray}#1}}
\else
\newcommand{\reva}[1]{{#1}}
\newcommand{\revam}[1]{}
\newcommand{\revanop}[1]{}
\newcommand{\rev}[2]{{#2}}
\newcommand{\revmv}[1]{{#1}}
\fi
\ifrevisionmarkerstwo
\definecolor{darkgreen}{rgb}{0,0.5,0}

\newcommand{\revtwoam}[1]{{\color{red} \sout{#1}}}
\newcommand{\revtwoanop}[1]{}
\newcommand{\revtwo}[2]{{\color{blue}#2}}

\else

\newcommand{\revtwoam}[1]{}
\newcommand{\revtwoanop}[1]{}
\newcommand{\revtwo}[2]{{#2}}

\fi

\newtheorem{definition}{Definition}
\newtheorem{example}{Example}

\newcommand{\leanparagraph}[1]{\smallskip\noindent\textbf{#1}}

\newcounter{myenumeratecounter}

\renewcommand{\vec}[1]{\ensuremath{\mb{#1}}}

\newcommand{\mi}[1]{\ensuremath{\mathit{#1}}}
\newcommand{\mb}[1]{\ensuremath{\mathbf{#1}}}
\def\lif{\ensuremath{\leftarrow}}

\def\naf{\ensuremath{\mathop{not}}}

\def\cI{\ensuremath{{\mathcal{I}}}}
\def\cA{\ensuremath{{\mathcal{A}}}}
\def\cE{\ensuremath{{\mathcal{E}}}}

\def\scI{\ensuremath{{\textsc{i}}}}
\def\scO{\ensuremath{{\textsc{o}}}}

\newcommand\bi{\begin{itemize}}
\newcommand\ei{\end{itemize}}
\newcommand\quo[1]{`#1'}

\newcommand{\eqs}{\,{=}\,}
\newcommand{\neqs}{\,{\neq}\,}

\newcommand{\cups}{\,{\cup}\,}

\newcommand{\lors}{\,{\vee}\,}

\newcommand\hex{{\sc hex}\xspace}
\newcommand\dlv{{\small\sffamily dlv}\xspace}
\newcommand\dlvhex{{\small\sffamily dlvhex}\xspace}

\newcommand{\AS}{\mathcal{A\!S}}
\newcommand{\dq}{\ifmmode\text{"}\else"\fi}
\newcommand{\lifs}{\,{\lif}\,}

\newcommand\heurold{\textsl{H1}\xspace}
\newcommand\heurnew{\textsl{H2}\xspace}

\newcommand{\dllite}{\ensuremath{\mathit{DL\text{-}Lite}}}

\newcommand{\mycuri}[0]{\ensuremath{\mi{cur\scI}}}
\newcommand{\mycuro}[0]{\ensuremath{\mi{cur\scO}}}
\newcommand{\myrefcounto}[0]{\ensuremath{\mi{refs\scO}}}

\newcommand\clasp{{\small\sffamily clasp}\xspace}

\newcommand\gringo{{\small\sffamily gringo}\xspace}
\newcommand\clingo{{\small\sffamily clingo}\xspace}

\newcommand{\amp}[1]{\ensuremath{\text{\textsl{{\&}}}\!\mathit{#1}}}
\newcommand{\ext}[3]{\ensuremath{\amp{#1}[#2](#3)}}
\newcommand{\extfun}[1]{\ensuremath{f_{\text{\sl\&}#1}}}
\newcommand{\extFun}[1]{\ensuremath{F_{\text{\sl\&}#1}}}
\newcommand{\extsem}[4]{\ensuremath{f_{\text{\sl\&}#1}(#2,#3,#4)}}

\newcommand{\rat}[1]{\ensuremath{d_{\text{\it\&}#1}}}

\newcommand{\GroundLiberallyDomainExpansionSafeProgram}{\ensuremath{\textsc{GroundHEX}}}
\newcommand{\EvaluateGroundHEX}{\ensuremath{\textsc{EvaluateGroundHEX}}}
\newcommand{\EvaluateLDESafe}{\ensuremath{\textsc{EvaluateLDESafe}}}

\def\dependsext{\rightarrow^e}

\def\dependsmon{\rightarrow_m}
\def\dependsnmon{\rightarrow_n}

\newcommand{\T}{\mathbf{T}}
\newcommand{\F}{\mathbf{F}}

\def\join{{\:\bowtie\:}}

\def\ufinal{u_{\mi{final}}}

\def\myiint{i-inter\-pre\-ta\-tion\xspace}

\def\myoint{o-inter\-pre\-ta\-tion\xspace}

\def\myimodels{\mi{i\text{-}ints}}
\def\myomodels{\mi{o\text{-}ints}}

\def\myunit{\mi{unit}}
\def\mytype{\mi{type}}
\def\myint{\mi{int}}
\newcommand{\myinputsE}[1]{\mi{preds}_{#1}}
\newcommand{\myinputs}{\myinputsE{\cE}}

\def\myevaluatePregroundable{\ensuremath{\textsc{evaluate}\-\textsc{Pre}\-\textsc{Groundable}}\xspace}

\def\myBuildAnswerSets{\ensuremath{\textsc{Build}\-\textsc{Answer}\-\textsc{Sets}}\xspace}
\def\myGetNextUnitModel{\ensuremath{\textsc{Get}\-\textsc{Next}\-\textsc{Unit}\-\textsc{Model}}\xspace}
\def\myGetNextIModel{\ensuremath{\textsc{Get}\-\textsc{Next}\-\textsc{Input}\-\textsc{Model}}\xspace}
\def\myGetNextOModel{\ensuremath{\textsc{Get}\-\textsc{Next}\-\textsc{Output}\-\textsc{Model}}\xspace}
\def\myEnsureModelIncrement{\ensuremath{\textsc{Ensure}\-\textsc{Model}\-\textsc{Increment}}\xspace}
\def\myOnDemandAS{\ensuremath{\textsc{Answer}\-\textsc{Sets}\-\textsc{On}\-\textsc{Demand}}\xspace}
\def\myNextAnswerSet{\ensuremath{\mathit{NextAnswerSet}}}
\def\undef{\ensuremath{\textsc{undef}}}

\def\myCAUtext{FAI\xspace}
\def\myCAUstext{FAIs\xspace}
\def\mycau{\mi{fai}}

\newcommand\facts[0]{\ensuremath{\mi{facts}}}

\def\papertitle{Inlining External Sources in Answer Set Programs}
\def\shortpapertitle{Inlining External Sources in Answer Set Programs}

\iftlp
	\title[\shortpapertitle]{\papertitle%
	\iffinal
	\thanks{
		This article is an extension of preliminary work presented at AAAI~2017
		\protect\cite{r2017a-aaai,r2017b-aaai}.
		This work has been supported by the Austrian Science Fund (FWF) Grant
		P27730.}
	\fi
	}

	\author[Redl]{%
		Christoph Redl \\
		\rev{Institut f\"ur Informationssysteme, Technische Universit\"at Wien}{Institute of Logic and Computation, Vienna University of Technology} \\
    Favoritenstra\ss e 9-11, A-1040 Vienna, Austria \\
    \email{redl@kr.tuwien.ac.at}
}

\else
	\title{\papertitle%
	}

	\author{Christoph Redl}

\fi

\input{figuremacros.tex}

\begin{document}
	\iftlp
		\maketitle
	\fi


	\medskip
	\begin{abstract}
		\hex{}-programs are an extension of answer set \rev{programming}{programs} (ASP) \rev{towards}{with}
		external sources.
		To this end, \emph{external atoms} provide
		a bidirectional interface between the program and an external source.
		The traditional evaluation algorithm for \hex-programs
		is based on guessing truth values of external atoms and verifying them by explicit calls of the external source.
		The approach was optimized by techniques that reduce the number of necessary verification calls
		or speed them up, but the remaining external calls are still expensive.
		In this paper we present an alternative evaluation approach based on \emph{inlining} of external atoms,
		motivated by existing but less general approaches for specialized formalisms such as DL-programs.
		External atoms are then compiled away such that no verification calls are necessary.
		The approach is implemented in the \dlvhex{} reasoner. Experiments show a significant performance gain.
		Besides performance improvements,
		we further exploit inlining for extending previous \rev{notions of}{(semantic) characterizations of} \revtwo{programs}{program} equivalence from ASP to \hex-programs,
		including \reva{those of} \emph{strong equivalence}, \emph{uniform equivalence} and \emph{$\langle \mathcal{H}, \mathcal{B} \rangle$-equivalence}.
		\revtwo{Based on \rev{this extended equivalence notion}{these equivalence criteria} we finally characterize also inconsistency of programs wrt.~extensions.}{
		Finally, based on \rev{this extended equivalence notion}{these equivalence criteria}, we characterize also inconsistency of programs wrt.~extensions.}
		Since well-known ASP extensions (such as constraint ASP) \rev{amount to}{are} special cases of \hex{},
		the results are interesting beyond the particular formalism.
		\ifcorrversion
		
		Under consideration in Theory and Practice of Logic Programming (TPLP).
		\fi
	\end{abstract}  

	\iftlp
	\fi

	\abovedisplayshortskip=2pt
	\belowdisplayshortskip=2pt
	\abovedisplayskip=4pt
	\belowdisplayskip=4pt
	\maketitle

	\section{Introduction}
	\label{sec:introduction}

		\hex{}-programs \revtwo{are an extension of}{extend} answer set \rev{programming}{progams} (ASP) \revtwo{\cite{gelf-lifs-91}}{as introduced by \citeN{gelf-lifs-91}} \rev{towards}{with}
		external sources.
		\rev{As}{Like} ASP, \hex-programs are based on nonmonotonic programs and have a multi-model semantics.
		External sources are used to represent knowledge and computation sources such as,
		for instance, description logic ontologies and Web resources.
		To this end, so-called \emph{external atoms} are used to send information from the logic program
		to an external source, which returns values to the program.
		Cyclic rules that involve external atoms are allowed, such that
		recursive data exchange between the program and external sources is possible.
		\revam{Moreover, \emph{value invention} allows for
		returning values which are not contained in the input program, i.e., which expand the domain.}
		A concrete example is the external atom $\rev{\ext{\mathit{edge}}{G}{X,Y}}{\ext{\mathit{edge}}{g}{x,y}}$
		which \rev{returns for a filename $G$, pointing to a file which stores a graph, the contained edges $(x,y)$}
		{evaluates to true for all edges $(x,y)$ contained in a graph that is stored in a file identified by a filename $g$}.

		The traditional evaluation procedure for \hex-programs is based on rewriting external atoms to ordinary atoms
		and guessing their truth values. This yields answer set candidates \rev{, which}{that} are subsequently checked
		to ensure that the guessed values coincide with the actual semantics of the external atoms.
		Furthermore, an additional minimality check is necessary to exclude self-justified atoms, which involves even more external calls.
		Although this approach has been refined by integrating advanced techniques for learning~\cite{efkr2012-tplp}
		and efficient minimality checking~\cite{efkrs2014-jair}, which tightly integrate the solver with the external sources and reduce the number of external calls,
		the remaining calls are still expensive.
		In addition to the complexity of the external sources themselves,
		also overhead on the implementation side,
		such as calls of external libraries and cache misses after jumps out of core algorithms, may decrease efficiency compared to
		ordinary ASP-programs. 
		
		In this paper we present a \textbf{novel method for \hex-program evaluation based on inlining of external atoms}.
		In contrast to existing approaches for DL-programs~\cite{DBLP:conf/ecai/HeymansEX10,DBLP:conf/rr/XiaoE11,DBLP:conf/dlog/BajraktariOS17},
		ours is generic and can be applied to arbitrary external sources.
		Therefore, it is interesting beyond \hex-programs and also applicable to specialized formalisms such as constraint ASP~\cite{geossc09a,os2012-tplp}.
		The approach uses \emph{support sets} (cf.~e.g.~\citeN{DBLP:journals/corr/abs-1106-1819}), i.e.,
		sets of literals \rev{which}{that} define assignments of input atoms that guarantee that an external atom is true.
		Support sets were previously exploited for \hex-program evaluation~\cite{eiterFR014};
		however, this was only for speeding up but not for eliminating the necessary verification step.
		In contrast, our new approach compiles external atoms
		away altogether such that there are no guesses at all \rev{which}{that} need to be verified\rev{.}{,}
		i.e., the semantics of external atoms is embedded
		in the ASP-program.
		We use a \textbf{benchmark suite to show significant performance improvements for certain classes of external atoms}.

		Next, we have a look at equivalence notions for ASP such as \emph{strong equivalence}~\cite{Lifschitz:2001:SEL:383779.383783}, \emph{uniform equivalence}~\cite{DBLP:conf/iclp/EiterF03}
		and the more general notion of $\langle \mathcal{H}, \mathcal{B} \rangle$-equivalence~\cite{DBLP:journals/tplp/Woltran08};
		all these notions identify programs as equivalent also wrt.~program extensions.
		Equivalence notions have received quite some attention and in fact have also been developed for other formalisms
		such as abstract argumentation~\cite{DBLP:conf/ijcai/BaumannDLW17}.
		Thus it is a natural goal to also \textbf{use equivalence notions from ordinary ASP-programs for \hex-programs} (and again, also special cases thereof),
		which turns out to be possible based on our inlining approach.
		We are able to show that equivalence can be \reva{(semantically)} characterized similarly as for ordinary ASP-programs\rev{; while
		this is convenient,
		due to the support for external atoms and the use of the FLP-reduct~\cite{flp2011-ai} instead of the GL-reduct~\cite{gelf-lifs-88} in the semantics of \hex-programs, this result is not immediate.}{.
		To this end, we show that the existing criteria for equivalence of ASP-programs characterize also the equivalence of \hex-programs.}
		Based on the \rev{extended equivalence notion}{equivalence characterization of \hex-programs},
		we further derive a \textbf{(semantic) characterization of inconsistency of a program wrt.~program extensions}, which we call \emph{persistent inconsistency}.
		\reva{More precisely, due to nonmonotonicity, an inconsistent program can in general become consistent when additional rules are added.
		Our notion of persistent inconsistency captures programs which remain inconsistent even under (certain) program extensions.}
		While the main results are decision criteria based on programs and their reducts, we further derive a criterion for checking persistent inconsistency based on unfounded sets\rev{,
		which are -- due to the fact that implementations usually do not explicitly construct the reduct -- convenient with view on applications, which we also discuss.}{.
		Unfounded sets are sets of atoms which support each other only cyclically and are often used in implementations to realize minimality checks of answer sets.
		Thus, a criterion based on unfounded sets is convenient in view of practical applications in the course of reasoner development; we discuss one such application as the end \revtwo{}{of this paper}.}

		\smallskip
		\noindent To summarize the \textbf{main contributions}, we present%
		\begin{enumerate}
			\item \textbf{a technique for external source inlining} and three applications thereof, namely
			\item \textbf{a new evaluation technique for \hex-programs},
			\item \textbf{a generalization of \rev{equivalence notions}{equivalence characterizations} from ASP- to \hex-programs}, and
			\item \textbf{a novel notion of inconsistency of \hex-programs wrt.~program extensions \reva{and an according characterization}}.
		\end{enumerate}
		
		\reva{Here, item 1.~is the foundation for the contributions in items 2.,~3.,~and 4.}
		
		\smallskip
		\noindent After the preliminaries in Section~\ref{sec:preliminaries} we proceed as follows:
		\begin{itemize}
			\item In Section~\ref{sec:inlining} we show how external atoms can be \emph{inlined} (embedded) into a program.
				To handle nonmonotonicity we use a \emph{saturation encoding} based on \emph{support sets}.
				For the sake of a simpler presentation we first restrict the discussion to positive external atoms and then extend our approach to handle also negated ones.
			\item In Section~\ref{sec:implementation} we exploit this approach for performance gains.
				To this end, we implement the approach in the \dlvhex{} system and perform an experimental evaluation, which shows a significant speedup
				for certain classes of external atoms.
				The speedup is both over traditional evaluation and over a previous approach based on support sets for guess verification.
			\item \rev{Based on this inlining approach}{As another application of the inlining technique},
				Section~\ref{sec:equivalence} characterizes equivalence of \hex-programs, which generalizes results by~\citeN{DBLP:journals/tplp/Woltran08}.
				The generalizations of strong~\cite{Lifschitz:2001:SEL:383779.383783} and uniform equivalence~\cite{DBLP:conf/iclp/EiterF03} correspond to special cases thereof.
			\item In Section~\ref{sec:inconsistency} we present a characterization of \emph{inconsistency of \hex-programs wrt.~program extensions}, which we call \emph{persistent inconsistency}.
				This characterization is derived from the previously presented notion of equivalence.
				We then discuss an application of the criteria in context of potential further improvements of the evaluation algorithm.
			\item Section~\ref{sec:conclusion} discusses related work and concludes the paper.
			\item Proofs are outsourced to \ref{sec:proofs}.
		\end{itemize}

		A preliminary version of the results in this paper has been presented at AAAI~2017~\cite{r2017a-aaai,r2017b-aaai};
		the extensions in this work \rev{comprise}{consist} of more extensive discussions of the theoretical contributions, additional experiments and formal proofs of the results.

	\section{Preliminaries}
	\label{sec:preliminaries}

		\revam{We start with basic concepts.}
		Our alphabet consists of possibly infinite, mutually disjoint sets of constant symbols $\mathcal{C}$,
		predicate symbols $\mathcal{P}$, and external predicates $\mathcal{X}$;
		in this paper we \rev{resign}{refrain} from using variables in the formal part, as will be justified below.

		In the following, a (ground) \reva{ordinary} atom
		$a$ is of form $p(c_1, \dotsc, c_\ell)$
		with predicate $p \in \mathcal{P}$ and
		constant symbols $c_1, \dotsc, c_\ell \in \mathcal{C}$,
		abbreviated as $p(\vec{c})$;
		we write $c \in \vec{c}$ if $c = c_i$ for some $1 \le i \le \ell$.
		For $\ell = 0$ we might drop the parentheses and write $p()$ simply as $p$.
		\reva{In the following we may drop `ordinary' and call it simply an atom whenever clear from context.}

		An {\em assignment} $Y$ over \rev{the}{a} \revam{(finite) }set $A$ of atoms 
		is a set $Y \subseteq A$,
		where $a \in Y$ expresses that~$a$ is true under $Y$, also denoted $Y \models a$, and $a \not\in Y$ that $a$ is false, also denoted $Y \not\models a$.
		For a \emph{default-literal} $\naf a$ over an atom $a$ we let $Y \models \naf a$ if $Y \not\models a$ and $Y \not\models \naf a$ otherwise.

		\leanparagraph{\hex-Programs.}
		We recall \hex-programs~\cite{efikrs2015-tplp}, which
		generalize (disjunctive) logic programs under the answer set
		semantics~\cite{gelf-lifs-91}, as follows.

		\paragraph{Syntax.}
		\hex-programs extend ordinary ASP-programs by \emph{external atoms}
		which provide a bidirectional interface between the program
		and external sources.
		A \emph{ground external atom} is of the form 
		$\ext{g}{\vec{p}}{\vec{c}}$,
		where
		$\amp{g} \in \mathcal{X}$ is an external predicate,
		$\vec{p} = p_1, \dotsc, p_k$ is a list of input parameters (predicates from $\mathcal{P}$ or object constants from $\mathcal{C}$), called \emph{input list},
		and 
		$\vec{c} = c_1, \dotsc, c_l$ are 
		output constants from $\mathcal{C}$.
		
		\begin{definition}
			\label{def:rule}
			A \hex-program $P$ consists of rules
			\begin{equation*}
			  a_1\lor\cdots\lor a_k \leftarrow b_1,\dotsc, b_m, \naf\, b_{m+1}, \dotsc, \naf\, b_n \ ,
			\end{equation*}
			where each $a_i$ is an \reva{ordinary} atom and each~$b_j$
			is either an ordinary
			atom or an external atom.
		\end{definition}

		For such a rule $r$, its \emph{head} is $H(r) = \{ a_1, \ldots, a_k \}$,
		its \emph{body} is $B(r) = \{b_1, \dotsc, b_m,$ $\naf\, b_{m+1},$ $\dotsc,$ $\naf\, b_n\}$,
		its \emph{positive body} is $B^{+}(r) = \{ b_1, \ldots, b_m \}$
		and its \emph{negative body} is $B^{-}(r) = \{ b_{m+1}, \ldots, b_n \}$.
		For a program $P$ we let $X(P) = \bigcup_{r \in P} X(r)$
		for $X \in \{ H, B, B^{+}, B^{-} \}$.
		
		For a program $P$ and a set of constants $\mathcal{C}$,
		let $\mathit{HB}_{\mathcal{C}}(P)$ denote the \emph{Herbrand base}
		containing all atoms constructible from the predicates occurring in $P$ and constants $\mathcal{C}$.

		We restrict the formal discussion to programs without variables
		as suitable safety conditions guarantee the existence of a finite grounding \rev{which}{that}
		suffices for answer set computation, see e.g.~\citeN{efkr2016-aij}.

		\paragraph{Semantics.}
		In the following, assignments are over the set \rev{$A(P)$ of ordinary atoms 
		that occur in the program $P$ at hand}{of ordinary atoms constructible from predicates $\mathcal{P}$ and constants $\mathcal{C}$}.
		The semantics of an external atom $\ext{g}{\vec{p}}{\vec{c}}$.
		wrt.~an assignment $Y$ is given by the value of a decidable $1{+}k{+}l$-ary
		\emph{two-valued (Boolean) oracle function} $\extfun{g}$ that is defined for all possible values
		of $Y$, $\vec{p}$ and $\vec{c}$.  We say that
		$\ext{g}{\vec{p}}{\vec{c}}$ is true relative
		to $Y$ if
		$\extsem{g}{Y}{\vec{p}}{\vec{c}} = \T$,
		and it is false otherwise.
		\revam{
		We make the restriction that for a fixed assignment $Y$ and input list $\vec{p}$,
		$\extsem{g}{Y}{\vec{p}}{\vec{c}} = \T$ holds only for finitely many different vectors $\vec{c}$.}
		\reva{We make the restriction that $\extsem{g}{Y}{\vec{p}}{\vec{c}} = \extsem{g}{Y'}{\vec{p}}{\vec{c}}$
		for all assignments $Y$ and $Y'$ which coincide on all atoms over predicates in $\vec{p}$.
		That is, only atoms over the predicates in $\vec{p}$ may influence the value of the external atom,
		which resembles the idea of $\vec{p}$ being the `input' to the external source;
		we call such atoms also the \emph{input atoms} of $\ext{g}{\vec{p}}{\vec{c}}$.}
		
		Satisfaction of ordinary  rules and ASP-programs~\cite{gelf-lifs-91}
		is then extended to 
		\hex-rules and \rev{}{-}programs as follows.
		A rule $r$ as by Definition~\ref{def:rule} is true under $Y$, denoted $Y \models r$,
		if $Y \models h$ for some $h \in H(r)$ or $Y \not\models b$ for some $b \in B(r)$.		

		The answer sets of a \hex-program $P$ are
		defined as follows.
		Let the \emph{FLP-reduct}
		of $P$ wrt.~an assignment $Y$
		be the set $f P^{Y} = \{ r \in P \mid Y \models b \text{ for all } b \in B(r) \}$.
		Then:
		\begin{definition}
			\label{def:answerset}
			An assignment $Y$ is an answer set of a \hex-program $P$\revam{,} if
			$Y$  is a subset-minimal model of  
			the FLP-reduct $f P^Y$ of $P$ wrt.~$Y$.
		\end{definition}

		\begin{example}
			\label{ex:id}
			Consider the program 
			$P = \{ p \leftarrow \ext{\mathit{id}}{p}{} \}$,
			where $\ext{\mathit{id}}{p}{}$ is true iff $p$ is true. Then 
			$P$ has the answer set $Y_1 = \emptyset$; indeed it is
			a subset-minimal model of~$f P^{Y_1} = \emptyset$.
		\end{example}

		For an ordinary program $P$, the above definition of answer sets is equivalent to Gelfond~\& Lifschitz' answer sets.

		\leanparagraph{Traditional Evaluation Approach.}
		A \hex-programs $P$ 
		is transformed to an ordinary ASP-program $\hat{P}$ as follows.
		Each external atom
		$\ext{g}{\vec{p}}{\vec{c}}$
		in~$P$ is replaced by an ordinary \emph{replacement atom}
		$e_{\amp{g}[\vec{p}]}(\vec{c})$
		and a rule~
		$e_{\amp{g}[\vec{p}]}(\vec{c}) \vee \mathit{ne}_{\amp{g}[\vec{p}]}(\vec{c}) \leftarrow$
		is added.
		The answer sets of the resulting \emph{guessing program} $\hat{P}$
		are computed by an 
		ASP solver. However, the assignment $Y$
		extracted from an answer set $\hat{Y}$ of $\hat{P}$
		by projecting it to the ordinary atoms $A(P)$ in $P$
		may not satisfy $P$
		as $\amp{g}[\vec{p}](\vec{c})$  under $f_{\amp{g}}$ may differ
		from the guessed value of $e_{\amp{g}[\vec{p}]}(\vec{c})$.
		The answer set is merely a \emph{candidate}.
		If a compatibility check against the external source
		succeeds,
		it is a \emph{compatible set} as formalized as follows:
		\begin{definition}
			\label{def:compatibleset}
			A \emph{compatible set} of a program $P$
			is an
			answer set $\hat{Y}$
			of the guessing program $\hat{P}$
			such that
			$\extsem{g}{\hat{Y}}{\vec{p}}{\vec{c}} = \T$ iff
			$e_{\amp{g}[\vec{p}]}(\vec{c}) \in \hat{Y}$ for
			all external atoms $\amp{g}[\vec{p}](\vec{c})$ in $P$.
		\end{definition}
		
		\begin{example}
			\label{ex:id-contd}
			Consider
			$P = \{\, p(a) \vee p(b) \leftarrow \ext{\mathit{atMostOne}}{p}{}\,\}$,
			where $\ext{\mathit{atMostOne}}{p}{}$ is true under an assignment $Y$ if $\{ p(a), p(b) \} \nsubseteq Y$, i.e., at most one of $p(a)$ or $p(b)$ is true under $Y$, and it is false otherwise.
			Then we have $\hat{P} = \{ p(a) \vee p(b) \leftarrow e_{\amp{\mathit{atMostOne}}[p]};\; e_{\amp{\mathit{atMostOne}}[p]} \vee \mathit{ne}_{\amp{\mathit{atMostOne}}[p]} \leftarrow \}$,
			which has the answer sets $\hat{Y_1} = \{ p(a), \mathit{e}_{\amp{\mathit{atMostOne}}[p]} \}$, $\hat{Y_2} = \{ p(b), \mathit{e}_{\amp{\mathit{atMostOne}}[p]} \}$,
			$\hat{Y_3} = \{ \mathit{ne}_{\amp{\mathit{atMostOne}}[p]}{}\}$ (while $\{ p(a), p(b), \mathit{e}_{\amp{\mathit{atMostOne}}[p]} \}$ is not an answer set of $\hat{P}$).
			However, although $\hat{Y_3}$ is an answer set of $\hat{P}$,
			its projection $Y_3 = \emptyset$ to atoms $A(P)$ in $P$ is not an answer set of $P$ because $Y_3 \models \ext{\mathit{atMostOne}}{p}{}$ but $e_{\amp{\mathit{atMostOne}}[p]} \not\in \hat{Y_3}$,
			and thus the compatibility check for $\hat{Y_3}$ fails.
			In contrast, the compatibility checks for $\hat{Y_1}$ and $\hat{Y_2}$ pass, i.e., they are compatible sets of $P$, and their
			projections $Y_1 = \{ p(a) \}$ and $Y_2 = \{ p(b) \}$ to atoms $A(P)$ in $P$ are answer sets of $P$.
		\end{example}

		However, if the compatibility check succeeds, the projected interpretation is not always automatically an answer set of the original program.
		Instead, after the compatibility check of an answer set $\hat{Y}$ of $P$ was passed, another final check is needed to guarantee also subset-minimality of its projection $Y$ wrt.~$f P^{Y}$.
		Each answer set $Y$ of $P$ is the projection of 
		some compatible set $\hat{Y}$ to
		$A(P)$, but not vice versa.

		\begin{example}
			\label{ex:id-contd}
			Reconsider
			$P = \{\, p \leftarrow \ext{\mathit{id}}{p}{}\,\}$
			from above.
			Then $\hat{P} = \{ p \leftarrow e_{\amp{\mathit{id}}[p]}();\;
				e_{\amp{\mathit{id}}[p]}{} \vee \mathit{ne}_{\amp{\mathit{id}}[p]}{} \leftarrow~ \}$
			has the answer sets $\hat{Y_1} = \{ \mathit{ne}_{\amp{\mathit{id}}[p]}{} \}$ and 
			$\hat{Y_2} = \{ p, e_{\amp{\mathit{id}}[p]}{}\}$. 
			Here, $Y_1 = \emptyset$ is a $\subseteq$-minimal model of $f
			P^{Y_1}\,{=}\, \emptyset$, 
			but $Y_2 = \{ p \}$ not of $f P^{Y_2} = P$.
		\end{example}
		
		There are several approaches for checking this minimality,
		e.g.~based on \emph{unfounded sets}, which are sets of atoms that support each other only cyclically~\cite{faber2005-lpnmr}.
		However, the details of this check are not relevant for this paper,
		which is why we refer the interested reader to~\citeN{efkrs2014-jair}
		for a discussion and evaluation of various approaches.

		\leanparagraph{Learning Techniques.}
		In practice, the guessing program $\hat{P}$ has usually many answer sets, but many of them fail the compatibility check against external sources (often because of the same wrong guess),
		which turns out to be an evaluation bottleneck.
		To overcome the problem, techniques \rev{which}{that} extend \emph{conflict-driven learning} have been introduced as \emph{external behavior learning (EBL)}~\cite{efkr2012-tplp}.
		
		As in ordinary ASP solving, the traditional \hex-algorithm translates the guessing program to a set of \emph{nogoods}, i.e., a set of literals \rev{which}{that} must not be true at the same time.
		Given this representation, techniques from SAT solving are applied to find an assignment \rev{which}{that} satisfies all nogoods~\cite{gks2012-aij}.
		Notably, as the encoding as a set of nogoods is of exponential size due to \emph{loop nogoods} \rev{which}{that} avoid cyclic justifications of atoms, those parts are generated only on-the-fly.
		Moreover, additional nogoods are learned from conflict situations, i.e., violated nogoods \rev{which}{that} cause the solver to backtrack; this is called \emph{conflict-driven nogood learning}, see \rev{e.g.~\cite{FM09HBSAT}}{e.g.~\citeN{FM09HBSAT}}.
		
		EBL extends this algorithm by learning additional nogoods not only from conflict situations in the ordinary part, but also from verification calls to external sources.
		Whenever an external atom $e_{\amp{e}[\vec{p}]}(\vec{c})$ is evaluated under an assignment $Y$ for the sake of compatibility checking, the actual truth value under the assignment becomes evident.
		Then, regardless of whether the guessed value was correct or not,
		one can add a nogood \rev{which}{that} represents that $e_{\amp{e}[\vec{p}]}(\vec{c})$ must be true under $Y$ if $Y \models \ext{e}{\vec{p}}{\vec{c}}$
		or that $e_{\amp{e}[\vec{p}]}(\vec{c})$ must be false under $Y$ if $Y \not\models \ext{e}{\vec{p}}{\vec{c}}$.
		If the guess was incorrect, the newly learned nogood will trigger backtracking, if the guess was correct,
		the learned nogood will prevent future wrong guesses.
					
		\begin{example}\label{ex:conflictdriven}
			Suppose $\ext{\mathit{atMostOne}}{p}{}$ is evaluated under $Y = \{ p(a), p(b) \}$.
			Then the real truth value of $\ext{\mathit{atMostOne}}{p}{}$ under $Y$ becomes evident: in this case $Y \not\models \ext{\mathit{atMostOne}}{p}{}$.
			One can then learn the nogood $\{ p(a), p(b), e_{\amp{\mathit{atMostOne}}[p]}() \}$ to represent that $p(a)$, $p(b)$ and $\ext{\mathit{atMostOne}}{p}{}$ cannot be true at the same time.
		\end{example}	

		Learning realizes a tight coupling of the reasoner and the external source
		by adding parts of the semantics on-demand to the program instance, which is similar to theory propagation in SMT (see e.g.~\citeN{Nieuwenhuis05theorypropagation})
		\reva{and lazy clause generation~\cite{Ohrimenko:2009:PVL:1553323.1553342,DBLP:conf/iclp/DrescherW12}. However, while these approaches consider only specific theories
		such as integer constraints, EBL in \hex{} \revtwo{support}{supports} arbitrary external sources. Moreover, EBL does not depend on application-specific procedures for generating
		learned clauses but rather derives them from the observed behavior of the source}.
		Experimental results show that \rev{this technique}{EBL} leads to a significant, up to exponential speedup, which is explained by
		the exclusion of up to exponentially many guesses by the learned nogoods,
		but the remaining verification calls are still expensive and -- depending on the type of the external source -- can account for large parts of the overall runtime~\cite{efkrs2014-jair}.

		\leanparagraph{Evaluation Based on Support Sets.}
		Later, an alternative evaluation approach was developed.
		While the basic idea of guessing the values of external atoms as in the traditional approach remains,
		the verification is now accomplished by using so-called \emph{support sets} instead of explicit evaluation~\cite{eiterFR014}.
		Here, a \emph{positive resp.~negative support set} for an external atom $e$ is a set of literals over the input atoms of $e$
		whose satisfaction implies satisfaction resp.~falsification of $e$.
		Informally, the verification is done by checking \rev{if}{whether} the answer set candidate matches with a support set of the external atom.
		If this is the case, the guess is verified resp.~falsified.
		
		More precisely, for a set $S$ of literals $a$ or $\neg a$, where $a$ is an atom,
		let $\neg S = \{ \neg a \mid a \in S \} \cup \{ a \mid \neg a \in S \}$
		be the set of literals $S$ with swapped sign.
		\reva{We call a set $S$ of literals \emph{consistent} if there is no atom $a$ such that $\{ a, \neg a \} \subseteq S$}.
		We formalize support sets as follows:
		\begin{definition}[Support Set]
			\label{def:supportset}
			Let $e = \ext{g}{\vec{y}}{\vec{x}}$ be an external atom in a program $P$.
			A \emph{support set} for $e$ is a consistent set $S_{\sigma} = S_{\sigma}^{+} \cup S_{\sigma}^{-}$ with $\sigma \in \{ \T, \F \}$,
				$S_{\sigma}^{+} \subseteq \mathit{HB}_{\mathcal{C}}(P)$,
				and
				$S_{\sigma}^{-} \subseteq \neg\mathit{HB}_{\mathcal{C}}(P)$
				s.t. $Y \supseteq S_{\sigma}^{+}$ and $Y \cap \neg S_{\sigma}^{-} = \emptyset$ implies $Y \models e$ if $\sigma = \T$ and $Y \not\models e$ if $\sigma = \F$
				for all assignments $Y$.
		\end{definition}

		We call the support set $S_{\sigma}$ \emph{positive} if $\sigma = \T$ and \emph{negative} if $\sigma = \F$.
		
		 \begin{example}
		 \label{ex:grsupset}
			Suppose $\ext{\mathit{diff}}{p, q}{c}$ computes the set of all elements $c$ \rev{which}{that} are in the extension\reva{\footnote{The \emph{extension} of a (unary) predicate $p$ wrt.~an assignment $Y$ is
			the set $\{ c \mid p(c) \in Y \}$; likewise for predicates with other arities.}} of $p$ but not in that of $q$.
			Then $\{ p(a), \neg q(a) \}$ is a positive support set for $\ext{\mathit{diff}}{p, q}{a}$
			because any assignment $Y$ with $\{ p(a) \} \subseteq Y$ but $Y \cap \{ q(a) \} = \emptyset$ satisfies $\ext{\mathit{diff}}{p, q}{a}$.
		\end{example}

		We are in particular interested in \emph{families (=sets) of support sets}
		which describe the behavior of external atoms completely:

		\begin{definition}[(Complete) Support Set Family]
			\label{def:supportsetfamily}
			A \emph{positive resp.~negative family of support sets} $\mathcal{S}_{\sigma}$ with $\sigma \in \{ \T, \F \}$ for external atom $e$
			is a set of positive resp.~negative support sets of $e$;
			$\mathcal{S}_{\sigma}$ is \emph{complete} if for each
			assignment $Y$ with $Y \models e$ resp.~$Y \not\models e$
			there is an $S_{\sigma} \in \mathcal{S}_{\sigma}$ s.t.~$Y \supseteq S_{\sigma}^{+}$ and $Y \cap \neg S_{\sigma}^{-} = \emptyset$.
		\end{definition}
		
		Complete support set families $\mathcal{S}_{\sigma}$ can be used for the verification of external atoms as follows.
		One still uses the rewriting $\hat{P}$, but instead of explicit evaluation and comparison of the guess of a replacement atom to the actual value
		under the current assignment,
		one checks \rev{if}{whether} for some $S_{\sigma} \in \mathcal{S}_{\sigma}$ we have $Y \supseteq S_{\sigma}^{+}$ and $Y \cap \neg S_{\sigma}^{-} = \emptyset$ for the current assignment $Y$.
		If this is the case, the external atom must be true if $\sigma = \T$ and false if $\sigma = \F$;
		otherwise, it must be false if $\sigma = \T$ and true if $\sigma = \F$.
		This method is in particular advantageous if the support sets in $\mathcal{S}_{\sigma}$ are \emph{small and few}.
		
		As a further improvement,
		positive support sets $S_{\T}$ for $\ext{g}{\vec{p}}{\vec{c}}$ can
		be added as constraints $\leftarrow S^{+}_{\T}, \{ \naf a \mid \neg a \in S^{-}_{\T} \}, \naf \ext{g}{\vec{p}}{\vec{c}}$
		to the program
		in order to exclude false negative guesses.
		Analogously, for negative support sets
		we can add $\leftarrow S^{+}_{\F}, \{ \naf a \mid \neg a \in S^{-}_{\F} \}, \ext{g}{\vec{p}}{\vec{c}}$
		to exclude false positive guesses.
		This \revam{improvement }was exploited in existing approaches for performance improvements~\cite{eiterFR014};
		we will also use this technique in Section~\ref{sec:implementation}
		when comparing our new approach to the previous support-set-based approach.
		This amounts to a learning technique similar to EBL.
		However, note that this learns only a fixed number of nogoods at the beginning, while learning by EBL is not \rev{possible}{done} here as external sources are not evaluated during solving.
		\rev{
		Also note that the verification check is still necessary even if all $S_{\sigma} \in \mathcal{S}_{\sigma}$ are added as
		constraints because these constraints prevent only false negative guesses for $\sigma = \T$ and false positive guesses for $\sigma = \F$, but not vice versa.
		}{Note that even if \emph{all} $S_{\T} \in \mathcal{S}_{\T}$ are added as constraints, the verification check is still necessary.
		This is because adding a positive support set $S_{\T}$ as a constraint eliminates only false negative guesses,
		but not false positive guesses (since they encode only when the external atom is true but not when it is false).
		Conversely, adding all $S_{\F} \in \mathcal{S}_{\F}$ prevents only false positive guesses but not false negative ones.
		}

		The approach was also lifted to the non-ground level~\cite{eiterFR014}.
		Intuitively, non-ground support sets may contain variables
		as shortcuts for all ground instances. Prior to the use of non-ground support sets, the variables are substituted
		by all relevant constants \rev{which}{that} appear in the program.
		However, in the following we
		restrict the formal discussion to the ground level for simplicity.

		To summarize, improvements in the traditional evaluation approach (learning) have reduced the number of verification calls,
		and the alternative support set approach has replaced explicit verification calls
		by matching an assignment with support sets, but neither of them did eliminate the need for guessing and subsequent verification altogether.
		In the next section we go a step further and eliminate this need.

		\revmv{
		\leanparagraph{Construction of Support Sets.}
		Obviously, in order to make use of support sets
		there must be procedures that can effectively and efficiently
		construct them, which is why we have a look at this aspect\revam{ first}.
		Constructing support sets depends on the external source~\cite{eiterFR014}.
		In general, the developer of an external atom is aware
		of its semantic structure, which usually allows her/him
		to provide this knowledge in form of support sets.
		Then, providing support sets can be seen as an alternative way
		to define and implement oracle functions.
		For certain classes of external atoms,
		procedures for constructing support sets are in fact already in place.

		\rev{Compactness of families of support sets is an important aspect for our contribution.
		It is crucial for our contribution that,
		although our encoding may be exponential in the worst case due to exponentially many support sets,
		many realistic external sources have small support set families}
		{Compactness of families of support sets is an important aspect for evaluation techniques based on families of support sets.
		It is therefore crucial for the approach by~\citeN{eiterFR014} and our contribution that,
		although there may be exponentially many support sets in the worst case,
		many realistic external sources have small support set families.}
		For certain types of external sources, their small size is
		even \emph{provable} and \emph{known before evaluating the program}.
		External sources with provably small support set families include, for instance, the description logic $\dllite_{\cA}$~\cite{clmr2007}.
		\revam{As the number of support sets is directly linked to the size of our rewriting, also the latter is guaranteed to be small in such cases.}
		Generally, support set families tend to be small for sources whose behavior is structured,
		i.e., whose output often depends only on parts of the input and does not change completely with small changes in the input~\cite{eiterFR014}.
		Note that such a structure in many realistic applications is also the key to parameterized complexity.
		In this paper, we focus on such sources; also the sources used in our benchmarks are guaranteed to have small families of support sets
		(whose sizes we will discuss together with the respective benchmark results).

		As an example we have a closer \reva{look} at constructing support sets for a $\dllite_{\cA}$-ontology \rev{, which}{that} is accessed from the logic program
		using dedicated external atoms
		(also called \emph{DL-atoms}~\cite{eilst2008-aij}).
		DL-atoms allow for answering queries over the ontology under a (possibly) extended Abox based on input from the program.
		We use \revtwo{an external atom of form}{the external atom} $\amp{\mathit{DL}}[\mathit{ont}, \mathit{inpc}, \mathit{inpr}, \mathit{con}](X)$
		to access an ontology $\mathit{ont}$ and retrieve all individuals $X$ in the concept $\mathit{con}$, where
		the binary resp.~ternary predicates $\mathit{inpc}$ and $\mathit{inpr}$ allow for answering the query under the assumption that certain concept resp.~role assertions are added
		to the Abox of the ontology before answering the query.		
		More precisely, the query is answered wrt.~an assignment $Y$ under the assumption that
		concept assertion $c(i)$ is added for each $\mathit{inpc}(c, i) \in Y$
		and
		role assertion $r(i_1, i_2)$ is added for each $\mathit{inpr}(r, i_1, i_2) \in Y$.

		For instance,
		suppose the program contains atoms of form $\mathit{inpc}(``\mathit{Person}", \cdot)$ to specify persons
		and atoms of form $\mathit{inpr}(``\mathit{childOf}", \cdot, \cdot)$ to specify parent-child relations.
		Then the external atom
		$\amp{\mathit{DL}}[\mathit{ont}, \mathit{inpc}, \mathit{inpr}, ``\mathit{OnlyChild}"](X)$
		queries all members of concept $\mathit{OnlyChild}$ under the assumption that concepts $\mathit{Person}$ and \revtwo{rols}{roles} $\mathit{childOf}$
		has been extended according to the truth values of the $\mathit{inpc}$ and $\mathit{inpr}$ atoms in the program.%
		\footnote{
				This is often written as
				$\mathit{DL}[\mathit{ont}; \mathit{Person} \uplus p, \mathit{childOf} \uplus c; \mathit{OnlyChild}](X)$
				using a more convenient syntax tailored to DL-atoms,
				where additions to concepts and rules \revtwo{is}{are} expressed by operator $\uplus$
				and $p$ and $c$ are unary and binary (instead of binary and ternary) predicates, respectively.
			}
		
		For this type of description logic,~\citeN{clmr2007} have proven that
		at most one assertion is needed to derive an instance query from a consistent ontology.
		Hence, for each concept $c$ and
		individual $i$ there is a (positive) support set either of form $\emptyset$ or of form $\{ p(\vec{x}) \}$,
		where the latter encodes that if $p(\vec{x}) \in Y$, then $Y \models \amp{\mathit{DL}}[\mathit{ont}, \mathit{inpc}, \mathit{inpr}, c](i)$
		for all assignments $Y$.
		Moreover, at most two added ABox assertions are needed to make such an ontology inconsistent (in which case all queries are true).
		For each possibility where the ontology becomes inconsistent there
		is a (positive) support set of form $\{ p(\vec{x}), p'(\vec{x'}) \}$.
		Then, each support set is of one of only three different forms, which are all at most binary.
		Moreover,~\citeN{Lembo2011} have proven that the number of different constants appearing in $\vec{x}$ resp.~$\vec{x'}$ in these
		support sets is limited by three.
		The limited cardinality and number of constants also limits the number of possible support sets required to describe the overall ontology to a quadratic number
		in the size of the program and the Abox.

		Moreover, as one can see, the support sets are easy to construct by a syntactic analysis of the ontology and the DL-atoms.
		For details regarding the construction of support sets for $\dllite_{\cA}$ we refer to~\citeN{Eiter:2014:TPD:3006652.3006701}.
		}

	\section{External Source Inlining}
	\label{sec:inlining}

		In this section we present a rewriting which compiles \hex-programs into equivalent ordinary ASP-programs (modulo auxiliary atoms)
		based on support sets,
		and thus embeds external sources into the program; we call the technique \emph{inlining}.
		Due to nonmonotonic behavior of external atoms, inlining is not straightforward.
		In particular, it is \emph{not} sufficient to substitute external atoms by ordinary replacement atoms and derive their truth values based on their support sets,
		which is surprising at first glance.
		Intuitively, this is because rules \rev{, which}{that} define replacement atoms\revam{,} can be missing in the reduct and it is not guaranteed any longer that the replacement atoms resemble the original semantics;
		we will demonstrate this in more detail in Section~\ref{sec:inlining:attempts}.
		Afterwards we present a sound and complete encoding based on the saturation technique (cf.~e.g.~\citeN{eik2009-rw}) in Section~\ref{sec:inlining:encoding}.

		\subsection{Observations}
		\label{sec:inlining:attempts}

			We \revam{now }start with observations that can be made when attempting to inline external sources in a straightforward way.
			The first intuitive attempt to inline an external atom $e$ might be to replace it by an ordinary atom $x_e$
			and add rules of kind $x_e \leftarrow L$, where $L$ is constructed from a positive support set $S_{\T}$ of $e$
			by adding $S^{+}_{\T}$ as positive atoms and $S^{-}_{\T}$ as default-negated ones.
			However, this alone is in general incorrect even if repeated for all $S_{\T} \in \mathcal{S}_{\T}$ for a complete family of support sets $\mathcal{S}_{\T}$,
			as the following example demonstrates.
			
			\begin{example}
				\label{ex:encodingAttempt1}
				Consider $P = \{ a \leftarrow \ext{\mathit{true}}{a}{} \}$ where $e = \ext{\mathit{true}}{a}{}$ is always true\reva{;
				a complete family of positive support sets is $\mathcal{S}_{\T} = \{ \{ a \}, \{ \neg a \} \}$}.
				The program is expected to have the answer set $Y = \{ a \}$.
				However, the translated program $P' = \{ x_e \leftarrow a; \ x_e \leftarrow \naf a; \ a \leftarrow x_e \}$ has no answer set because the only candidate is $Y' = \{ a, x_e \}$
				and $f P'^Y = \{ x_e \leftarrow a; \ a \leftarrow x_e \}$ has the smaller model $\emptyset$.
			\end{example}
			
			In the example, $P'$ fails to have an answer set because the former external atom $\ext{\mathit{true}}{a}{}$
			is true also if $\naf a$ holds, but the rule $x_e \leftarrow \naf a$, which represents this case, is dropped from the reduct wrt.~$Y'$ because its body $\naf a$
			is unsatisfied by $Y'$. Hence, although the external atom $e$ holds both under $Y'$ and under the smaller model $\emptyset$ of the reduct which dismisses $Y'$,
			this is not detected since the representation of the external atom in the reduct is incomplete. In such a case, the value of $x_e$ and $e$ under a model of the reduct can differ.
			
			An attempt to fix this problem might be to explicitly guess the value of the external atom and represent both when it is true and when it is false.
			Indeed, $P'' = \{ x_e \vee \overline{x_e} \reva{\leftarrow}; \ \leftarrow a, \naf x_e; \ \leftarrow \naf a, \naf x_e; \ a \leftarrow x_e \}$ is a valid rewriting of the previous program
			($Y'$ is an answer set). However, this rewriting is also incorrect in general, as the next example shows.
			
			\begin{example}
				\label{ex:encodingAttempt2}
				Consider $P = \{ a \leftarrow \ext{\mathit{id}}{a}{} \}$ where $e = \ext{\mathit{id}}{a}{}$ is true iff $a$ is true.
				The program is expected to have the answer set $Y = \emptyset$.
				However, the translated program $P' = \{ x_e \vee \overline{x_e} \reva{\leftarrow}; \ \leftarrow a, \naf x_e; \ \leftarrow \naf a, x_e; \ a \leftarrow x_e \}$ has not only the intended answer set $\{ \overline{x_e} \}$
				but also $Y' = \{ a, x_e \}$ because $f P'^{Y'} = \{ x_e \vee \overline{x_e} \reva{\leftarrow}; \ a \leftarrow x_e \}$ has no smaller model.
			\end{example}
			
			While the second rewriting attempt from Example~\ref{ex:encodingAttempt2} works for Example~\ref{ex:encodingAttempt1},
			and, conversely, the one applied in Example~\ref{ex:encodingAttempt1} works for Example~\ref{ex:encodingAttempt2}, a general rewriting schema must be more elaborated.
			\rev{In fact, since \hex-programs with recursive, nonmonotonic external atoms are on the second level of the polynomial hierarchy~\cite{flp2011-ai},
			such a rewriting must involve disjunctions with head-cycles.
			}{
			
			In fact, since \hex-programs with recursive nonmonotonic external atoms are on the second level of the polynomial hierarchy,
			we present a rewriting which involves head-cycles.
			}%
			\reva{%
			Before we start, let us first discuss this aspect in more detail.
			\citeN{flp2011-ai} reduced 2QBF polynomially to a program without disjunctions but with nonmonotonic aggregates, which are special cases of external atoms.
			This, together with a membership proof, shows that programs with external atoms are complete for the second level,
			even in the disjunction-free case.
			Since ordinary ASP-programs without head-cycles are only complete for the first level,
			this implies that a further polynomial reduction to ordinary ASP must introduce disjunctions with head-cycles.
			
			Interestingly, all aggregates used by \citeN{flp2011-ai} depend only on two input atoms each,
			which implies that they can be described by a complete family of support sets of constant size (at most two support sets are needed if an optimal encoding is used).
			This shows that \hex-programs are already on the second level even if they are disjunction-free and all external atoms can be described by families of support sets with constant size.

			The size of the encoding we are going to present depends linearly on the size of the given complete family of support sets;
			since there can be exponentially many support sets even for polynomial external sources (e.g.~for the parity function),
			this can lead to an exponential encoding.
			However, for polynomial families of support sets our encoding remains polynomial as well.
			Because \hex-programs are already on the second level even if they are disjunction-free and all external atoms can be
			described by families of support sets with constant size (as discussed above),
			this is only possible because our rewriting to ordinary ASP uses head-cycles.
			}

		\subsection{Encoding in Disjunctive ASP}
		\label{sec:inlining:encoding}

			\rev{We now present such a general rewriting.}{
			In this section we present a general rewriting for inlining external atoms.
			}
			In the following, for an external atom $e$ in a program $P$, let $I(e, P)$ be the set of all ordinary atoms
			in $P$ whose predicate occurs as a predicate parameter in $e$, i.e., the set of all input atoms to $e$.
			\rev{Let further}{Furthermore, let} $\mathcal{S}_{\T}(e, P)$ be an arbitrary but fixed complete positive support set family over atoms in $P$.

			For a simpler presentation\revam{,} we proceed in two steps. We first restrict the discussion to positive external atoms,
			and then extend the approach to negative ones in Section~\ref{sec:inlining:encoding:negative}.
		
			\subsubsection{Inlining Positive External Atoms}
			\label{sec:inlining:encoding:positive}
		
				We present the encoding for inlining single positive external atoms into a program
				and explain it rule by rule afterwards.
				\reva{In the following, a \emph{new atom} is an atom that does not occur in the program $P$
					at hand and such that its predicate does not occur in the input list
					of any external atom in $P$ (but its building blocks occur in the vocabulary).
					This insures that inlining does not introduce any undesired interference with
					existing parts of the program.}

				\begin{definition}[External Atom Inlining]
					\label{def:inlining}
					For a \hex-program $P$ and external atom $e$ \rev{which}{that} occurs only positively in $P$, let
					\begin{flalign}
						P_{[e]} = & \ \{x_e \leftarrow S_{\T}^{+} \cup \{ \overline{a} \mid \neg a \in S_{\T}^{-} \} \mid S_{\T} \in \mathcal{S}_{\T}(e, P) \} \hspace{-3mm} & \label{def:inlining:1} \\
												\cup & \ \{ \overline{a} \leftarrow \naf a; \overline{a} \leftarrow x_e; a \vee \overline{a} \leftarrow \naf \overline{x_e} \mid a \in I(e, P) \} & \label{def:inlining:2} \\
												\cup & \ \{ \overline{x_e} \leftarrow \naf x_e \} & \label{def:inlining:3} \\
												\cup & \ P|_{e \rightarrow x_e} & \label{def:inlining:4}
					\end{flalign}
					where $\overline{a}$ is a new atom for each $a$, $x_e$ and $\overline{x_e}$ are new atoms for \revtwo{}{external atom} $e$,
					and $P|_{e \rightarrow x_e} = \bigcup_{r \in P} r|_{e \rightarrow x_e}$
					where $r|_{e \rightarrow x_e}$ denotes rule $r$ with every occurrence of $e$ replaced by $x_e$.
				\end{definition}

				The rewriting works as follows. The atom $x_e$ represents the former external atom, i.e., that $e$ is true, while $\overline{x_e}$ represents that it is false.
				The rules in (\ref{def:inlining:1}) represent all input assignments \rev{which}{that} satisfy $x_e$ (resp.~$e$).
				More specifically, each rule in $\{x_e \leftarrow S_{\T}^{+} \cup \{ \overline{a} \mid \neg a \in S_{\T}^{-} \} \mid S \in \mathcal{S}_{\T}(e, P) \}$
				represents one possibility to satisfy the former external atom $e$, using the complete positive family of support sets $\mathcal{S}_{\T}$\rev{.}{; in each such case $x_e$ is derived.}
				Next, for an input atom $a$, the atom $\overline{a}$ represents that $a$ is false \emph{or} that $x_e$ (resp.~$e$) is true, as formalized by the rules~(\ref{def:inlining:2}).
				\rev{This is}{The latter is} in order to ensure that for an assignment $Y$, all relevant rules in (\ref{def:inlining:1}), i.e.~those \rev{which}{that} might apply to subsets of $Y$, are contained in the reduct wrt.~$Y$
				(because $a$ could become false in a smaller model of the reduct);
				recall that in Example~\ref{ex:encodingAttempt1} the reason for incorrectness of the rewriting was
				exactly that these rules were dropped.
				The derivation of $\overline{a}$ despite $a$ being true is only necessary if $x_e$ is true wrt.~$Y$; if $x_e$ is false then all rules containing $x_e$ are dropped from the reduct anyway.
				The idea amounts to a saturation encoding~\cite{eik2009-rw}\revam{; the use of disjunctions with head-cycles is in general unavoidable for complexity reasons (unless the polynomial hierarchy collapses)}.
				Next, rule~(\ref{def:inlining:3}) enforces $\overline{x_e}$ to be true whenever $x_e$ is false.
				Finally, rules~(\ref{def:inlining:4}) resemble the original program with $x_e$ in place of $e$.

				For the following Proposition~\ref{prop:externalAtomInlining} we first assume that the complete family of support sets $\mathcal{S}_{\T}(e, P)$
				\rev{is chosen such that}{contains only support sets that contain all input atoms of $e$ in $P$ explicitly in positive or negative form. That is,}
				for all $S_{\T} \in \mathcal{S}_{\T}(e, P)$
				we have that $S^{+}_{\T} \cup \neg S^{-}_{\T} = I(e, P)$\revam{, i.e., all input atoms to $e$ in $P$ are explicitly constrained to be true or false}.
				Note that each complete family of support sets can be modified to fulfill this criterion: replace each $S_{\T} \in \mathcal{S}_{\T}(e, P)$
				with $S^{+}_{\T} \cup \neg S^{-}_{\T} \subsetneq I(e, P)$ by \rev{the set of all support sets constructible by
				distributing undefined atoms  to $S^{+}_{\T}$ or $S^{-}_{\T}$ in all possible ways.}{all of the support sets
				$\mathcal{C} = \{ S^{+}_{\T} \cup S^{-}_{\T} \cup R \mid R \subseteq U \cup \neg U, R \text{ consistent} \}$
				where $U = I(e, P) \setminus (S^{+}_{\T} \cup \neg S^{-}_{\T})$.
				These are all the support sets constructible by adding `undefined atoms' (those which occur neither positively nor negatively in $S_{\T}$)
				either in positive or negative form in all possible ways.
				The intuition is that $S_{\T}$ encodes the following condition for satisfaction of $e$: all of $S^{+}_{\T}$ but none of $S^{-}_{\T}$ must be true, while
				the value of the atoms $U$ are irrelevant for satisfaction of $e$. Thus, adding the atoms from $U$ in all combinations of positive and negative polarities makes it only explicit
				that $e$ is true in all of these cases.				
				Formally, this means that for any $Y \subseteq I(e, P)$ we have that
				$Y \supseteq S^{+}_{\T}$ and $Y \cap \neg S^{-}_{\T} = \emptyset$
				iff
				$Y \supseteq C^{+}_{\T}$ and $Y \cap \neg C^{-}_{\T} = \emptyset$
				for some $C \in \mathcal{C}$.
				}
				This might lead to an exponential blowup of the size of the family of support sets, but is made in order to simplify the first result and its proof;
				however, we show below that the result still goes through without this blowup.

				\rev{}{We show now that for such families of support sets the rewriting is sound and complete.}
				\reva{Here, we say that the answer sets of programs $P$ and $Q$ are \emph{equivalent modulo a set of atoms $A$},
				if there is a one-to-one correspondence between their answer sets
				in the sense that every answer set of $P$ can be extended to one of $Q$ in a \emph{unique} way by adding atoms from $A$,
				and every answer set of $Q$ can be shrinked to one of $P$ by removing atoms that are also in $A$.}
				
				\addProposition{prop:externalAtomInlining}{
					For all \hex-programs $P$, external atoms $e$ in $P$
					and a positive complete family of support sets $\mathcal{S}_{\T}(e, P)$ such that $S^{+}_{\T} \cup \neg S^{-}_{\T} = I(e, P)$ for all $S_{\T} \in \mathcal{S}_{\T}(e, P)$,
					the answer sets of $P$ are equivalent to those of $P_{[e]}$, modulo the atoms newly introduced in $P_{[e]}$.
				}
			
				\addProof{prop:externalAtomInlining}{
					\revmv{
					($\Leftarrow$)
						Let $Y'$ be an answer set of $P_{[e]}$. We show that its restriction $Y$ to ordinary atoms in $P$ is an answer set of $P$.
					
						\begin{itemize}
							\item We first show that $Y$ is a model of $P$.
							It suffices to show that $Y' \models x_e$ iff $\revtwo{Y'}{Y} \models e$.
							\revtwo{}{Since $Y$ and $Y'$ coincide on the input atoms of $e$ (they coincide on all ordinary atoms in $P$),
							we have that $Y \models e$ iff $Y' \models e$, and thus
							it further suffices to show $Y' \models x_e$ iff $Y' \models e$.}

							The if-direction is obvious as the rules in (\ref{def:inlining:1})\revtwo{ together with (\ref{def:inlining:2})}{}
							force $x_e$ to be true whenever $e$ is.
							For the only-if-direction, observe that if $Y' \not\models e$ but $x_e \in Y'$, then
							$Y' \setminus (\{ x_e \} \cup \{ \overline{a} \mid a \in Y' \}) \subsetneq Y'$ is a model of $f P_{[a]}^{Y'}$
							because it does not satisfy any body in (\ref{def:inlining:1}),
							which contradicts the assumption that $Y'$ is an answer set of $P_{[e]}$.
						
							\item Suppose there is a smaller model $Y_{<} \subsetneq Y$ of $f P^Y$.
							We show by case distinction that also $f P_{[e]}^{Y'}$ has a smaller model than $Y'$.
						
							\begin{enumerate}[(a)]
							\item Case $x_e \in Y'$:
						
							We show that
							$Y_{<}' = Y_{<} \cup \{ \overline{a} \mid a \in I(e, P) \setminus Y_{<} \} \cup \{ \overline{a} \mid a \in I(e, P), Y_{<} \models e \} \cup \{ x_e \mid Y_{<} \models e \}$
							is a model of $f P_{[e]}^{Y'}$ and that $Y_{<}' \subsetneq Y'$. 
							For the rules in (\ref{def:inlining:1}), if $Y_{<} \cup \{ \overline{a} \mid a \in I(e, P) \setminus Y_{<} \}$ satisfies one of their bodies,
							then we have that $Y_{<} \models e$ and we set $x_e$ to true, thus the rules are all satisfied.
							If $Y_{<} \cup \{ \overline{a} \mid a \in I(e, P) \setminus Y_{<} \}$ does not satisfy one of their bodies but $Y_{<}'$ does, then
							the additional atoms in $Y_{<}'$ can only come from $\{ \overline{a} \mid a \in I(e, P), Y_{<} \models e \}$, which implies $Y_{<} \models e$ (by construction)
							and thus $x_e \in Y_{<}$ also in this case. Hence, the rules in (\ref{def:inlining:1}) are all satisfied.
							The construction satisfies also the rules in (\ref{def:inlining:2}) because we set $\overline{a}$ to true whenever $a$ is false or $x_e$ is true in $Y_{<}'$ (due to $Y_{<}' \models e$).
							Rule (\ref{def:inlining:3}) is not in $f P_{[e]}^{Y'}$ because $x_e \in Y'$ by assumption.
							For the rules $P|_{e \rightarrow x_e}$ in (\ref{def:inlining:4}) satisfaction is given because $r \in f P^Y$ iff $r|_{e \rightarrow x_e} \in f P_{[e]}^{Y'}$ (since $Y \models e$ iff $Y' \models x_e$),
							and by construction of $Y_{<}'$, we set
							$x_e$ to true iff $Y_{<}' \models e$.

							Now suppose $Y_{<}' \not\subsetneq Y'$. We have that $Y_{<} \subsetneq Y \subseteq Y'$ and that $Y' \setminus Y$
							contains \revtwo{no atoms from $\{ \overline{a} \mid a \in I(e, P) \} \cup \{ x_e, \overline{x_e} \}$.}{only atoms from
							$S = \{ \overline{a} \mid a \in I(e, P) \} \cup \{ x_e, \overline{x_e} \}$,
							and therefore $Y' \setminus Y_{<}$ contains some atom not in $S$. But then}
							\revtwo{Therefore, $Y_{<}' \not\subseteq Y'$ implies that}{$Y_{<}' \not\subsetneq Y'$ is only possible if} $Y_{<}'$ adds an atom from \revtwo{this set}{$S$}
							to $Y_{<}$\rev{, which}{ that} is not in $Y'$,
							i.e., \revtwo{$Y' \setminus Y_{<}'$}{$Y_{<}' \setminus Y'$} contains an atom from \revtwo{$\{ \overline{a} \mid a \in I(e, P) \} \cup \{ x_e, \overline{x_e} \}$}{$S$}.
							But this is impossible since $x_e \in Y'$, thus we also have $\overline{a} \in Y'$ for all $a \in I(e, P)$,
							while $\overline{x_e} \not\in Y_{<}'$ by construction.

							Moreover, $Y_{<}' \subsetneq Y'$ because they differ in an atom other than
							$\{ \overline{a} \mid a \in I(e, P) \} \cup \{ x_e, \overline{x_e} \}$ due to $Y_{<} \subsetneq Y$.
						
							\item Case $\overline{x_e} \in Y'$:

							We show that $Y_{<}' = Y_{<} \cup \{ \overline{a} \mid a \in I(e, P) \setminus Y' \} \cup \{ \overline{x_e} \}$
							is a model of $f P_{[e]}^{Y'}$ and that $Y_{<}' \subsetneq Y'$.

							The rules in (\ref{def:inlining:1}) are all eliminated from $f P_{[e]}^{Y'}$ because
							$x_e \not\in Y'$ implies $Y' \not\models e$ and thus $Y'$ does not satisfy any body of the rules in (\ref{def:inlining:1});
							this is because due to minimality of $Y'$ and falsehood of $x_e$, no $\overline{a}$ is set to true if $a$ is already true in $Y'$.
							The rules in (\ref{def:inlining:2}) are satisfied because for every $a \in I(e, P)$ we have that either
							$\overline{a} \leftarrow \naf a$ is not in $f P_{[e]}^{Y'}$ (if $a \in Y'$) or $\overline{a} \in Y_{<}'$
							by construction (each such $\overline{a}$ is also in $Y'$).
							We further have $\overline{x_e} \in Y_{<}'$ by construction and thus rule (\ref{def:inlining:3}) is satisfied.
							For the rules $r' \in f (P|_{e \rightarrow x_e})^{Y'}$ in (\ref{def:inlining:4}),
							observe that there are corresponding rules $r \in f P^Y$, and that $Y_{<}'$ coincides with $Y_{<}$
							on atoms other than $x_e$.
							If $Y_{<} \models r$ because $Y_{<} \models H(r)$ or $Y_{<} \not\models B(r) \setminus \{ e \}$, then this implies $Y_{<}' \models r'$.
							If $Y_{<} \models r$ because $Y_{<} \not\models e$, then $Y_{<}' \models r'$ because
							$Y_{<}' \not\models x_e$ by construction of $Y_{<}'$.

							Moreover, $Y_{<}' \subsetneq Y'$ because they differ in an atom other than
							$\{ \overline{a} \mid a \in I(e, P) \} \cup \{ x_e, \overline{x_e} \}$ due to $Y_{<} \subsetneq Y$.
							\end{enumerate}
						\end{itemize}

					($\Rightarrow$)
						Let $Y$ be an answer set of $P$. We show that
						\begin{align*}
							Y' = \ Y & \cup \{ \overline{a} \mid a \in I(e, P) \setminus Y \} \cup \{ \overline{a} \mid a \in I(e, P), Y \models e \} \\
									& \cup \{ x_e \mid Y \models e \} \cup \{ \overline{x_e} \mid Y \not\models e \}
						\end{align*}
						is an answer set of $P_{[e]}$\reva{; afterwards we show that $Y'$ is actually the only extension of $Y$ to an answer set of $P_{[e]}$}.

						\begin{itemize}
							\item We first show that $Y'$ is a model of $P_{[e]}$.
							If $Y \cup \{ \overline{a} \mid a \in I(e, P) \setminus Y \}$ satisfies one of the rule bodies in (\ref{def:inlining:1}),
							then $\revtwo{S_{\T}}{S_{\T}^{+}} \subseteq Y$ and $(I(e, P) \setminus S^{-}_{\T}) \cap Y = \emptyset$
							(if some $a \in I(e, P) \setminus \revtwo{S_{\T}}{S_{\T}^{-}}$ would be in $Y$,
							then $\overline{a}$ would not be in $Y \cup \{ \overline{a} \mid a \in I(e, P) \setminus Y \}$ and the rule body would not be satisfied)
							for some $\revtwo{S^{+}_{\T}}{S_{\T}} \in \mathcal{S}^{-}_{\T}(e, P)$;
							this implies $Y \models e$ and, by construction, $x_e \in Y'$.
							If only $Y'$ but not $Y \cup \{ \overline{a} \mid a \in I(e, P) \setminus Y \}$ satisfies one of the rule bodies in (\ref{def:inlining:1}),
							then additional atoms of kind $\overline{a}$ must be in $Y'$, which are only added if $Y \models e$; this also implies,
							by construction, $x_e \in Y'$.
							Thus we have $x_e \in Y'$ whenever $Y'$ satisfies one of the rule bodies in (\ref{def:inlining:1}),
							and thus these rules are all satisfied.
							We further add $\overline{a}$ whenever $a \not\in Y$ or $x_e$ is added (due to $Y \models e$)
							for all $a \in I(e, P)$, which satisfies rules (\ref{def:inlining:2}), and we add $\overline{x_e}$
							whenever $x_e$ is not added (due to $Y \not\models e$), thus the rule (\ref{def:inlining:3}) is satisfied.
							Moreover, the rules in (\ref{def:inlining:4}) are satisfied because $Y$ is a model of $P$ and the value of
							$x_e$ under $Y'$ coincides with the value of $e$ under $Y$ by construction.
						
							\item Suppose there is a smaller model $Y_{<}' \subsetneq Y'$ of $f P_{[e]}^{Y'}$ and assume that this $Y_{<}'$ is subset-minimal.
							We show that then, for the restriction $Y_{<}$ of $Y_{<}'$ to the atoms in $P$ it holds that
							(i) $Y_{<}$ is a model of $f P^Y$ and
							(ii) $Y_{<} \subsetneq Y$,
							which contradicts the assumption that $Y$ is an answer set of $P$.

							(i) Suppose there is a rule $r \in f P^Y$ such that $Y_{<} \not\models r$.
							Observe that for $r' = r|_{e \rightarrow x_e}$ we have $r' \in f P_{[e]}^{Y'}$ because we have
							$Y \models B(r)$ (since $r \in fP^Y$) and
							$Y' \models x_e$ iff $Y \models e$ (by construction of $Y'$), which implies $Y' \models B(r')$.
							Moreover, since $Y_{<}' \models r'$, we either have $Y_{<}' \models H(r')$ or $Y_{<}' \not\models B(r')$.
							In the former case we also have $Y_{<} \models H(r)$, and thus $Y_{<} \models r$, because the two assignments resp.~rules
							coincide on ordinary atoms in $P$; with the same argument $Y_{<} \models r$ holds also in the latter case if a
							body atom in $B(r') \setminus \{ x_e \}$ is unsatisfied under $Y_{<}'$.
							Hence, $Y_{<}' \models r'$ and $Y_{<} \not\models r$ is only possible if $e \in B(r)$, $Y_{<}' \not\models x_e$,
							and $Y_{<} \models e$; the latter implies $Y_{<}' \models e$ as $Y_{<}$ and $Y_{<}'$ coincide on $I(e, P)$.
							Moreover, $Y' \models B(r')$ implies $Y' \models x_e$; by construction of $Y'$ this further
							implies $\overline{x_e} \not\in Y'$.

							Since $x_e$ could not be false in $Y_{<}'$ if (at least) one of the rules $r_1, \ldots, r_n$ in (\ref{def:inlining:1})
							would be in $P_{[e]}^{Y'}$ and had a satisfied body, for each $r_i$, $1 \le i \le n$, one of
							$Y' \not\models B(r_i)$ (then $r_i$ is not even in $P_{[e]}^{Y'}$) or
							$Y_{<}' \not\models B(r_i)$ must hold;
							but since $Y_{<}' \subsetneq Y'$ and $B(r_i)$ consists only of positive atoms, the former case in fact implies the latter,
							thus $Y_{<}' \not\models B(r_i)$ must hold for all $1 \le i \le n$.

							Moreover, we have that $\overline{a} \in Y_{<}'$ whenever $a \not\in Y_{<}'$ for all $a \in I(e, P)$.
							This is because $\overline{x_e} \not\in Y'$ and thus $a \vee \overline{a} \leftarrow \naf \overline{x_e} \in P_{[e]}^{Y'}$
							for all $a \in I(e, P)$ (cf.~rules (\ref{def:inlining:2})) and $\overline{x_e} \not\in Y_{<}'$;
							$a \not\in Y_{<}'$ and $\overline{a} \not\in Y_{<}'$ would violate this rule.
							But then $Y_{<}'$ does not fulfill any of the cases in which $e$ is true, hence $Y_{<}' \not\models e$,
							which contradicts our previous observation, thus the initial assumption that $Y_{<} \not\models r$
							is false and we have $Y_{<} \models P^Y$.

							(ii) Finally, we show that $Y_{<} \subsetneq Y$,
							i.e., $Y' \setminus Y_{<}'$ contains not only atoms from $\{ \overline{a} \mid a \in I(e, P) \} \cup \{ x_e, \overline{x_e} \}$.
							We first consider the case $\overline{x_e} \in Y'$.
							Then $Y' \setminus Y_{<}'$ cannot contain
							$x_e$ (because it is not even in $Y'$ by construction)
							or $\overline{x_e}$
							(because it would leave rule (\ref{def:inlining:3}) unsatisfied).
							It further cannot contain any $\overline{a}$ because $Y_{<}'$ is assumed to be subset-minimal and thus contains $\overline{a}$ only if
							$a \not\in Y'$ (and thus $a \not\in Y_{<}'$);
							this is because $x_e \not\in Y'$ and thus all rules in (\ref{def:inlining:2})
							which force $\overline{a}$ to be true, except $\overline{a} \leftarrow \naf a$, are dropped from $f P_{[e]}^{Y'}$;
							but then removal of any $\overline{a}$ would leave the rule $\overline{a} \leftarrow \naf a$ in
							(\ref{def:inlining:2}), which is contained in $f P_{[e]}^{Y'}$, unsatisfied.

							In case $x_e \in Y'$, if $Y' \setminus Y_{<}'$ contains only
							atoms from $\{ \overline{a} \mid a \in I(e, P) \} \cup \{ x_e, \overline{x_e} \}$,
							then it contains $x_e$ (because $\overline{x_e} \not\in Y'$
							and all $\overline{a}$ for $a \in I(e, P)$ must be true whenever $x_e$ is due to the rules in (\ref{def:inlining:2}),
							which are all also in $P_{[e]}^{Y'}$).
							Moreover, we have that $\overline{a} \in Y_{<}'$ whenever $a \not\in Y_{<}'$
							because $a \vee \overline{a} \leftarrow \naf \overline{x_e} \in f P_{[e]}^{Y'}$
							for all $a \in I(e, P)$ (cf.~rules in (\ref{def:inlining:2}));
							$a \not\in Y_{<}'$ and $\overline{a} \not\in Y_{<}'$ would violate this rule.
							But then $Y_{<}'$ does not fulfill any of the cases in which $e$ is true
							(otherwise $Y_{<}'$ would satisfy a body of (\ref{def:inlining:1}),
							which would also be satisfied by $Y' \supsetneq Y_{<}'$, such that the rule would be in
							$P_{[e]}^{Y'}$ and $x_e$ could not be false in $Y_{<}'$), hence $Y_{<}' \not\models e$;
							since $x_e \in Y'$ implies that $Y' \models e$ we have that $Y' \setminus Y_{<}'$ must contain
							at least one of $I(e, P)$ such that the truth values of $e$ can differ under the two assignments,
							thus it does not only contain atoms from $\{ \overline{a} \mid a \in I(e, P) \} \cup \{ x_e, \overline{x_e} \}$.
						\end{itemize}
					}
						
						\reva{It remains to show that $Y'$ is the \emph{only} extension of $Y$ that is an answer set of $P_{[e]}$.
							To this end, consider an arbitrary answer set $Y''$ of $P_{[e]}$ which coincides with $Y'$ on the atoms in $P$ (i.e., they are both extensions of $Y$);
							we have to show that $Y'' = Y'$.
							Since the only rules in the encoding which support $x_e$ are the rules in (\ref{def:inlining:1}),
							minimality of answer sets implies that $Y' \models e$ iff $Y' \models x_e$ and $Y'' \models e$ iff $Y'' \models x_e$;
							since the value of $e$ depends only on atoms in $P$ and is thus the same under $Y'$ and $Y''$, this further implies $Y' \models x_e$ iff $Y'' \models x_e$,
							i.e., the value of $x_e$ under $Y'$ and $Y''$ is the same.
							\revtwo{This further implies that}{Then} the value of $\overline{x_e}$, which is only defined in rule (\ref{def:inlining:3}),
							is also the same in $Y'$ and $Y''$.
							Finally, since $Y'$ and $Y''$ coincide on each atom $a$ in $P$, and the value of $\overline{a}$,
							which is defined only in rules (\ref{def:inlining:2}),
							depends only on atoms which have already been shown to be the same under $Y'$ and $Y''$,
							we have that also the value of $\overline{a}$ is the same under $Y'$ and $Y''$.
							Thus $Y' = Y''$.
							}
				}

				\rev{However,}{Next we show that} the idea still works for arbitrary complete positive families of support sets $\mathcal{S}_{\T}(e, P)$.
				To this end, we first show that two rules $x_e \leftarrow B, b$ and $x_e \leftarrow B, \overline{b}$ in the above encoding,
				stemming from two support sets that differ only in $b$ resp.~$\overline{b}$,
				can be replaced by a single rule $x_e \leftarrow B$ without affecting the semantics of the program.
				Intuitively, this corresponds to the case \rev{that}{where} two support sets
				\reva{$\{ a \in B \} \cup \{ \neg a \mid \revtwo{a \naf}{\overline{a}} \in B \} \cup \{ b \}$ and $\{ a \in B \} \cup \{ \neg a \mid \revtwo{a \naf}{\overline{a}} \in B \} \cup \{ \neg b \}$}
				imply that $e$ is true whenever all of $B$ and one of $b$ or $\overline{b}$
				hold, which might be also be expressed by a single support set\revam{s} $\{ a \in B \} \cup \{ \neg a \mid \revtwo{a \naf}{\overline{a}} \in B \}$%
				\rev{consisting of $B$ only}{ that expresses that $B$ suffices as a precondition;} \reva{\revtwo{the}{this} idea is similarly to resolution}.

				\addProposition{prop:externalAtomInliningOptimization}{
					Let $X$ be a set of atoms and $P$ be a \hex-program such that
					\begin{align*}
						P \supseteq & \ \{ r_1\colon x_e \leftarrow B, b; r_2\colon x_e \leftarrow B, \overline{b} \} \\
								\cup & \ \{ \overline{a} \leftarrow \naf a; \ \overline{a} \leftarrow x_e; \ a \vee \overline{a} \leftarrow \naf \overline{x_e} \mid a \in X \} \\
								\cup & \ \{ \overline{x_e} \leftarrow \naf x_e \}
					\end{align*}
					where $B \subseteq \{ a, \overline{a} \mid a \in X \}$, $b \in X$, and $\overline{x_e}$ occurs only in the rules explicitly shown above.
					Then $P$ is equivalent to $P' = (P \setminus \{ r_1, r_2 \}) \cup \{ r\colon x_e \leftarrow B \}$.
				}
			
				\addProof{prop:externalAtomInliningOptimization}{
					We have to show that an assignment $Y$ is an answer set of $P$ iff it is an answer set of $P'$.
					It suffices to restrict the discussion to $r_1, r_2 \in P$ and the corresponding rule $r \in P'$
					because the other rules in $P$ vs.~$P'$ and their reducts $P^Y$ vs.~$P'^Y$ wrt.~a fixed assignment $Y$ coincide.

					$(\Rightarrow)$
						Let $Y$ be an answer set of $P$.
						We first show that $Y \models P'$. It suffices show that $Y \models r$.
						Towards a contradiction, suppose $Y \not\models x_e$ and $Y \models B$.
						Since we have (at least) one of $b \in Y$ or $\overline{b} \in Y$ (otherwise $Y$ could not satisfy the rule $\overline{b} \leftarrow \naf b \in P$), we also have $Y \not\models r_1$ or $Y \not\models r_2$,
						which is impossible because $Y$ is an answer set of $P$.

						Thus $Y \models P'$.
						Towards a contradiction, suppose there is a smaller model $Y_{<} \subsetneq Y$ of $f P'^{Y}$.
						If $r \not\in f P'^{Y}$ then $Y \not\models B(r)$, which implies that $Y \not\models B(r_1)$ and $Y \not\models B(r_2)$, and thus neither $r_1$ nor $r_2$ is in $f P^Y$.
						Otherwise, since $Y_{<} \models f P'^{Y}$ we have $Y_{<} \models r$ and thus either $Y_{<} \models x_e$ or $Y_{<} \not\models B$.
						But in both cases also $Y_{<} \models r_1$ and $Y_{<} \models r_2$, thus $Y_{<} \models P^Y$, which contradicts the assumption that $Y$ is an answer set of $P$.
					
					$(\Leftarrow)$
						Let $Y$ be an answer set of $P'$.
						We immediately get $Y \models P$ because $Y \models r$ and $r_1$ and $r_2$ are even easier to satisfy than $r$.
					
						Towards a contradiction, suppose there is a smaller model $Y_{<} \subsetneq Y$ of $f P^Y$.
						If $Y \not\models B$ we have that $r \not\in f P'^Y$ and thus $Y_{<} \models f P'^Y$,
						which contradicts the assumption that $Y$ is an answer set of $P'$.
					
						Then $Y \models B$ and we have that $r \in f P'^Y$ and have to show that $Y_{<} \models r$.

						If $Y \models \overline{x_e}$ then $Y \not\models x_e$
						(otherwise $Y \setminus \{ \overline{x_e} \} \models f P'^Y$, contradicting our assumption that $Y$ is an answer set of $P'$).
						Moreover, due to the rule $\overline{b} \leftarrow \naf b$, one of $b$ or $\overline{b}$ must be true in $Y$.
						But then $Y$ cannot not satisfy both $r_1$ and $r_2$, hence $Y \not\models \overline{x_e}$.
					
						Then $Y \models x_e$ (since $Y \models \overline{x_e} \leftarrow \naf x_e$).
						If $Y_{<} \models x_e$ or $Y_{<} \not\models B$ then also $Y_{<} \models r$ and we are done.
						Otherwise, since $Y_{<} \models f P^Y$,
						we have (i) either $Y \not\models b$ (thus $r_1 \not\in f P^Y$) or $Y_{<} \not\models b$ (thus $Y_{<} \models r_1$), where the former case implies the latter since $Y_{<} \subsetneq Y$,
						and (ii) either $Y \not\models \overline{b}$ (thus $r_2 \not\in f P^Y$) or $Y_{<} \not\models \overline{b}$ (thus $Y_{<} \models r_2$), where the former case implies the latter since $Y_{<} \subsetneq Y$.
						Thus, we have in any case both $Y_{<} \not\models b$ and $Y_{<} \not\models \overline{b}$.
						But then $Y_{<} \not\models b \vee \overline{b} \leftarrow \naf \overline{x_e}$, and since this rule is in $f P^Y$ because $Y \not\models \overline{x_e}$, we get $Y_{<} \not\models f P^Y$.
						This contradicts our initial assumption that $f P^Y$ has a smaller model than $Y$, hence $Y$ is an answer set.
				}

				\rev{This idea can be iterated such that pairs of support sets can be recursively combined.
				Based on this result, one can then show that inlining is correct for arbitrary complete families of support sets.
				}{
				The idea of the next corollary is then as follows.
				Suppose we start with a rewriting based on a
				positive complete family of support sets $\mathcal{S}_{\T}(e, P)$ such that $S^{+}_{\T} \cup \neg S^{-}_{\T} = I(e, P)$ for all $S_{\T} \in \mathcal{S}_{\T}(e, P)$.
				We know by Proposition~\ref{prop:externalAtomInlining} that this rewriting is sound an complete.
				Any other positive complete family of support sets can be constructed by iteratively combining support sets in $\mathcal{S}_{\T}(e, P)$
				which differ only in the polarity of a single atom.
				Since the likewise combination of the respective rules in the rewriting does not change
				the semantics of the resulting program as shown by Proposition~\ref{prop:externalAtomInliningOptimization},
				the rewriting can be constructed from an arbitrary positive complete family of support sets right from the beginning.
				}

				\addCorollary{cor:externalAtomInlining}{
					For all \hex-programs $P$, external atoms $e$ in $P$
					and a positive complete family of support sets $\mathcal{S}_{\T}(e, P)$,
					the answer sets of $P$ are equivalent to those of $P_{[e]}$, modulo the atoms newly introduced in program $P_{[e]}$.
				}
			
				\addProof{cor:externalAtomInlining}{
					\rev{
					First apply Proposition~\ref{prop:externalAtomInliningOptimization} and then Proposition~\ref{prop:externalAtomInlining}.
					}{
					A support set of kind $S_{\T}$ with $S^{+}_{\T} \cup \neg S^{-}_{\T} \subsetneq I(e, P)$
					is equivalent to the set $\mathcal{C} = \{ S^{+}_{\T} \cup S^{-}_{\T} \cup R \mid R \subseteq U \cup \neg U, R \text{ consistent} \}$
					of support sets, where $U = I(e, P) \setminus (S^{+}_{\T} \cup \neg S^{-}_{\T})$,
					in the sense that $S_{\T}$ is applicable if one of $\mathcal{C}$ is applicable.
					Conversely, each such support set can be retrieved by recursive resolution-like
					replacement of support sets in $\mathcal{C}$ which differ only in the polarity of a single atom.
					According to Proposition~\ref{prop:externalAtomInliningOptimization},
					such a replacement in $\mathcal{S}_{\T}(e, P)$ does not change the semantics of
					the program $P_{[e]}$ constructed based on $\mathcal{S}_{\T}(e, P)$.
					Thus the encoding can be constructed from an arbitrary positive complete family of support sets right from the beginning.
					}
				}

				We demonstrate the rewriting with an example.
			
				\begin{example}
					Consider $P = \{ a \leftarrow \ext{\mathit{aOrNotB}}{a,b}{} \}$, where $e = \ext{\mathit{aOrNotB}}{a,b}{}$ evaluates to true if $a$ is true or $b$ is false.
					Let $\mathcal{S}_{\T}(e, P) = \{ \{ a \}, \{ \neg b \} \}$.
					Then we have:
					\begin{align*}
						P_{[e]} = \{ & x_e \leftarrow a; \ x_e \leftarrow \overline{b} \\
									& \overline{a} \leftarrow \naf a; \ \overline{a} \leftarrow x_e; \ \overline{b} \leftarrow \naf b; \ \overline{b} \leftarrow x_e; a \vee \overline{a} \leftarrow \naf \overline{x_e}; \ b \vee \overline{b} \leftarrow \naf \overline{x_e} \\
									& \overline{x_e} \leftarrow \naf x_e \\
									& a \leftarrow x_e \}
					\end{align*}
					The program has the unique answer set $Y' = \{ a, x_e, \overline{a}, \overline{b} \}$, which represents the answer set $Y = \{ a \}$ of $P$.
				\end{example}

				Multiple external atoms can be inlined by iterative application.
				For a program $P$ and a set $E$ of external atoms in $P$ we denote by $P_{[E]}$ the program after all external atoms from $E$ have been inlined.
				Importantly, separate auxiliaries must be introduced for atoms that are input to multiple external atoms.

			\subsubsection{Inlining Negated External Atoms}
			\label{sec:inlining:encoding:negative}

				Until now we restricted the discussion to positive external atoms
				based on positive support sets.
				One can observe that the rewriting from Definition~\ref{def:inlining} does indeed not work
				for external atoms $e$ \rev{, which}{that} occur (also) in form $\naf e$ because programs $P$ and $P{[e]}$ are in this case \emph{not} equivalent in general.
			
				\begin{example}
					\label{ex:negatedExternalAtom}
					Consider $P = \{ p \leftarrow \naf \ext{\mathit{neg}}{p}{} \}$, where $\ext{\mathit{neg}}{p}{}$ is true if $p$ is false and vice versa.
					The only answer set of $P$ is $Y = \emptyset$ but
					the rewriting from Definition~\ref{def:inlining} yields
					\begin{align*}
						P_{[\ext{\mathit{neg}}{p}{}]} = \{ & x_e \leftarrow \overline{p} \\
										&  \overline{p} \leftarrow \naf p; \ \overline{p} \leftarrow x_e; \ p \vee \overline{p} \leftarrow \naf \overline{x_e} \\
								& \overline{x_e} \leftarrow \naf x_e \\
								& p \leftarrow \naf x_e \} 
					\end{align*}
					which has the answer sets $Y_1' = \{ x_e, \overline{p} \}$ and $Y_2' = \{ \overline{x_e}, p \}$
					that represent the assignments $Y_1 = \emptyset$ and $Y_2 = \{ p \}$ over $P$.
					However, only $Y_1$ $(=Y)$ is an answer set of $P$.
				\end{example}
			
				Intuitively, the rewriting does not work for negated external atoms
				because their input atoms may support themselves.
				More precisely, due to rule~(\ref{def:inlining:3}), an external atom is false by default if none of the rules~(\ref{def:inlining:1})
				apply. If one of the external atom's input atoms depends on falsehood of the external atom, as in Example~\ref{ex:negatedExternalAtom},
				then the input atom might be supported by falsehood of the external atom, although this falsehood itself depends on the input atom.

				In order to extend our approach to the inlining of negated external atoms $\naf e$ in a program $P$,
				we make use of an arbitrary but fixed negative complete family $\mathcal{S}_{\F}(e, P)$ of support sets as by Definition~\ref{def:supportsetfamily}.
				The idea is \reva{then} to replace a negated external atom $\naf e$ by a positive one $e'$ \rev{such}{that is defined \revtwo{in such}{such}} that $Y \models e'$ iff $Y \not\models e$ for all assignments $Y$; obviously, the resulting program has the same answer sets as before.
				\reva{This reduces the case for negated external atoms to the case for positive ones.}
				\rev{Then t}{T}he semantics of $e'$ is fully described by the negative complete family of support sets of $e$ and we may apply the rewriting of Definition~\ref{def:inlining}.

				The idea is formalized by the following definition:
				\begin{definition}[Negated External Atom Inlining]
					\label{def:neginlining}
					For a \hex-program $P$ and negated external atom $\naf e$ in $P$, let
					\begin{flalign}
						P_{[\naf e]} \hspace{-0.8mm}=\hspace{-0.8mm} & \ \{x_e \leftarrow S_{\F}^{+} \cup \{ \overline{a} \mid \neg a \in S_{\F}^{-} \} \hspace{-0.8mm}\mid\hspace{-0.8mm} S_{\F} \in \mathcal{S}_{\F}(e, P) \} \hspace{-4mm} & \label{def:neginlining:1} \\
												\cup & \ \{ \overline{a} \leftarrow \naf a; \overline{a} \leftarrow x_e; a \vee \overline{a} \leftarrow \naf \overline{x_e} \mid a \in I(e, P) \} & \label{def:neginlining:2} \\
												\cup & \ \{ \overline{x_e} \leftarrow \naf x_e \} & \label{def:neginlining:3} \\
												\cup & \ P|_{\naf e \rightarrow x_e} & \label{def:neginlining:4}
					\end{flalign}
					where $\overline{a}$ is a new atom for each $a$, $x_e$ and $\overline{x_e}$ are new atoms for \revtwo{}{external atom} $e$,
					and $P|_{\naf e \rightarrow x_e} = \bigcup_{r \in P} r|_{\naf e \rightarrow x_e}$
					where $r|_{\naf e \rightarrow x_e}$ denotes rule $r$ with every occurrence of $\naf e$ replaced by $x_e$.
				\end{definition}

				\reva{Informally, the effects of changing a negated external atom to a positive one and using a negative family of support sets
				cancel each other out.}
				One can show that this rewriting is sound and complete.

				\addProposition{prop:negExternalAtomInlining}{
					For all \hex-programs $P$, negated external atoms $\naf e$ in $P$
					and a negative complete family of support sets $\mathcal{S}_{\F}(e, P)$,
					the answer sets of $P$ are equivalent to those of $P_{[\naf e]}$, modulo the atoms newly introduced in program $P_{[\naf e]}$.
				}
			
				\addProof{prop:negExternalAtomInlining}{
					Using a negative complete family of support sets for defining the auxiliary variable $x_e$ in the rules~(\ref{def:neginlining:1}),
					and replacing $\naf e$ by $x_e$ amounts to the
					replacement of $\naf e$ by a new external atom $e'$, \rev{any}{and} applying the rewriting from Definition~\ref{def:inlining} afterwards.
				}

				As before, iterative application allows for inlining multiple negated external atoms.
				In the following, for a program $P$ and a set $E$ of either positive or negated external atoms in $P$,
				we denote by $P_{[E]}$ the program after all external atoms from $E$ have been inlined.
				
				\leanparagraph{Transforming Complete Families of Support Sets.}
				For the sake of completeness we show that one can change the polarity of complete families of support sets:

				\addProposition{prop:supportSetPolarityConversion}{
					Let $\mathcal{S}_{\sigma}$ be a positive resp.~negative complete family of support sets for some external atom $e$ in a program $P$, where $\sigma \in \{ \T, \F \}$.
					Then $\mathcal{S}_{\overline{\sigma}} = \{ S_{\overline{\sigma}} \in \prod_{S_{\sigma} \in \mathcal{S}_{\sigma}} \neg S_{\sigma} \mid S_{\overline{\sigma}} \text{ is consistent} \}$
					is a negative resp.~positive complete family of support sets, where $\overline{\T} = \F$ and $\overline{\F} = \T$.
				}
			
				\addProof{prop:supportSetPolarityConversion}{
					We restrict the proof to the case $\sigma = \T$; the case $\sigma = \F$ is symmetric.
				
					If $\mathcal{S}_{\T}$ is a positive complete family of support sets, then
					the support sets $S_{\T} \in \mathcal{S}_{\T}$ describe the possibilities to satisfy $e$ exhaustively.
					Thus, in order to falsify $e$, at least one literal of each $S_{\T} \in \mathcal{S}_{\T}$ must be falsified,
					i.e., at least one literal in $\neg S_{\T}$ must be satisfied. Thus amounts to the Cartesian product of all sets $\neg S_{\T}$ with $S_{\T} \in \mathcal{S}_{\T}$.
				}

				Intuitively, since a complete family family of positive support sets $\mathcal{S}_{\T}$ fully describes under
				which conditions the external atom is true, one can construct a negative support set by picking an arbitrary literal
				from each $S_{\T} \in \mathcal{S}_{\T}$ and changing its sign. Then, whenever the newly generated set is contained
				in the assignment, none of the original support sets in $\mathcal{S}_{\T}$ can match. The case for families of negative support sets is symmetric.
				
				However, similarly to the transformation of the formula from conjunctive normal form to disjunctive normal form or vice versa,
				this may result in an exponential blow-up.
				In the spirit of our initial assumption that compact complete families of support sets exist,
				it is suggested to construct families of support sets of the required polarity right from the beginning,
				which we will also do in our experiments.

	\section{Exploiting External Source Inlining for Performance Boosts}
	\label{sec:implementation}

		An application of the techniques from the previous section are algorithmic improvements by skipping explicit verification calls
		for the sake of performance gains.
		As stated in Section~\ref{sec:preliminaries}, learning techniques may reduce the number of required verification calls,
		and -- alternatively -- using support sets for verification instead of explicit calls
		may lead to an efficiency improvement when checking external source guesses,
		but neither of these techniques eliminates the checks altogether~\cite{eiterFR014}.
		In contrast, inlining embeds the semantics of external sources directly in the logic program.
		Thus, no more checks are needed; the resulting program can actually be evaluated by an ordinary ASP solver.
		
		\subsection{Implementation}

			We implemented this approach in the \dlvhex{}\footnote{\url{www.kr.tuwien.ac.at/research/systems/dlvhex}} system, which
			is based on \gringo{} and \clasp{} from the Potassco suite\footnote{\url{https://potassco.org}}.
			External sources are supposed to provide a complete set of support sets.
			The system allows also for using universally quantified variables in the specification of support sets,
			which are automatically substituted by all constants occurring in the program.
			\rev{After support set learning in the initialization for the sake of}{After} external source inlining \reva{during preprocessing},
			the \hex-program is \revam{then }evaluated entirely by the backend without any external calls.

			The rewriting makes both the compatibility check (cf.~Definition~\ref{def:compatibleset})
			and the minimality check wrt.~the reduct and external sources (cf.~Section~\ref{sec:preliminaries} and \citeN{efkrs2014-jair}) obsolete.
			With the traditional approach, compatible sets are not necessarily answer sets\rev{ because
			of possibly self-justified atoms whose cyclic support
			involves external atoms
			and is not detected by the ordinary ASP solver that evaluates the guessing program $\hat{P}$.
			On the other hand, after inlining, the minimality check performed by the ordinary ASP solver suffices.}{. This is
			because cyclic support of atoms that involves external sources is not detected by the ordinary ASP solver when evaluating $\hat{P}$.
			But after inlining, due to soundness and completeness of our rewriting, the minimality check performed by the ordinary ASP solver suffices.
			}

			We evaluated the approach using the experiments described in the following.
			
		\subsection{Experimental Setup}

			We present several benchmarks with $100$ randomly generated instances each, which
			were run on a Linux server with two 12-core AMD 6176 SE CPUs and 128GB RAM
			using a $300$ seconds timeout.
			The instances are available from \url{http://www.kr.tuwien.ac.at/research/projects/inthex/inlining},
			while the program encodings and scripts used for running the benchmarks are included in the sourcecode repository of the \dlvhex{} system,
			which is available from \url{https://github.com/hexhex}.
			Although some of the benchmark problems are similar to those used by~\citeN{eiterFR014} and in the conference versions of this paper, the runtime results are not directly comparable
			because of technical improvements in the implementation of support set generation and other (unrelated) solver improvements.
			Moreover, for the taxi benchmark we use a different scenario since the previous one was too easy in this context.
			However, for the pre-existing approaches the fundamental trend that the approach based on support sets outperforms the traditional approach is the same.

			In our tables we compare three evaluation approaches (\emph{configurations}), which we evaluate both for computing all and the first answer set only\rev{; the}{.
			The runtimes specify the wall-clock time needed for the whole reasoning task including grounding, solving and side tasks; the observed runtime differences, however, stem only
			from the solving technique since grounding and other reasoning tasks are the same for all configurations. The}
			numbers in parentheses indicate the number of timeout instances, which were counted as $300$ seconds when computing the average runtime of the
			instances\reva{; otherwise timeout instances could even decrease the average runtime compared to instances which finish shortly before the deadline.\footnote{Due to this
			it might happen in few cases that two configurations behave similar wrt.~runtime but the number of timeout instances is different.
			This is explained by instances which terminate shortly before the deadline with one configuration and do not terminate in time with the other.
			}}
			The \textbf{traditional} evaluation algorithm guesses the truth values of external atoms and verifies them by evaluation.
			In our experiments we use the learning technique EBL~\cite{efkr2012-tplp} to learn parts of the external atom's behavior,
			i.e., there is a tight coupling of the reasoner with external sources.
			The second approach as by~\citeN{eiterFR014} is based on \textbf{support sets} (\textbf{sup.sets}), which are provided by the external source and learned at the beginning
			of the evaluation process.
			It then guesses external atoms as in the traditional approach, but verifies them by matching candidate compatible sets against
			support sets rather than by evaluation.
			While we add learned support sets as nogoods at the beginning, which exclude some but not all wrong guesses,
			recall that on-the-fly learning as by EBL is not \rev{possible}{done} in this approach since external sources are only called at the beginning;
			this may be a drawback compared to \textbf{traditional}.
			The new \textbf{inlining} approach, based on the results from this paper, also learns support sets at the beginning similar to \textbf{sup.sets},
			but uses them for rewriting external atoms as demonstrated in Section~\ref{sec:inlining}.
			Then, all answer sets of the rewritten ASP-program are accepted without the necessity for additional checks.
			Wrong guesses \rev{, which}{that} are not detected by the ordinary ASP solver backend\reva{,} cannot occur here.

			Note that our goal is to show improvements compared to previous \hex-algorithms,
			but not to compare \hex{} to other formalisms or encodings in ordinary (disjunctive) ASP, which might be feasible for some of the benchmark programs.
			\rev{Our fundamental assumption that compact complete families of support sets exist (see Section~\ref{sec:inlining})
			is satisfied for all our scenarios, as discussed in more detail when we present the results.}{
			Compact (i.e., polynomial) complete families of support sets exist for all scenarios considered in the following;
			we make the statement about the sizes more precise \revtwo{}{when we} discuss the individual benchmarks \revtwo{}{below}.}
			
			Our hypothesis is that \textbf{inlining}
			outperforms both \textbf{traditional} and \textbf{sup.sets} for external sources with compact complete support set families.
			More precisely, we expect that \textbf{inlining} leads to a further speedup over \textbf{sup.sets} in many cases,
			especially when there are many candidate answer sets.
			Moreover, we expect that in cases where \textbf{inlining} cannot yield further improvements over \textbf{sup.sets},
			then it does at least not harm much.
			This is because with \textbf{inlining}, (i) no external calls and (ii) no additional minimality checks are needed,
			which potentially leads to speedups.
			On the other hand, the only significant costs when generating the rewriting are caused by support set learning;
			however, this is also necessary with \textbf{sup.sets},
			which was already shown to outperform \textbf{traditional} if small complete families of support sets exist.
			Hence, we expect further benefits but negligible additional costs.

			\leanparagraph{House Problem.}
			We first consider an abstraction of configuration problems,
			consisting of sets of \emph{cabinets}, \emph{rooms}, \emph{objects} and \emph{persons}~\cite{mbsf2009}.
			The goal is to assign cabinets to persons, cabinets to rooms, and objects to cabinets,
			such that there are no more than four cabinets in a room or more than five objects in a cabinet.
			Objects belonging to a person must be stored in a cabinet belonging to the same person,
			and a room must not contain cabinets of more than one person.
			We assume that we have already a partial assignment to be completed.
			We use an existing guess-and-check encoding%
			\footnote{The encoding was taken from
			{\footnotesize\url{http://143.205.174.183/reconcile/tools}}.}
			which implements
			the check as external source.
			Instances of size $n$ have $n$ persons, $n{+}2$ cabinets, $n{+}1$
			rooms, and $2n$ objects
			randomly assigned to persons;
			$2n{-}2$ objects are already stored.
			
			The number and size of support sets is polynomially bounded by $(2n)^5$; this is due to the constraints
			that no more than four cabinets can be in a room and no more than five objects can be in a cabinet.
			
			Table~\ref{tab:house} shows the results\revam{,
			where numbers in parentheses indicate timeout instances}.
			As expected, we have that \textbf{sup.sets} clearly outperforms \textbf{traditional} both when computing all answer sets and the first answer set only,
			which is because of faster candidate checking as already observed by~\citeN{eiterFR014}.
			When computing all answer sets, the new \textbf{inlining} approach leads to a further speedup as it eliminates wrong guesses and the checking step altogether,
			while the additional initialization overhead is negligible. This is consistent with our hypothesis.
			When computing only a single answer set, \textbf{inlining} does not yield a further visible speedup,
			which can be explained by the fact that only few candidates \revam{(in the extreme case for consistent instances: a single one) }must be checked before an answer set is found.
			In this case the additional initialization overhead compared to \textbf{sup.sets} is slightly visible,
			but as can be seen it is little such that the new technique does in fact not harm, as expected.
			

			{
				\begin{table}[t]
					\centering
					\begin{tabular}[t]{r|rrr|rrr}
					\toprule
					$n$ & \multicolumn{3}{c|}{all answer sets} & \multicolumn{3}{c}{first answer set} \\
					& traditional & sup.sets & inlining & traditional & sup.sets & inlining \\
					\midrule
5 & 99.88 ~~(17) & 5.81 ~~(0) & 3.57 ~~(0) & 5.17 ~~(0) & 0.39 (0) & 0.40 (0) \\
6 & 193.56 ~~(35) & 19.40 ~~(1) & 11.51 ~~(0) & 13.03 ~~(0) & 0.75 (0) & 0.77 (0) \\
7 & 252.61 ~~(81) & 35.72 ~~(3) & 22.04 ~~(2) & 23.68 ~~(2) & 1.50 (0) & 1.54 (0) \\
8 & 267.01 ~~(85) & 93.39 (13) & 59.25 (11) & 64.89 (10) & 3.06 (0) & 3.14 (0) \\
9 & 274.23 ~~(85) & 129.37 (29) & 85.85 (13) & 79.52 (13) & 6.15 (0) & 6.34 (0) \\
10 & 281.55 ~~(83) & 154.29 (42) & 120.66 (16) & 107.86 (12) & 11.80 (0) & 12.17 (0) \\
11 & 297.28 ~~(86) & 206.15 (53) & 166.84 (45) & 160.25 (49) & 21.84 (0) & 22.55 (0) \\
12 & 300.00 (100) & 246.40 (57) & 179.59 (41) & 162.33 (47) & 39.31 (0) & 40.62 (0) \\
13 & 297.43 ~~(99) & 281.02 (91) & 239.08 (69) & 214.30 (65) & 68.07 (0) & 70.43 (0) \\
14 & 300.00 (100) & 287.11 (91) & 253.58 (65) & 213.63 (63) & 114.56 (0) & 118.81 (0) \\
15 & 300.00 (100) & 296.36 (92) & 287.66 (75) & 240.21 (75) & 187.94 (0) & 195.09 (0) \\
					\bottomrule
					  \end{tabular}
					\caption{House configuration}
					\label{tab:house}
				\end{table}
			}

			\leanparagraph{Taxi Assignment.}
			We consider a program which uses external atoms to access a $\dllite_{\cA}$-ontology\rev{ (}{, }called
			a \emph{DL-atom}~\cite{eilst2008-aij}\revam{)}.
			As \rev{demonstrated}{discussed} in \rev{Section~\ref{sec:inlining:constructing}}{Section~\ref{sec:preliminaries}},~\citeN{clmr2007} have proven that
			for this type of description logic at most one assertion is needed to derive an instance query from a consistent ontology.
			Moreover, at most two added ABox assertions are needed to make such an ontology inconsistent.
			Hence, the support sets required to describe the ontology
			are of only few different and small forms, which limits also the number of possible support sets to a quadratic number
			in the size of the program and the Abox.
			Moreover, the support sets are easy to construct by a syntactic analysis of the ontology and the DL-atoms, for details we refer to~\citeN{eiterFR014}.

			The task in this benchmark is to assign taxi \emph{drivers} to \emph{customers}.
			Each customer and driver is in a \emph{region}.
			A customer may only be assigned to a driver in the same region.
			Up to four customers may be assigned to a driver.
			We let some customers be \emph{e-customers} who use only electronic cars,
			and some drivers be \emph{e-drivers} who drive electronic cars.
			The ontology stores information about individuals such as their locations (randomly chosen but balanced among regions).
			The encoding is taken from~\url{http://www.kr.tuwien.ac.at/research/projects/inthex/partialevaluation}.
			An instance of size $4 \le n \le 9$ consists of $n$ drivers, $n$ customers including $n/2$ e-customers and $n/2$ regions.

			Table~\ref{tab:taxi} shows the results.
			The \textbf{sup.sets} approach is faster than the \textbf{traditional} one.
			When computing all answer sets, the difference is still clearly visible but less dramatic than when computing only the first answer set or in other benchmarks.
			This is because there is a large number of candidates and answer sets in this benchmark,
			which allow the learning techniques used in \textbf{traditional} to learn the behavior of the external sources well over time.
			The reasoner can then prevent wrong guesses and verification calls effectively,
			such that the advantage of improved verification calls as in \textbf{sup.sets} decreases the longer the solver runs.
			%
			However, the \textbf{inlining} approach leads to a significant speedup
			since wrong guesses are impossible from the beginning
			and all verification calls are spared.
		
			\begin{table}[t]
				\centering
					\begin{tabular}[t]{r|rrr|rrr}
					\toprule
					$n$ & \multicolumn{3}{c|}{all answer sets} & \multicolumn{3}{c}{first answer set} \\
					& traditional & sup.sets & inlining & traditional & sup.sets & inlining \\
					\midrule
4 & 0.54 ~~~~(0) & 0.47 ~~~~(0) & 0.22 ~~~~(0) & 0.19 ~~(0) & 0.16 (0) & 0.16 (0) \\
5 & 5.40 ~~~~(0) & 5.92 ~~~~(0) & 1.10 ~~~~(0) & 0.82 ~~(0) & 0.21 (0) & 0.18 (0) \\
6 & 88.93 ~~~~(9) & 63.24 ~~~~(2) & 8.92 ~~~~(0) & 8.86 ~~(0) & 0.28 (0) & 0.21 (0) \\
7 & 295.94 ~~(98) & 277.64 ~~(84) & 149.56 ~~(19) & 154.71 (42) & 0.90 (0) & 0.26 (0) \\
8 & 300.00 (100) & 299.99 ~~(99) & 290.00 ~~(94) & 249.79 (81) & 3.55 (1) & 0.32 (0) \\
9 & 300.00 (100) & 300.00 (100) & 300.00 (100) & 281.35 (92) & 2.77 (0) & 0.39 (0) \\
10 & 300.00 (100) & 300.00 (100) & 300.00 (100) & 289.54 (96) & 3.33 (1) & 0.49 (0) \\
					\bottomrule
				  \end{tabular}

				\caption{Driver-customer assignment}
				\label{tab:taxi}
			\end{table}

			\leanparagraph{LUBM Diamond.}
			While description logics correspond to fragments of first-order logic and are monotonic,
			their cyclic interaction with rules allow for default reasoning,
			i.e., making assumptions which might have to be withdrawn if more information becomes available (such as classifying an object based on absence of information).
			We consider default reasoning over
			the LUBM $\dllite_{\cA}$ ontology (\url{http://swat.cse.lehigh.edu/projects/lubm/}).
			Defaults express that assistants are normally employees
			and students are normally not employees.
			The ontology entails that assistants are students, resembling Nixon's diamond.
			The instance size is the number of persons \rev{, which}{who}
			are randomly marked as students, assistants or employees.
			The task is to classify all persons in the ontology. Due to incomplete information the result is not unique.

			Table~\ref{tab:lubm} shows the results.
			As already observed by~\citeN{eiterFR014},
			\textbf{sup.sets} outperforms \textbf{traditional}.
			Compared to the taxi benchmark there is a significantly smaller number of model candidates,
			which makes learning in the \textbf{traditional} approach less effective.
			This can in particular be seen when computing all answer sets, since when computing the first answer set only, learning is less effective anyway (as described in the previous benchmark).
			The decreased effectiveness of learning from external calls is then more easily compensated by the more efficient compatibility check
			as by \textbf{sup.sets}, which is why the relative speedup is larger now.
			However, \textbf{inlining} is again the most efficient approach due to elimination of the compatibility check.
			Thanks to the existence of a \rev{compact}{quadratic} family \reva{of} support sets for $\dllite_{\cA}$-ontologies \reva{(see previous benchmark)}, the speedup is dramatic.

			{
				\begin{table}[t]
					\centering
						\begin{tabular}[t]{r|rrr|rrr}
						\toprule
						$n$ & \multicolumn{3}{c|}{all answer sets} & \multicolumn{3}{c}{first answer set} \\
						& traditional & sup.sets & inlining & traditional & sup.sets & inlining \\
						\midrule
20 & 1.17 ~~~~(0) & 0.33 ~~~~(0) & 0.30 (0) & 0.34 ~~~~(0) & 0.31 ~~~~(0) & 0.30 (0) \\
30 & 30.05 ~~~~(3) & 0.98 ~~~~(0) & 0.33 (0) & 6.29 ~~~~(0) & 0.61 ~~~~(0) & 0.33 (0) \\
40 & 148.57 ~~(40) & 16.66 ~~~~(2) & 0.37 (0) & 86.69 ~~(22) & 8.88 ~~~~(0) & 0.37 (0) \\
50 & 250.26 ~~(75) & 80.51 ~~(15) & 0.44 (0) & 214.68 ~~(65) & 51.94 ~~~~(4) & 0.43 (0) \\
60 & 286.58 ~~(89) & 183.79 ~~(47) & 0.52 (0) & 265.91 ~~(87) & 153.05 ~~(36) & 0.52 (0) \\
70 & 297.94 ~~(99) & 253.66 ~~(73) & 0.65 (0) & 297.16 ~~(99) & 225.54 ~~(65) & 0.65 (0) \\
80 & 300.00 (100) & 282.01 ~~(91) & 0.81 (0) & 300.00 (100) & 271.19 ~~(84) & 0.81 (0) \\
90 & 300.00 (100) & 298.71 ~~(99) & 1.04 (0) & 300.00 (100) & 296.06 ~~(97) & 1.04 (0) \\
100 & 300.00 (100) & 300.00 (100) & 1.27 (0) & 300.00 (100) & 298.45 ~~(99) & 1.27 (0) \\
110 & 300.00 (100) & 300.00 (100) & 1.59 (0) & 300.00 (100) & 300.00 (100) & 1.58 (0) \\
120 & 300.00 (100) & 300.00 (100) & 2.00 (0) & 300.00 (100) & 300.00 (100) & 2.00 (0) \\
						\bottomrule
					  \end{tabular}

					\caption{Default rules over LUBM in $\mathit{DL}$-$Lite_{\cA}$}
					\label{tab:lubm}

				\end{table}
			}

			\leanparagraph{Non-3-Colorability.}
			We consider the problem of deciding if a given graph is \emph{not} 3-colorable, i.e., if it is
			not possible to color the nodes such that adjacent nodes have different colors.
			\reva{To make the problem more challenging, we want to represent the answer by a dedicated atom within the program.
			That is, we do not simply want to compute all valid 3-colorings and leave the program inconsistent in case there is no valid 3-coloring,
			but the program should rather be consistent in this case and
			a dedicated atom should represent that there is no 3-coloring; this allows, for instance, continuing reasoning based on the result.}
			
			We use \rev{an}{a saturation} encoding which splits the guessing part $P_{\mathit{col}}$ from the checking part $P_{\mathit{check}}$.
			The latter, which is itself implemented as logic program
			$$\reva{P_{\mathit{check}} = \{ \mathit{inv} \leftarrow \mathit{inp}(\mathit{col}, U, C), \mathit{inp}(\mathit{col}, V, C), \mathit{inp}(\mathit{edge}, U, V) \}}\text{,}$$
			is used as an external source from the guessing part.
			For a color assignment, given by facts of kind $\mathit{inp}(\mathit{col}, v, c)$ where $v$ is a vertex and $c$ is a color,
			$P_{\mathit{check}}$ derives the atom $\mathit{inv}$ in its only answer set, otherwise it has an empty answer set.
			We then use the following program $P_{\mathit{col}}$ to guess a coloring
			and check it using the external atom $\ext{\mathit{query}}{P_{\mathit{check}}, \mathit{inp}, \mathit{inv}}{}$ \revtwo{which is true}{for query answering over subprograms.
			We let $\ext{\mathit{query}}{P_{\mathit{check}}, \mathit{inp}, \mathit{inv}}{}$ evaluate to true}
			iff program $P_{\mathit{check}}$, extended with facts over predicate $\mathit{inp}$, delivers an answer set \rev{which}{that} contains $\mathit{inv}$.\revtwo{}{\footnote{Here,
			the parameter $\mathit{inv} \in \mathcal{P}$ is a predicate symbol, whose purpose is to inform the external source about the propositional atom it should look
			for in the answer sets of the subprogram.}}
			\reva{In this case we saturate the model.}
			We add a constraint \rev{which}{that} eliminates answer sets other than the saturated one,
			thus each instance has either no or exactly one answer set.
			The size of the instances is the number of nodes $n$.

			A compact complete family of support sets for $\ext{\mathit{query}}{P_{\mathit{check}}, \mathit{inp}, \mathit{inv}}{}$ exists:
			the number of edges to be checked is no greater than quadratic in the number of nodes and the number of colors is constant,
			which allows the check to be encoded by a quadratic number of binary support sets.
			
			The encoding is as follows:
			\begin{align*}
				P_{\mathit{col}} = \big\{ \mathit{col}(V, \mathit{r}) \vee \mathit{col}(V, \mathit{g}) \vee \mathit{col}(V, \mathit{b}) \leftarrow\ & \mathit{node}(V) \\[0.3ex]
											\mathit{inp}(\mathit{p}, X, Y) \leftarrow\ & \mathit{p}(X, Y) \mid  p \in\{ \mathit{col}, \mathit{edge}\} \\[0.3ex]
											\mathit{inval} \leftarrow\ & \ext{\mathit{query}}{P_{\mathit{check}}, \mathit{inp}, \mathit{inv}}{} \\[0.3ex]
											\mathit{col}(V,c) \leftarrow\ & \mathit{inval}, \mathit{node}(V) \mid c \in \{r, g, b\} \\[0.3ex]
											\leftarrow\ & \reva{\naf{}} \mathit{inval} \big\}
			\end{align*}

			The results are shown in Table~\ref{tab:graphcoloring}.
			While \textbf{sup.sets} already outperforms
			\textbf{traditional}, \textbf{inlining} leads to a further small speedup
			when computing all answer sets.
			Compared to previous benchmarks, there are significantly fewer support sets,
			which makes candidate checking in \textbf{sup.sets} inexpensive.
			This explains the large speedup of \textbf{sup.sets} over \textbf{traditional},
			and that avoiding the check in \textbf{inlining} does not lead to a large further speedup.
			However, due to a negligible additional overhead, \textbf{inlining} does at least not harm,
			which is in line with our hypothesis.
			
			\revmv{
			Interestingly, the runtimes when computing all and the first answer set only are almost the same.
			Although this effect occurs with all configurations and is not related to our new approach, we briefly discuss it.
			Each instance has either one or no answer set. Despite this, computing all answer sets can in principle
			be slower than computing the first answer set since the reasoner has to determine that there are no further ones.
			However, in this case, the instances terminate almost immediately after the (only) answer set has been found.
			\rev{Since the only answer set is the saturated one, whose minimality wrt.~the reduct has already been established,
			learning techniques in the ordinary ASP solver seem to help the reasoner to determine that there cannot be further (smaller) answer sets.}{
			Since the only answer set of a non-3-colorable instance is the saturated one, which is also the only classical model,
			the reasoner needs to perform only a single minimality check.
			}
			}

			{
				{
				\begin{table}[t]
					\centering
					\rev{
					\begin{tabular}[t]{r|rrr|rrr}
					\toprule
					$n$ & \multicolumn{3}{c|}{all answer sets} & \multicolumn{3}{c}{first answer set} \\
					& traditional & sup.sets & inlining & traditional & sup.sets & inlining \\
					\midrule
20 & 299.01 ~~(99) & 0.20 ~~(0) & 0.16 ~~(0) & 0.12 ~~(0) & 0.12 ~~(0) & 0.12 ~~(0) \\
60 & 300.00 (100) & 1.65 ~~(0) & 1.38 ~~(0) & 0.47 ~~(0) & 0.46 ~~(0) & 0.46 ~~(0) \\
100 & 300.00 (100) & 8.74 ~~(0) & 8.05 ~~(0) & 2.03 ~~(0) & 2.02 ~~(0) & 2.02 ~~(0) \\
140 & 300.00 (100) & 29.65 ~~(0) & 28.62 ~~(0) & 6.42 ~~(0) & 6.46 ~~(0) & 6.46 ~~(0) \\
180 & 300.00 (100) & 76.91 ~~(0) & 75.65 ~~(0) & 16.60 ~~(0) & 16.48 ~~(0) & 16.39 ~~(0) \\
220 & 300.00 (100) & 154.24 (20) & 153.17 (20) & 35.64 ~~(0) & 35.60 ~~(0) & 35.30 ~~(0) \\
260 & 300.00 (100) & 203.91 (49) & 201.88 (50) & 67.63 ~~(0) & 67.20 ~~(0) & 67.48 ~~(0) \\
300 & 300.00 (100) & 233.00 (60) & 231.12 (60) & 117.26 ~~(0) & 117.75 ~~(1) & 117.09 ~~(0) \\
340 & 300.00 (100) & 252.45 (70) & 250.02 (70) & 164.72 (29) & 164.59 (30) & 164.62 (30) \\
380 & 300.00 (100) & 265.30 (80) & 262.85 (80) & 196.86 (41) & 196.76 (40) & 196.46 (40) \\
420 & 300.00 (100) & 276.76 (80) & 274.24 (80) & 219.18 (58) & 219.65 (58) & 219.48 (60) \\
460 & 300.00 (100) & 282.09 (90) & 280.31 (90) & 235.81 (60) & 236.19 (61) & 236.10 (61) \\
					\bottomrule
					  \end{tabular}
					}{
					\begin{tabular}[t]{r|rrr|rrr}
					\toprule
					$n$ & \multicolumn{3}{c|}{all answer sets} & \multicolumn{3}{c}{first answer set} \\
					& traditional & sup.sets & inlining & traditional & sup.sets & inlining \\
					\midrule
20 & 298.94 ~~(99) & 0.19 ~~(0) & 0.16 ~~(0) & 298.96 ~~(99) & 0.19 ~~(0) & 0.16 ~~(0) \\
60 & 300.00 (100) & 1.61 ~~(0) & 1.35 ~~(0) & 300.00 (100) & 1.61 ~~(0) & 1.35 ~~(0) \\
100 & 300.00 (100) & 8.45 ~~(0) & 7.81 ~~(0) & 300.00 (100) & 8.44 ~~(0) & 7.83 ~~(0) \\
140 & 300.00 (100) & 28.18 ~~(0) & 27.30 ~~(0) & 300.00 (100) & 28.17 ~~(0) & 27.34 ~~(0) \\
180 & 300.00 (100) & 73.03 ~~(0) & 72.32 ~~(0) & 300.00 (100) & 72.88 ~~(0) & 72.43 ~~(0) \\
220 & 300.00 (100) & 148.87 (20) & 147.98 (20) & 300.00 (100) & 149.16 (19) & 148.35 (20) \\
260 & 300.00 (100) & 200.16 (44) & 199.02 (45) & 300.00 (100) & 200.20 (45) & 198.96 (46) \\
300 & 300.00 (100) & 230.51 (60) & 228.65 (60) & 300.00 (100) & 230.54 (60) & 228.76 (60) \\
340 & 300.00 (100) & 250.51 (70) & 248.46 (70) & 300.00 (100) & 250.64 (70) & 248.50 (70) \\
380 & 300.00 (100) & 264.10 (80) & 262.12 (80) & 300.00 (100) & 264.23 (80) & 262.10 (80) \\
420 & 300.00 (100) & 275.91 (80) & 273.07 (80) & 300.00 (100) & 276.02 (80) & 273.17 (80) \\
460 & 300.00 (100) & 282.03 (90) & 280.20 (90) & 300.00 (100) & 282.11 (90) & 280.14 (90) \\
					\bottomrule
					\end{tabular}
					}

					\caption{Non-3-colorability}
					\label{tab:graphcoloring}
				\end{table}
			}

			\leanparagraph{Nonexistence of a Vertex Covering.}
			Next, we consider the coNP-complete problem of checking \rev{if}{whether} for a given undirected graph
			there is no vertex covering of a certain maximal size.
			More precisely, given a graph $\langle V, E \rangle$, a vertex covering is a node selection $C \subseteq V$
			such that for each edge $\{v, u\} \in E$ we have $\{v,u\} \cap C \not= \emptyset$.
			\reva{As before we want the program to be consistent in case there is no vertex covering of the given maximum size, and a dedicated atom should represent this.}
			Our instances consist of such a graph $\langle V, E \rangle$, given by \rev{predicates}{atoms of kind} $\mathit{node}(\cdot)$ and $\mathit{edge}(\cdot, \cdot)$,
			and \rev{an}{a positive} integer $L$ (\emph{limit}), given by $\mathit{limit}(L)$. The task is to decide \rev{if}{whether} there is
			\emph{no} vertex covering containing at most $L$ nodes.
			The size of the instances is the number of nodes $n = |V|$.
			
			Similarly as for the previous benchmark, we use an encoding which splits the guessing part $P_{\mathit{nonVC}}$ from the checking part,
			where the latter is realized as an external source.
			An important difference to the previous benchmark is that the checking component must now aggregate over the node selection to check the size constraint.
			\rev{An encoding as an ordinary ASP-program is challenging since the coNP-completeness
			requires \reva{again} a saturation encoding and
			}{
			Since we want the program to be consistent whenever there is \emph{no} vertex covering, we need again a saturation encoding. However,}
			the size check requires aggregate atoms, which means that aggregate atoms must be used in a cycle; many reasoners do not support this.
			However, \hex-program\rev{}{s}, which inherently support cyclic external atoms, allow for pushing the check into an external source.
			
			The number and size of support sets is polynomial in the size of the graph, but exponential in the limit $L$.
			In this benchmark we consider $L$ to be a constant number \rev{which}{that} is for each instance randomly chosen from the range $1 \le L \le 20$\rev{;}{.}
			\rev{we}{We} exclude instances with graphs $\langle V, E \rangle$ and limits $L$ such that $L \ge |V|$ as in such cases the final answer to the considered problem is trivially false
			(since $V$ is trivially a vertex covering of size no greater than $L$).		

			The encoding is as follows.
			The guessing part is similar as
			\revtwo{before. For a}{before and construct a} candidate vertex covering given by atoms of kind $\mathit{in}(n)$ or $\mathit{out}(n)$ for nodes $n$\revtwo{,}{. In the checking part,}
			the external atom $\ext{\mathit{checkVC}}{\mathit{in}, \mathit{out}, \mathit{edge}, \rev{S}{L}}{}$ is true iff $\mathit{in}$ and $\mathit{out}$ encode \rev{a valid}{an invalid}
			vertex covering of the graph specified
			by $\mathit{edge}$ of size no greater than limit $L$.
			A complete family of support sets for $\ext{\mathit{checkVC}}{\mathit{in}, \mathit{out}, \mathit{edge}, L}{}$ is of size at most $n^L$, where $L$ is bounded in our scenario.
			\begin{align*}
				P_{\mathit{nonVC}} = \big\{ \mathit{in}(V) \vee \mathit{out}(V) \leftarrow\ & \mathit{node}(V) \\[0.3ex]
											\mathit{inval} \leftarrow\ & \ext{\mathit{checkVC}}{\mathit{in}, \mathit{out}, \mathit{edge}, L}{}, \mathit{limit}(L) \\[0.3ex]
											\mathit{in}(V) \leftarrow\ & \mathit{inval}, \mathit{node}(V) \\
											\mathit{out}(V) \leftarrow\ & \mathit{inval}, \mathit{node}(V) \\
											\leftarrow\ & \reva{\naf{}} \mathit{inval} \big\}
			\end{align*}

			The results are shown in Table~\ref{tab:vertexcover}.%
			\revam{Although the maximum size $L$ of the vertex covering is limited and the number of support sets is thus polynomial in the instance size,
			it is significantly larger compared to the non-3-colorability problem; the benchmark shows that the approach is still feasible in such cases.}
			\reva{Note that although $L$ is bounded and the size of the family of support sets $n^L$ is therefore polynomial in the size of the graph,
			it is in general still much larger than in the previous benchmark. This is because the order $L$ of the polynom is randomly chosen
			such that $1 \le L \le \mathit{min}(20, |V|)$, where $|V|$ is the size of the respective instance,
			while for non-3-colorability the family of support sets is always quadratic in the size of the input graph.
			The benchmark shows that the approach is still feasible in such cases.}
			\rev{Checking guesses based on support sets in the \textbf{sup.sets} configuration is then more expensive than for non-3-colorability, which
			makes the relative speedup of \textbf{sup.sets} over \textbf{traditional} smaller (but still clearly visible).
			Thus, there is now more room for further improvement by eliminating this check as in the \textbf{inlining} configuration.
			Once more, eliminating the compatibility check altogether yields a a further speedup.}
			{Here, checking guesses based on support sets in the \textbf{sup.sets} configuration is more expensive than for non-3-colorability
			because the verification of guesses requires a significantly larger number of comparisons to support sets. This
			makes the relative speedup of \textbf{sup.sets} over \textbf{traditional} smaller (but still clearly visible).
			On the other hand\revtwo{}{,} there is now more room for further improvement by the \textbf{inlining} configuration.
			Eliminating the (more expensive) check against support sets altogether yields now a larger further speedup.}

			{
				\begin{table}[t]
					\centering
					\begin{tabular}[t]{r|rrr|rrr}
					\toprule
					$n$ & \multicolumn{3}{c|}{all answer sets} & \multicolumn{3}{c}{first answer set} \\
					& traditional & sup.sets & inlining & traditional & sup.sets & inlining \\
					\midrule
8 & 15.45 ~~~~(0) & 4.13 ~~(0) & 0.61 ~~(0) & 15.42 ~~~~(0) & 4.12 ~~(0) & 0.61 ~~(0) \\
9 & 62.89 ~~(11) & 31.23 ~~(8) & 7.72 ~~(0) & 62.81 ~~(11) & 31.26 ~~(8) & 7.64 ~~(0) \\
10 & 102.15 ~~(22) & 80.65 (24) & 36.09 ~~(8) & 102.17 ~~(22) & 80.57 (24) & 36.11 ~~(8) \\
11 & 181.35 ~~(55) & 89.87 (26) & 47.42 (13) & 181.41 ~~(55) & 89.96 (26) & 47.45 (13) \\
12 & 222.05 ~~(66) & 135.79 (43) & 89.43 (25) & 222.05 ~~(66) & 135.82 (43) & 89.36 (25) \\
13 & 256.16 ~~(82) & 158.63 (50) & 110.26 (32) & 256.16 ~~(82) & 158.71 (51) & 110.19 (32) \\
14 & 288.93 ~~(96) & 189.18 (62) & 152.59 (50) & 288.94 ~~(96) & 189.24 (62) & 152.60 (50) \\
15 & 284.97 ~~(93) & 178.66 (59) & 145.46 (47) & 284.96 ~~(93) & 178.66 (59) & 145.42 (47) \\
16 & 294.77 ~~(98) & 219.03 (72) & 191.25 (62) & 294.74 ~~(98) & 218.98 (72) & 191.21 (62) \\
17 & 300.00 (100) & 219.19 (73) & 175.57 (56) & 300.00 (100) & 219.19 (73) & 175.45 (56) \\
18 & 300.00 (100) & 231.10 (77) & 195.14 (63) & 300.00 (100) & 231.10 (77) & 195.13 (63) \\
19 & 300.00 (100) & 243.12 (81) & 220.70 (71) & 300.00 (100) & 243.11 (81) & 220.72 (71) \\
20 & 300.00 (100) & 237.07 (79) & 217.87 (70) & 300.00 (100) & 237.07 (79) & 217.86 (70) \\
					\bottomrule
					  \end{tabular}

					\caption{Nonexistence of a vertex covering}
					\label{tab:vertexcover}

				\end{table}
			}

			\leanparagraph{Discussion and Summary.}
			As stated above, this paper focuses on external sources \rev{who}{that} possess a compact complete family of support sets.
			For the sake of completeness we still discuss also the case where a complete family of support sets is not small.
			As an extreme case, consider $P = \{ p(n+1) \leftarrow \rev{\ext{\mathit{even}}{p}}{\ext{\mathit{even}}{p}{}} \} \cup \{ p(i) \leftarrow {} \mid 1 \le i \le n \}$ for a given integer $n$,
			where $\ext{\mathit{even}}{p}{}$ is true iff the number of true atoms over $p$ is even.
			The program has a single answer set $Y = \{ p(i) \mid 1 \le i \le n \}$ if $n$ is odd, and no answer set if $n$ is even.
			\reva{This is because $p(n+1)$ would be derived based on $\ext{\mathit{even}}{p}{}$, which makes the number of $p$-atoms odd and destroys support of $p(n+1)$.}
			In any case, $\hat{P}$ has only two candidates which are easily checked in the \textbf{traditional} approach, while exponentially many
			support sets must be generated to represent the semantics of $\ext{\mathit{even}}{p}$ (one for each subset of $\{ p(i) \leftarrow {} \mid 1 \le i \le n \}$ with an even number of elements).
			In such cases, \textbf{traditional} might be exponentially faster than \textbf{sup.sets} and \textbf{inlining}.
			
			However, this is not the case for many realistic types of external sources,
			where the existence of \revam{of }a compact family of support sets is often even provable,
			such as the ones we used in our experiments.
			The size of the \textbf{inlining} encoding is directly linked to the size of the complete family of support sets,
			and if this size is small then the \textbf{inlining} approach is clearly superior to \textbf{sup.sets}
			as it eliminates the compatibility check and minimality check wrt.~external sources altogether, while
			it has only slightly higher initialization overhead. This overhead can be neglected even in
			cases where there is no further speedup by \textbf{inlining}.
			\textbf{Sup.sets} is in turn superior to \textbf{traditional} (even with learning technique EBL)
			as already observed by~\citeN{eiterFR014}.
			We can therefore conclude that \textbf{inlining} is a significant improvement over \textbf{sup.sets}
			and, for the considered types of external sources, also over \textbf{traditional}.

	\section{Equivalence of \hex-Programs}
	\label{sec:equivalence}

		\revam{In previous works, characterizations of equivalence of (ordinary) answer set programs under programs extensions have been developed.}
		\reva{In this section we present another application of the technique of external source inlining from Section~\ref{sec:inlining}.}
		\rev{That is, two}{Two} programs $P$ and $Q$ are considered to be equivalent if $P \cup R$ and $Q \cup R$ have the same answer sets
		for all programs $R$ of a certain type, which depends on the notion of equivalence at hand.
		Most importantly, for \emph{strongly equivalent} programs we have that $P \cup R$ and $Q \cup R$ have the same answer sets for any program $R$~\cite{Lifschitz:2001:SEL:383779.383783},
		while \emph{uniformly equivalent} programs guarantee this only if $R$ is a set of facts~\cite{DBLP:conf/iclp/EiterF03}.
		Later, these notions were extended to the non-ground case~\cite{DBLP:conf/aaai/EiterFTW05}.
		\rev{A more fine-grained approach is the one}{We will use the more fine-grained notion of $\langle \mathcal{H}, \mathcal{B} \rangle$-equivalence} by~\citeN{DBLP:journals/tplp/Woltran08},
		where $R$ can contain rules other than facts, but the sets of atoms \rev{which}{that} can occur in rule heads and bodies are restricted \reva{by sets of atoms
		$\mathcal{H}$ and $\mathcal{B}$, respectively. This notion generalizes both strong and uniform equivalence}.
		\reva{Formal criteria allow for semantically characterizing equivalence of two programs.}

		\reva{We extend a characterization
		of $\langle \mathcal{H}, \mathcal{B} \rangle$-equivalence from ordinary ASP- to \hex-programs.}%
		\revam{Besides efficiency improvements, external source inlining can also be exploited for
		extending such equivalence notions from ordinary ASP to \hex-programs.
		In this section we present such an extension of the notion
		by~\citeN{DBLP:journals/tplp/Woltran08}, which subsumes strong and uniform equivalence as special cases.
		More specifically, we present a notions for equivalence of \hex-programs $P$ and $Q$ under extensions by additional rules $R$.
		The possibly added rules are constrained such that their head and body atoms and input atoms to external atoms can only come from fixed sets.}
		Due to the support for external atoms, which can even be nonmonotonic,
		and the use of the FLP-reduct~\cite{flp2011-ai} instead of the GL-reduct~\cite{gelf-lifs-88} in the semantics of \hex-programs, this result is not immediate.
		Since well-known ASP extensions \revam{(}such as programs with aggregates~\cite{flp2011-ai} and constraint ASP~\cite{geossc09a,os2012-tplp}\revam{)}
		\rev{amount to}{are} special cases of \hex-programs, the results \rev{are interesting beyond the specific formalism}{carry over}.
		
		We proceed as follows. In the first step (Section~\ref{sec:equivalence:generalization}), only the programs $P$ and $Q$ can be \hex-programs,
		but the added program $R$ must be ordinary.
		This amounts to a generalization of the results by~\citeN{DBLP:journals/tplp/Woltran08} from ordinary ASP to \hex-programs.
		In the second step (Section~\ref{sec:equivalence:generalhex}), we allow also the added program $R$ to contain external atoms.
		For this purpose, we exploit the possibility to inline external atoms.
	
		\subsection{Generalizing Equivalence Results}
		\label{sec:equivalence:generalization}

			In the following, for sets $\mathcal{H}$ and $\mathcal{B}$ of atoms we let
			$\mathcal{P}_{\langle \mathcal{H}, \mathcal{B} \rangle} = \{ P \text{ is an ASP-program} \hspace{-0.9mm} \mid \hspace{-0.9mm} H(P) \subseteq \mathcal{H}, B^{+}(P) \cup B^{-}(P) \subseteq \mathcal{B} \}$
			be the set of ordinary programs whose head and body atoms come only from $\mathcal{H}$ and $\mathcal{B}$, respectively.
			Ordinary ASP-programs $P$ and $Q$ are called $\langle \mathcal{H}, \mathcal{B} \rangle$-equivalent,
			if the answer sets of $P \cup R$ and $Q \cup R$ are the same for all ordinary ASP-programs $R$ \rev{which}{that} use only head atoms from $\mathcal{H}$ and only body atoms from $\mathcal{B}$,
			i.e., $R \in \mathcal{P}_{\langle \mathcal{H}, \mathcal{B} \rangle}$.
			
			We first lift this \rev{result}{definition} to the case where $P$ and $Q$ are general \hex-programs which possibly contain external atoms, while $R$ remains an ordinary ASP-program. Formally:

			\begin{definition}
				\label{def:hbequivalence}
				\hex-programs $P$ and $Q$ are \emph{equivalent wrt.~a pair $\langle \mathcal{H}, \mathcal{B} \rangle$ of sets of atoms},
				or \emph{$\langle \mathcal{H}, \mathcal{B} \rangle$-equivalent}, denoted $P \equiv_{\langle \mathcal{H}, \mathcal{B} \rangle} Q$,
				if $\mathcal{AS}(P \cup R) = \mathcal{AS}(Q \cup R)$ for all $R \in \mathcal{P}_{\langle \mathcal{H}, \mathcal{B} \rangle}$.
			\end{definition}
			
			Similarly, we write $P \subseteq_{\langle \mathcal{H}, \mathcal{B} \rangle} Q$
			if $\mathcal{AS}(P \cup R) \subseteq \mathcal{AS}(Q \cup R)$ for all $R \in \mathcal{P}_{\langle \mathcal{H}, \mathcal{B} \rangle}$.

			Towards a characterization of equivalence \reva{of \hex-programs}, one can first show that if there is a counterexample $R$ for $P \equiv_{\langle \mathcal{H}, \mathcal{B} \rangle} Q$,
			i.e., an $R \in \mathcal{P}_{\langle \mathcal{H}, \mathcal{B} \rangle}$ such that $\mathcal{AS}(P \cup R) \not= \mathcal{AS}(Q \cup R)$,
			then there is also a simple counterexample in form of a positive program $R' \in \mathcal{P}_{\langle \mathcal{H}, \mathcal{B} \rangle}$.

			\addProposition{prop:simpleCounterExample}{
				Let $P$ and $Q$ be \hex-programs, $R$ be an ordinary ASP-program, and $Y$ be an assignment s.t.~$Y \in \mathcal{AS}(P \cup R)$ but $Y \not\in \mathcal{AS}(Q \cup R)$.
				Then there is also a positive ordinary ASP-program $R'$ such that
				$Y \in \mathcal{AS}(P \cup R')$ but $Y \not\in \mathcal{AS}(Q \cup R')$
				and $B(R') \subseteq B(R)$ and $H(R') \subseteq H(R)$.
			}

			\addProof{prop:simpleCounterExample}{
				Let $P$ and $Q$ be \hex-programs, $R$ be an ordinary ASP-program, and $Y$ be an assignment such that $Y \in \mathcal{AS}(P \cup R)$ but $Y \not\in \mathcal{AS}(Q \cup R)$.
				We have to show that there is a positive $R'$ such that $Y \in \mathcal{AS}(P \cup R')$ but $Y \not\in \mathcal{AS}(Q \cup R')$.
				As~\citeN{DBLP:journals/tplp/Woltran08}, we show this in particular
				for $R' = R^Y$, where $R^Y = \{ H(r) \leftarrow B^{+}(r) \mid r \in R, \rev{Y \not\models B^{-}(r)}{Y \not\models b \text{ for all } b \in B^{-}(r)} \}$ is the GL-reduct~\cite{gelf-lifs-88},
				not to be confused with the FLP-reduct which is used in the definition of the \hex-semantics.
				Obviously we have $B(R') \subseteq B(R)$ and $H(R') \subseteq H(R)$.
				
				\begin{itemize}
					\item We first show that $Y \in \mathcal{AS}(P \cup R')$.
						For modelhood, we know that $Y$ is a model of $P$, thus it suffices to discuss $R'$.
						Let $r' \in R'$. Then there is a corresponding rule $r \in R$ such that $r'$ is the only rule in $\{ r \}^Y$.
						We have that $Y \not\models B^{-}(r)$, otherwise $r'$ would not be in $\{ r \}^Y$. But then, since $Y \models r$
						(because $Y \models R$ since $Y \in \mathcal{AS}(P \cup R)$ by assumption), we have that $Y \models H(r)$ or $Y \not\models B^{+}(r)$,
						which implies that $Y \models r'$.
						
						It remains to show that there is no $Y' \subsetneq Y$ such that $Y' \models f (P \cup R')^Y$. Towards a contradiction, suppose there is such an $Y'$;
						we show that it is also a model of $f (P \cup R)^Y$, which contradicts the assumption that $Y$ is an answer set of $P \cup R$.
						Obviously we have $Y' \models f P^Y$. 
						Now consider $r \in f R^Y$. Then $Y \models B^{+}(r)$ and $Y \not\models B^{-}(r)$. But then $H(r) \leftarrow B^{+}(r) \in R'$ and $H(r) \leftarrow B^{+}(r) \in f R'^Y$.
						Since $Y' \models f R'^Y$, we have that $Y' \models H(r)$ or $Y' \not\models B^{+}(r)$ and thus $Y' \models r$.
						Since this holds for all $r \in f R^Y$ this implies $Y' \models f (P \cup R)^Y$,
						which contradicts the assumption that $Y$ is an answer set of $P \cup R$, thus $Y'$ cannot exist and $Y$ is an answer set of $P \cup R'$.
						
					\item We now show that $Y \not\in \mathcal{AS}(Q \cup R')$.
						If $Y \not\models Q \cup R$ then also $Y \not\models Q \cup R'$ because for each $r \in R$ we either have that $Y \models B^{-}(r)$
						(and thus $r$ is not relevant for the inconsistency of $Q \cup R$)
						or $R'$ contains $H(r) \leftarrow B^{+}(r)$ instead, which is even harder to satisfy (i.e., is violated whenever $r$ is).
						
						If $Y \models Q \cup R$ then there is an $Y' \subsetneq Y$ such that $Y' \models f (Q \cup R)^Y$. We show that $Y'$ is also a model of $f (Q \cup R')^Y$.
						Towards a contradiction, suppose there is an $r' \in f (Q \cup R')^Y$ such that $Y' \not\models r'$.
						Then $r'$ must be in $f R'^Y$ because if it would be in $f Q^Y$ then $Y'$ could not be a model of $f (Q \cup R)^Y$.
						Then $Y' \not\models H(r')$ but $Y' \models B^{+}(r')$.
						But then there is a rule $r \in f R^Y$ with $H(r) = H(r')$ and $B^{+}(r) = B^{+}(r')$ such that $Y' \not\models B^{-}(r)$
						(otherwise $Y \models B^{-}(r)$ and $r'$ could not be in $R'$ and thus also not in $f (Q \cup R')^Y$).
						However, then $Y' \not\models r$ and thus $Y' \not\models f (Q \cup R)^Y$, which contradicts our assumption.
				\end{itemize}
			}

			\reva{The idea of the constructive proof is to show for given programs $P$, $Q$ and $R$ and an assignment $Y$ that the GL-reduct~\cite{gelf-lifs-88} $R^Y$,
			which is a positive program, is such a simple counterexample.}

			Next, we show that the concepts on equivalence generalize from ordinary ASP to \hex-programs.		
			In the following, for an assignment $Y$ and a set of atoms $A$
			we write $Y|_{A}$ for for the projection $Y \cap A$ of $Y$ to $A$.
			Moreover, for sets of atoms $X$, $Y$ we write $X \leq^{\mathcal{B}}_{\mathcal{H}} Y$ if $X|_{\mathcal{H}} \subseteq Y|_{\mathcal{H}}$ and $X|_{\mathcal{B}} \supseteq Y|_{\mathcal{B}}$.
			Intuitively, if $X \leq^{\mathcal{B}}_{\mathcal{H}} Y$ then $Y$ satisfies all positive programs from $\mathcal{P}_{\langle \mathcal{H}, \mathcal{B} \rangle}$ \rev{which}{that} are also satisfied by $X$
			because it satisfies no fewer heads and no more bodies than $X$.
			We write $X <^{\mathcal{B}}_{\mathcal{H}} Y$ if $X \leq^{\mathcal{B}}_{\mathcal{H}} Y$ and $X|_{\mathcal{H} \cup \mathcal{B}} \not= Y|_{\mathcal{H} \cup \mathcal{B}}$.

			We use the following concept for witnessing that $\mathcal{AS}(P \cup R) \subseteq \mathcal{AS}(Q \cup R)$ does \emph{not} hold.

			\begin{definition}
				\label{def:witness}
				A witness for $P \not\subseteq_{\langle \mathcal{H}, \mathcal{B} \rangle} Q$ is a pair $(X, Y)$ of assignments with $X \subseteq Y$ such that\footnote{
					Note that~\citeN{DBLP:journals/tplp/Woltran08} called this \emph{a witness for $P \subseteq_{\langle \mathcal{H}, \mathcal{B} \rangle} Q$},
					but since it is actually a witness for the violation of the containment, we change the terminology.
				}:
				\begin{enumerate}[(i)]
					\item $Y \models P$ and for each $Y' \subsetneq Y$ with $Y' \models f P^Y$ we have $Y'|_{\mathcal{H}} \subsetneq Y|_{\mathcal{H}}$; and
					\item if $Y \models Q$ then $X \subsetneq Y$, $X \models f Q^Y$ and for all $X'$ with $X \leq^{\mathcal{B}}_{\mathcal{H}} X' \subsetneq Y$ we have $X' \not\models f P^Y$.
				\end{enumerate}
			\end{definition}
			
			The idea is that a witness represents a counterexample to the containment\rev{, where}{. To this end,} $X$
			characterizes a program $R$ \rev{such that $Y$}{and $Y$ is an assignment \revtwo{that it}{that}} is an answer set of $P \cup R$ but not of $Q \cup R$.
			One can show that the existence of a witness and the violation of the containment are equivalent.
			
			Because some steps in the according considerations for ordinary ASP depend on the fact that GL-reducts of programs wrt.~assignments are positive programs (cf.~$\leq^{\mathcal{B}}_{\mathcal{H}}$),
			it is an interesting result that the following propositions still hold in its generalized form.
			Because we use FLP-reducts instead, and $P$ and $Q$ might even contain nonmonotonic external atoms, the results do not automatically carry over.
			However, a closer analysis reveals that the property of being a positive program is only required for the reduct\revam{s} of $R$ but not \rev{}{the reducts} of $P$ or $Q$.
			Since we restricted $R$ to ordinary ASP-programs for now, and Proposition~\ref{prop:simpleCounterExample} allows us to further restrict it to positive programs,
			the use of the FLP-reduct does not harm: if $R$ is positive from the beginning, then also its FLP-reduct (wrt.~any assignment) is positive.
			\reva{Hence, the main idea is that due to restrictions of the input program, the reduct is still guaranteed to be positive despite the switch from the GL- to the FLP-reduct.
			This allows for lifting the proof of the following proposition from ordinary ASP to \hex.}

			\addProposition{prop:containment}{
				For \hex-programs $P$ and $Q$ and sets $\mathcal{H}$ and $\mathcal{B}$ of atoms,
				there is a program $R \in \mathcal{P}_{\langle \mathcal{H}, \mathcal{B} \rangle}$ with $\mathcal{AS}(P \cup R) \not\subseteq \mathcal{AS}(Q \cup R)$
				iff there is a witness for $P \not\subseteq_{\langle \mathcal{H}, \mathcal{B} \rangle} Q$.
			}

			\addProof{prop:containment}{
				($\Rightarrow$)
					If $\mathcal{AS}(P \cup R) \not\subseteq \mathcal{AS}(Q \cup R)$ for a program $R$, then there is an assignment $Y$
					such that $Y \in \mathcal{AS}(P \cup R)$ but $Y \not\in \mathcal{AS}(Q \cup R)$. Due to Proposition~\ref{prop:simpleCounterExample}
					we can assume that $R$ is a positive program.
					
					We show that $Y$ satisfies Condition~(i) of Definition~\ref{def:witness}. Since $Y \in \mathcal{AS}(P \cup R)$ we have $Y \models P$.
					Towards a contradiction, suppose there is an $Y' \subsetneq Y$ such that $Y' \models f P^Y$ and $Y'|_{\mathcal{H}} \not\subsetneq Y|_{\mathcal{H}}$.
					Then, since $Y' \subseteq Y$, we have $Y'|_{\mathcal{H}} = Y|_{\mathcal{H}}$. We further have $Y'|_{\mathcal{B}} \subseteq Y|_{\mathcal{B}}$, i.e.,
					$Y \leq^{\mathcal{B}}_{\mathcal{H}} Y'$.
					Since $R$ is positive, $Y \models R$ implies $Y' \models R$, and since $f R^Y \subseteq R$ this further implies $Y' \models f R^Y$.
					Since we further have $Y' \models f P^Y$ this gives $Y' \models f (P \cup R)^Y$
					and thus $Y$ cannot be an answer set of $P \cup R$, which contradicts our assumption and therefore Condition~(i) is satisfied.
					
					We show now that there is an $X$ such that $(X,Y)$ satisfies also Condition~(ii), i.e., is a witness as by Definition~\ref{def:witness}.
					If $Y \not\models Q$ then Condition~(ii) is trivially satisfied for any $X \subseteq Y$ and e.g. $(Y,Y)$ is a witness.
					Otherwise ($Y \models Q$), note that we have $Y \models R$ since $Y \in \mathcal{AS}(P \cup R)$. Together with the precondition that $Y \not\in \mathcal{AS}(Q \cup R)$
					this implies that there is an $X \subsetneq Y$ such that $X \models f (Q \cup R)^Y$, which is equivalent to $X \models f Q^Y$ and $X \models f R^Y$.
					We show that for this $X$, Condition~(ii) is satisfied, hence $(X,Y)$ is a witness. As we already have $X \subsetneq Y$ and $X \models f Q^Y$,
					it remains to show that for any $X'$ with $X \leq^{\mathcal{B}}_{\mathcal{H}} X' \subsetneq Y$ we have $X' \not\models f P^Y$.
					If there would be an $X'$ with $X \leq^{\mathcal{B}}_{\mathcal{H}} X' \subsetneq Y$ with $X' \models f P^Y$,
					then, since we also have $X' \models f R^Y$ (because $X \models f R^Y$ and $f R^Y$ is positive),
					this implies $X' \models f (P \cup R)^Y$ and contradicts the precondition that $Y \in \mathcal{AS}(P \cup R)$.
					Thus such an $X'$ cannot exist and Condition~(ii) is satisfied by $(X,Y)$.

				($\Leftarrow$)
					Let $(X,Y)$ be a witness for $\mathcal{AS}(P \cup R) \not\subseteq \mathcal{AS}(Q \cup R)$.
					We make a case distinction: either $Y \not\models Q$ or $Y \models Q$.
					
					\begin{itemize}
						\item Case $Y \not\models Q$:
						
							We show for the following $R \in \mathcal{P}_{\langle \mathcal{H}, \mathcal{B} \rangle}$ that $Y \in \mathcal{AS}(P \cup R)$ but $Y \not\in \mathcal{AS}(Q \cup R)$:
								$$R = \{ a \leftarrow {} \mid a \in Y|_{\mathcal{H}} \}$$
								
							Since $(X,Y)$ is a witness, by Property~(i) we have $Y \models P$. We further have $Y \models R$, thus $Y \models P \cup R$.
							Moreover, we obviously have $f R^Y = R$, which contains all atoms from $Y|_{\mathcal{H}}$ as facts.
							Suppose there is a $Y' \subsetneq Y$ such that $Y' \models f (P \cup R)^Y$;
							then $Y' \models f P^Y$ and by Property~(i) we have $Y'|_{\mathcal{H}} \subsetneq Y|_{\mathcal{H}}$, i.e., at least one atom from $Y|_{\mathcal{H}}$
							is unsatisfied under $Y'$. But then $Y' \not\models f R^Y$ and thus $Y' \not\models f (P \cup R)^Y$, i.e.,
							$Y$ is an answer set of $P \cup R$.
							On the other hand, $Y \not\models Q$ implies $Y \not\models Q \cup R$ and therefore $Y$ cannot be an answer set of $Q \cup R$.
					
						\item Case $Y \models Q$:
						
							We show for the following $R \in \mathcal{P}_{\langle \mathcal{H}, \mathcal{B} \rangle}$ that $Y \in \mathcal{AS}(P \cup R)$ but $Y \not\in \mathcal{AS}(Q \cup R)$:
							\begin{align*}
								R = &\{ a \leftarrow {} \mid a \in X|_{\mathcal{H}} \} \ \cup \\
									&\{ a \leftarrow b \mid a \in (Y \setminus X)|_{\mathcal{H}}, b \in (Y \setminus X)|_{\mathcal{B}} \}
							\end{align*}
							
							We first show that $Y \in \mathcal{AS}(P \cup R)$.
							Since $(X,Y)$ is a witness as by Definition~\ref{def:witness}, we have $Y \models P$.
							We further have $Y \models R$ by construction of $R$ because all heads of its rules are in $Y$.

							Thus it remains to show that it is also a subset-minimal model of $f (P \cup R)^Y$.
							Towards a contradiction, assume that there is a $Z \subsetneq Y$ such that $Z \models f (P \cup R)^Y$,
							which is equivalent to $Z \models f P^Y$ and $Z \models f R^Y$, where $f R^Y = R$ (by construction of $R$).
							By construction of $R$, $Z \models R$ implies that $X|_{\mathcal{H}} \subseteq Z|_{\mathcal{H}}$.
							Property~(i) of Definition~\ref{def:witness} implies that $Z|_{\mathcal{H}} \subsetneq Y|_{\mathcal{H}}$
							and thus $X|_{\mathcal{H}} \subseteq Z|_{\mathcal{H}} \subsetneq Y|_{\mathcal{H}}$.
							This implies that there is an $a \in (Y \setminus X)|_{\mathcal{H}}$ which is not in $Z|_\mathcal{H}$.
							Since $Y \models Q$, $Z \subsetneq Y$, $Z \models f P^Y$ and $X|_{\mathcal{H}} \subseteq Z|_{\mathcal{H}}$,
							Property~(ii) further implies $Z|_{\mathcal{B}} \not\subseteq X|_{\mathcal{B}}$ (since violating $X \leq^{\mathcal{B}}_{\mathcal{H}} Z$ is the only remaining option to satisfy the property).
							As we also have $Z|_{\mathcal{B}} \subseteq Y|_{\mathcal{B}}$ (because $Z \subsetneq Y$), there is a $b \in (Y \setminus X)|_{\mathcal{B}}$ which is also in $Z$.
							Hence, we have an $a \in (Y \setminus X)|_{\mathcal{H}}$ and a $b \in (Y \setminus X)|_{\mathcal{B}}$ such that only $b$ is also in $Z$, hence the rule
							$a \leftarrow b \in R$ (and $a \leftarrow b \in f R^Y$) is violated by $Z$,
							thus $Z \not\models f R^Y$ and $Z \not\models f (P \cup R)^Y$, which contradicts our assumption.
							
							It remains to show that $Y \not\in \mathcal{AS}(Q \cup R)$. We already know that $Y \models Q \cup R$ and must show that $f (Q \cup R)^Y$ has a smaller model than $Y$.
							Since $(X,Y)$ is a witness, we have $X \subsetneq Y$ and $X \models f Q^Y$ by Property~(ii).
							As $X \models R$ (it satisfies all facts $\{ a \leftarrow {} \mid a \in X|_{\mathcal{H}} \}$ and no other rules of $R$ are applicable as their bodies contain only atoms that are not in $X$),
							we get $X \models f R^Y$
							and have $X \models f (Q \cup R)^Y$. Therefore $Y \not\in \mathcal{AS}(Q \cup R)$.
					\end{itemize}
			}
			
			While witnesses compare the sets of answer sets of two programs directly, the next concept of
			$\langle \mathcal{H}, \mathcal{B} \rangle$-models can be used to characterize a single program.
			In the following,
			for two sets of atoms $\mathcal{H}$ and $\mathcal{B}$,
			a pair $(X,Y)$ of assignments is called $\leq^{\mathcal{B}}_{\mathcal{H}}$-maximal
			for $P$ if $X \models f P^Y$ and for all $X'$ with $X <^{\mathcal{B}}_{\mathcal{H}} X' \subsetneq Y$ we have $X' \not\models f P^Y$.

			\begin{definition}
				\label{def:hbmodel}
				Given sets $\mathcal{H}$, $\mathcal{B}$ of atoms, a pair $(X,Y)$ of assignments is an
				$\langle \mathcal{H}, \mathcal{B} \rangle$-model of a program $P$ if
				\begin{enumerate}[(i)]
					\item $Y \models P$ and for each $Y' \subsetneq Y$ with $Y' \models f P^Y$ we have $Y'|_{\mathcal{H}} \subsetneq Y|_{\mathcal{H}}$; and
					\item if $X \subsetneq Y$ then there exists an $X' \subsetneq Y$ with $X'|_{\mathcal{H} \cup \mathcal{B}} = X$ such that $(X',Y)$ is $\leq^{\mathcal{B}}_{\mathcal{H}}$-maximal for $P$.
				\end{enumerate}
			\end{definition}
		
			\reva{Intuitively, $\langle \mathcal{H}, \mathcal{B} \rangle$-models $(X,Y)$ characterize potential answer sets $Y$ of a program $P$ and the models of its reducts $f P^Y$.
			More precisely, the assignments $Y$ represent classical models of a program
			which can potentially be turned into an answer set by adding a program from $R \in \mathcal{P}_{\langle \mathcal{H}, \mathcal{B} \rangle}$
			(which can be empty if $Y$ is already an answer set of $P$).
			Turning $Y$ into an answer set requires that smaller models of the reduct $f P^Y$ (if existing) can be eliminated,
			which is only possible if they contain fewer atoms from $\mathcal{H}$ since these are the only atoms which can get support by adding $R$ (cf.~Condition~(i)).
			Furthermore, for such a classical model $Y$,
			different models of the reduct $f P^Y$ that coincide on $\mathcal{H}$ and $\mathcal{B}$
			behave the same over $f (P \cup R)^Y$ for any $R \in \mathcal{P}_{\langle \mathcal{H}, \mathcal{B} \rangle}$: either all or neither of them are models of the extended reduct;
			such different models are represented by a single $\langle \mathcal{H}, \mathcal{B} \rangle$-model $(X,Y)$
			as formalized by Condition~(ii).
			}
			
			One can show that $\langle \mathcal{H}, \mathcal{B} \rangle$-equivalence of two programs can be reduced to a comparison of their $\langle \mathcal{H}, \mathcal{B} \rangle$-models.
			We denote the set of all $\langle \mathcal{H}, \mathcal{B} \rangle$-models of a program $P$ by $\sigma_{\langle \mathcal{H}, \mathcal{B} \rangle}(P)$.

			\appendToProofs{
				Towards a characterization of equivalence in terms of $\langle \mathcal{H}, \mathcal{B} \rangle$-models we introduce the following lemma.
			}
			
			\addLemmaOutsourcedOnly{lem:semodels}{
				For sets $\mathcal{H}$ and $\mathcal{B}$ of atoms and programs $P$, $Q$,
				$(Y,Y) \in \sigma_{\langle \mathcal{H}, \mathcal{B} \rangle}(P) \setminus \sigma_{\langle \mathcal{H}, \mathcal{B} \rangle}(Q)$
				iff there is a witness $(X,Y)$ for $P \subseteq_{\langle \mathcal{H}, \mathcal{B} \rangle} Q$ with $X|_{\mathcal{H}} = Y|_{\mathcal{H}}$.
			}
			
			\addProof{lem:semodels}{
				($\Rightarrow$)
					Since $(Y,Y) \in \sigma_{\langle \mathcal{H}, \mathcal{B} \rangle}(P)$,
					Property~(i) of Definition~\ref{def:witness} holds because Property~(i) of Definition~\ref{def:hbmodel} is the same and holds.
					For Property~(ii) of Definition~\ref{def:witness}, $(Y,Y) \not\in \sigma_{\langle \mathcal{H}, \mathcal{B} \rangle}(Q)$
					implies that either $Y \not\models Q$ or there is a $Y' \subsetneq Y$ such that $Y' \models f Q^Y$ and $Y'|_{\mathcal{H}} = Y|_{\mathcal{H}}$.
					In the former case, Property~(ii) of Definition~\ref{def:witness} holds trivially for all $X \subseteq Y$ and, e.g.,
					$(Y,Y)$ is witness for $P \subseteq_{\langle \mathcal{H}, \mathcal{B} \rangle} Q$, for which $Y|_{\mathcal{H}} = Y|_{\mathcal{H}}$ clearly holds.
					In case $Y \models Q$ we have that there is some $X \subsetneq Y$ with $X \models f Q^Y$ and $X|_{\mathcal{H}} = Y|_{\mathcal{H}}$.
					In order to show that $(X,Y)$ satisfies Property~(ii),
					it remains to show that for all $X'$ with $X \leq^{\mathcal{B}}_{\mathcal{H}} X' \subsetneq Y$ we have $X' \not\models f P^Y$.
					If there would be an $X'$ with $X \leq^{\mathcal{B}}_{\mathcal{H}} X' \subsetneq Y$ and $X' \models f P^Y$,
					then $X|_{\mathcal{H}} = Y|_{\mathcal{H}}$ would imply $X'|_{\mathcal{H}} = Y|_{\mathcal{H}}$, and thus
					Property~(i) of Definition~\ref{def:hbmodel} would be violated by $(Y,Y)$ wrt.~$P$,
					which contradicts the assumption that $(Y,Y) \in \sigma_{\langle \mathcal{H}, \mathcal{B} \rangle}(P)$.
					
				($\Leftarrow$)
					For a witness $(X,Y)$ for $P \subseteq_{\langle \mathcal{H}, \mathcal{B} \rangle} Q$ with $X|_{\mathcal{H}} = Y|_{\mathcal{H}}$,
					Property~(i) of Definition~\ref{def:witness}
					implies that $(Y,Y) \in \sigma_{\langle \mathcal{H}, \mathcal{B} \rangle}(P)$
					and it remains to show that $(Y,Y) \not\in \sigma_{\langle \mathcal{H}, \mathcal{B} \rangle}(Q)$.
					Since $(X,Y)$ is a witness with $X|_{\mathcal{H}} = Y|_{\mathcal{H}}$, we have either $Y \not\models Q$ or $X \subsetneq Y$ and $X \models f Q^Y$.
					In the former case $(Y,Y)$ cannot be an $\langle \mathcal{H}, \mathcal{B} \rangle$-model of $Q$ due to violation of Property~(i) of Definition~\ref{def:hbmodel}.
					In the latter case $(Y,Y)$ cannot be an $\langle \mathcal{H}, \mathcal{B} \rangle$-model of $Q$ since our assumption $X|_{\mathcal{H}} = Y|_{\mathcal{H}}$ also
					contradicts Property~(i) of Definition~\ref{def:hbmodel}.
			}
			
			\addProposition{prop:semodels}{
				For sets $\mathcal{H}$ and $\mathcal{B}$ of atoms and \hex-programs $P$ and $Q$,
				we have $P \equiv_{\langle \mathcal{H}, \mathcal{B} \rangle} Q$ iff $\sigma_{\langle \mathcal{H}, \mathcal{B} \rangle}(P) = \sigma_{\langle \mathcal{H}, \mathcal{B} \rangle}(Q)$.
			}

			\addProof{prop:semodels}{
				($\Rightarrow$)
					We make a proof by contraposition.
					Wlog.~assume there is an $(X,Y) \in \sigma_{\langle \mathcal{H}, \mathcal{B} \rangle}(P) \setminus \sigma_{\langle \mathcal{H}, \mathcal{B} \rangle}(Q)$
					(the case $(X,Y) \in \sigma_{\langle \mathcal{H}, \mathcal{B} \rangle}(Q) \setminus \sigma_{\langle \mathcal{H}, \mathcal{B} \rangle}(P)$ is symmetric).
					We have to show that then $P \equiv_{\langle \mathcal{H}, \mathcal{B} \rangle} Q$ does not hold.
					
					Since $(X,Y) \in \sigma_{\langle \mathcal{H}, \mathcal{B} \rangle}(P)$, we also have $(Y,Y) \in \sigma_{\langle \mathcal{H}, \mathcal{B} \rangle}(P)$ (cf.~Definition~\ref{def:hbmodel}).
					If $(Y,Y) \not\in \sigma_{\langle \mathcal{H}, \mathcal{B} \rangle}(Q)$ then by Lemma~\ref{lem:semodels} there is a witness $(X,Y)$ for $P \not\subseteq_{\langle \mathcal{H}, \mathcal{B} \rangle} Q$
					and thus by Proposition~\ref{prop:containment} there is a program $R \in \mathcal{P}_{\langle \mathcal{H}, \mathcal{B} \rangle}$ with $\mathcal{AS}(P \cup R) \not\subseteq \mathcal{AS}(Q \cup R)$,
					hence $P \equiv_{\langle \mathcal{H}, \mathcal{B} \rangle} Q$ does not hold.
					
					In case $(Y,Y) \in \sigma_{\langle \mathcal{H}, \mathcal{B} \rangle}(Q)$ we have $X \subsetneq Y$ ($X$ and $Y$ cannot be equal because $(X,Y) \not\in \sigma_{\langle \mathcal{H}, \mathcal{B} \rangle}(Q)$).
					We make a case distinction.
					
					\begin{itemize}
						\item Case 1: There exists an $X'$ with $X \leq^{\mathcal{B}}_{\mathcal{H}} X' \subsetneq Y$ such that $(X',Y) \in \sigma_{\langle \mathcal{H}, \mathcal{B} \rangle}(Q)$:
						
							Since $(Y,Y) \in \sigma_{\langle \mathcal{H}, \mathcal{B} \rangle}(Q)$ but $(X,Y) \not\in \sigma_{\langle \mathcal{H}, \mathcal{B} \rangle}(Q)$, the latter
							fails to satisfy Definition~\ref{def:hbmodel} due to Property~(ii).
							Then $X <^{\mathcal{B}}_{\mathcal{H}} X'$ must hold
							(rather than $X|_{\mathcal{H} \cup \mathcal{B}} = X'|_{\mathcal{H} \cup \mathcal{B}}$)
							because only in this case satisfaction of Property~(ii) of Definition~\ref{def:hbmodel} wrt.~$X$ can differ from satisfaction wrt.~$X'$.
							Then there is a $Z \subsetneq Y$ with $Z|_{\mathcal{H} \cup \mathcal{B}} = X'$ such that $(Z,Y)$ is $\leq^{\mathcal{B}}_{\mathcal{H}}$-maximal for $Q$ and thus $Z \models f Q^Y$.
							We show that $(Z,Y)$ is a witness for $P \not\subseteq_{\langle \mathcal{H}, \mathcal{B} \rangle} Q$.
							Since $(Y,Y) \in \sigma_{\langle \mathcal{H}, \mathcal{B} \rangle}(P)$, Property~(i) of Definition~\ref{def:witness} holds for $(Z,Y)$.
							Moreover, we have $Z \models f Q^Y$ and, since $(X,Y) \in \sigma_{\langle \mathcal{H}, \mathcal{B} \rangle}(P)$, we have by Property~(ii) of Definition~\ref{def:hbmodel} for all $X''$ with
							$X <^{\mathcal{B}}_{\mathcal{H}} X'' \subsetneq Y$ that $X'' \not\models f P^Y$.
							Since $X <^{\mathcal{B}}_{\mathcal{H}} Z$ (as a consequence of $Z|_{\mathcal{H} \cup \mathcal{B}} = X'$ and $X <^{\mathcal{B}}_{\mathcal{H}} X'$),
							Property~(ii) of Definition~\ref{def:witness} holds for $Z$ and thus $(Z,Y)$ is a witness for $P \not\subseteq_{\langle \mathcal{H}, \mathcal{B} \rangle} Q$.
							
						\item Case 2: For each $X'$ with $X \leq^{\mathcal{B}}_{\mathcal{H}} X' \subsetneq Y$ we have $(X',Y) \not\in \sigma_{\langle \mathcal{H}, \mathcal{B} \rangle}(Q)$:
						
							We already have $(X,Y) \in \sigma_{\langle \mathcal{H}, \mathcal{B} \rangle}(P)$ and thus there is a $Z \subsetneq X$ with $Z|_{\mathcal{H} \cup \mathcal{B}} = X$
							such that $Z \models f P^Y$. We show that $(Z,Y)$ is a witness for the reverse problem $Q \not\subseteq_{\langle \mathcal{H}, \mathcal{B} \rangle} P$.
							Since $(Y,Y) \in \sigma_{\langle \mathcal{H}, \mathcal{B} \rangle}(Q)$ we have that Property~(i) of Definition~\ref{def:witness} is satisfied.
							We have $Y \models P$ (due to Property~(i) of Definition~\ref{def:hbmodel} wrt.~$(X,Y) \in \sigma_{\langle \mathcal{H}, \mathcal{B} \rangle}(P)$),
							thus for satisfaction of Property~(ii) of Definition~\ref{def:witness} recall that we have $Z \models f P^Y$ and it remains to show
							that for each $X''$ with $X \leq^{\mathcal{B}}_{\mathcal{H}} X'' \subsetneq Y$ we have $X'' \not\models f Q^Y$.
							If there would be such an $X''$ with $X'' \models f Q^Y$, then there would also a $\leq^{\mathcal{B}}_{\mathcal{H}}$-maximal one $X'''$
							and $(X''',Y)$ would be an $\langle \mathcal{H}, \mathcal{B} \rangle$-model of $Q$, which
							contradicts our assumption that $(X',Y) \not\in \sigma_{\langle \mathcal{H}, \mathcal{B} \rangle}(Q)$ for each $X'$ with $X \leq^{\mathcal{B}}_{\mathcal{H}} X' \subsetneq Y$.
					\end{itemize}

				($\Leftarrow$)
					We make a proof by contraposition.
					Suppose $P \not\equiv_{\langle \mathcal{H}, \mathcal{B} \rangle} Q$, then either
					$P \subseteq_{\langle \mathcal{H}, \mathcal{B} \rangle} Q$ or $Q \subseteq_{\langle \mathcal{H}, \mathcal{B} \rangle} P$
					does not hold; we assume wlog.~that $P \subseteq_{\langle \mathcal{H}, \mathcal{B} \rangle} Q$ does not hold (the other case is symmetric).
					We have to show that $\sigma_{\langle \mathcal{H}, \mathcal{B} \rangle}(P) = \sigma_{\langle \mathcal{H}, \mathcal{B} \rangle}(Q)$ does not hold either.
					
					By Proposition~\ref{prop:containment} there is a witness $(X,Y)$ for $P \not\subseteq_{\langle \mathcal{H}, \mathcal{B} \rangle} Q$.
					Then by Property~(i) of Definition~\ref{def:witness} we have $Y \models P$ and for all $Y' \subsetneq Y$ with $Y' \models f P^Y$ we have $Y'|_{\mathcal{H}} = Y|_{\mathcal{H}}$,
					which implies that $(Y,Y) \in \sigma_{\langle \mathcal{H}, \mathcal{B} \rangle}(P)$.
					
					If $(Y,Y) \not\in \sigma_{\langle \mathcal{H}, \mathcal{B} \rangle}(Q)$, it is proven that $\sigma_{\langle \mathcal{H}, \mathcal{B} \rangle}(P) \not= \sigma_{\langle \mathcal{H}, \mathcal{B} \rangle}(Q)$.
					
					Otherwise we have $(Y,Y) \in \sigma_{\langle \mathcal{H}, \mathcal{B} \rangle}(Q)$.
					By Property~(i) of Definition~\ref{def:witness} we then have $Y \models Q$ and by Lemma~\ref{lem:semodels}
					we have $X|_{\mathcal{H}} \not= Y|_{\mathcal{H}}$ and thus $X \subsetneq Y$.
					Since $(X,Y)$ is a witness for $P \not\subseteq_{\langle \mathcal{H}, \mathcal{B} \rangle} Q$ we have $X \models f Q^Y$
					and for all $X'$ with $X \leq^{\mathcal{B}}_{\mathcal{H}} X' \subsetneq Y$ we have $X' \not\models f P^Y$.
					Take an arbitrary pair $(Z,Y)$ of assignments with $Z \subsetneq Y$ for which $X \leq^{\mathcal{B}}_{\mathcal{H}} Z$ holds and which is $\leq^{\mathcal{B}}_{\mathcal{H}}$-maximal for $Q$
					(such a pair exists because we already know that $X \models f Q^Y$).
					Moreover, $(Y,Y) \in \sigma_{\langle \mathcal{H}, \mathcal{B} \rangle}(Q)$ implies that Property~(i) of Definition~\ref{def:hbmodel} holds for $(Z|_{\mathcal{H} \cup \mathcal{B}},Y)$.
					Therefore $(Z|_{\mathcal{H} \cup \mathcal{B}},Y) \in \sigma_{\langle \mathcal{H}, \mathcal{B} \rangle}(Q)$.
					
					On the other hand, $(Z|_{\mathcal{H} \cup \mathcal{B}},Y) \not\in \sigma_{\langle \mathcal{H}, \mathcal{B} \rangle}(P)$
					because $(X,Y)$ is a witness for $P \not\subseteq_{\langle \mathcal{H}, \mathcal{B} \rangle} Q$ and therefore for all $X'$ with $X \leq^{\mathcal{B}}_{\mathcal{H}} X' \subsetneq Y$
					we have $X' \not\models f P^Y$. Since $X \leq^{\mathcal{B}}_{\mathcal{H}} Z \subsetneq Y$ we also have that $X'' \not\models f P^Y$ for all $X''$
					such that $Z|_{\mathcal{H} \cup \mathcal{B}} \leq^{\mathcal{B}}_{\mathcal{H}} X'' \subsetneq Y$.
					But then there cannot be an $X''$ with $X''|_{\mathcal{H} \cup \mathcal{B}} = Z|_{\mathcal{H} \cup \mathcal{B}}$ such that $X'' \models f P^Y$.
					Therefore Property~(ii) of Definition~\ref{def:hbmodel} cannot be satisfied due to failure to find a pair $(X'',Y)$ with $X'' \subsetneq Y$ and
					$X''|_{\mathcal{H} \cup \mathcal{B}} = Z|_{\mathcal{H} \cup \mathcal{B}}$
					that is $\leq^{\mathcal{B}}_{\mathcal{H}}$-maximal for $P$.
			}

			We demonstrate the lifted results using three examples.
			
			\begin{example}
				Consider the programs $P = \{ a \leftarrow \ext{\mathit{aOrNotB}}{a,b}{} \}$ and $Q = \{ a \leftarrow a; \ a \leftarrow \naf b \}$ where $\ext{\mathit{aOrNotB}}{a,b}{}$ evaluates to true whenever $a$ is true
				or $b$ is false, and to false otherwise.
				Let $\mathcal{H} = \mathcal{B} = \{ a, b \}$.
				We have that $\sigma_{\langle \mathcal{H}, \mathcal{B} \rangle}(P) = \sigma_{\langle \mathcal{H}, \mathcal{B} \rangle}(Q) = \{ (\emptyset, \{\rev{a}{b}\}), (\{a\}, \{a\}), (\{b\}, \{b\}),$ $(\{a\},\{a,b\}), (\{b\},\{a,b\}), (\{a,b\}, \{a,b\}) \}$,
				and thus $P$ and $Q$ are $\langle \mathcal{H}, \mathcal{B} \rangle$-equivalent.
				
				It is easy to see that for any of the $\langle \mathcal{H}, \mathcal{B} \rangle$-models of form $(Y,Y)$,
				$Y$ is a model both of $P$ and $Q$, and for any $Y' \not\subseteq Y$ we have $Y'|_{\mathcal{H}} \subsetneq Y|_{\mathcal{H}}$;
				for the fourth candidate $(\emptyset, \emptyset)$ one can observe that $\emptyset$ is neither a model of $P$ nor of $Q$.
				
				For the $\langle \mathcal{H}, \mathcal{B} \rangle$-models of form $(\{a\},\{a,b\})$ resp.~$(\{b\},\{a,b\})$,
				one can observe that $X' = \{a\}$ resp.~$X' = \{b\}$ satisfies Condition~(ii) of Definition~\ref{def:hbmodel} both for $P$ and $Q$,
				while for $(\emptyset,\{a,b\})$ the only candidate for $X' \subsetneq \{a,b\}$ with $X'|_{\mathcal{H} \cup \mathcal{B}} = X$
				is $X' = \emptyset$, but $(\emptyset, \{a,b\})$ is neither $\leq^{\mathcal{B}}_{\mathcal{H}}$-maximal for $P$ nor for $Q$
				because $\emptyset \not\models f P^{\{a,b\}}$ and $\emptyset \not\models f Q^{\{a,b\}}$.

				For unary $Y$, the only $\langle \mathcal{H}, \mathcal{B} \rangle$-model $(X,Y)$ with $X \not= Y$ of $P$ or $Q$
				is $(\emptyset, \{b\})$ because for $X' = \emptyset$ we have $\emptyset \models f P^{\{b\}}$ and $\emptyset \models f Q^{\{\rev{a}{b}\}}$, and $(\emptyset, \{b\})$ is also $\leq^{\mathcal{B}}_{\mathcal{H}}$-maximal for $P$ and for $Q$.
				On the other hand, $(\emptyset, \{a\})$ fails to be an $\langle \mathcal{H}, \mathcal{B} \rangle$-model because the only candidate for
				$X'$ is $\emptyset$, but $\emptyset \not\models f P^{\{a\}}$ and $\emptyset \not\models f Q^{\{a\}}$.
			\end{example}
			
			\begin{example}
				\label{ex:hbmodel}
				Consider the programs $P = \{ a \leftarrow \ext{\mathit{neg}}{b}{}; \ b \leftarrow \ext{\mathit{neg}}{a}{}; \ a \leftarrow b \}$
				and $Q = \{ a \vee b \reva{\leftarrow}; \ a \leftarrow b \}$ where $\ext{\mathit{neg}}{x}{}$ evaluates to true whenever $x$ is false and to true otherwise.
				
				For $\mathcal{H} = \{ a, b \}$ and $\mathcal{B} = \{ b \}$
				we have that $\sigma_{\langle \mathcal{H}, \mathcal{B} \rangle}(P) = \sigma_{\langle \mathcal{H}, \mathcal{B} \rangle}(Q) = \{ (\{a\}, \{a\}),$ $(\{a\}, \{a,b\}), (\{a,b\}, \{a,b\}) \}$,
				and thus the programs are $\langle \mathcal{H}, \mathcal{B} \rangle$-equivalent.
				The most interesting candidate which fails to be an $\langle \mathcal{H}, \mathcal{B} \rangle$-model of either progam is $(\emptyset, \{a,b\})$.
				For $P$ we have that $f P^{\{a,b\}} = \{ a \leftarrow b \}$, of which $\emptyset$ is a model,
				but for $\{a\}$ we have $\emptyset \leq^{\mathcal{B}}_{\mathcal{H}} \{a\} \subsetneq Y$ and $\{a\} \models f P^{\{a,b\}}$, thus $\emptyset$ is not $\leq^{\mathcal{B}}_{\mathcal{H}}$-maximal for $P$\rev{. For}{; for}
				$Q$ we have that $f Q^{\{a,b\}} = \{ a \vee b; a \leftarrow b \}$, which is unsatisfied under $\emptyset$.
			\end{example}

			\begin{example}
				Consider the programs $P$ and $Q$ from Example~\ref{ex:hbmodel} and $\mathcal{H} = \{ a, b \}$ and $\mathcal{B} = \{ a, b \}$.
				We have that $\sigma_{\langle \mathcal{H}, \mathcal{B} \rangle}(P) = \{ (\{a\}, \{a\}), (\emptyset, \{a,b\}), (\{a\}, \{a,b\}), (\{a,b\}, \{a,b\}) \}$.
				Note that $(\emptyset, \{a,b\})$ is now an $\langle \mathcal{H}, \mathcal{B} \rangle$-model of $P$
				because $\emptyset$ is a model of $f P^{\{a,b\}} = \{ a \leftarrow b \}$ and there is no $X'$ with $\emptyset \leq^{\mathcal{B}}_{\mathcal{H}} X' \subsetneq Y$ with $X' \models f P^{\{ a \leftarrow b \}}$
				(because now $\emptyset \not\leq^{\mathcal{B}}_{\mathcal{H}} \{a\}$); thus $\emptyset$ is $\leq^{\mathcal{B}}_{\mathcal{H}}$-maximal for $P$.
				On the other hand, $\sigma_{\langle \mathcal{H}, \mathcal{B} \rangle}(Q) = \{ (\{a\}, \{a\}), (\{a\}, \{a,b\}), (\{a,b\},$ $\{a,b\}) \}$.
				That is, $(\emptyset, \{a,b\})$ is still not an $\langle \mathcal{H}, \mathcal{B} \rangle$-model of $Q$
				because $\emptyset$ is not a model of $f Q^{\{a,b\}} = \{ a \vee b \reva{\leftarrow}; \ a \leftarrow b \}$.
				And thus the programs are not $\langle \mathcal{H}, \mathcal{B} \rangle$-equivalent.
				
				Indeed, for $R = \{ b \leftarrow a \} \in \mathcal{P}_{\langle \mathcal{H}, \mathcal{B} \rangle}$
				we have that $Y = \{ a, b \}$ is an answer set of $Q \cup R$ but not of $P \cup R$.
			\end{example}

		\subsection{Adding General \hex-Programs}
		\label{sec:equivalence:generalhex}

			Up to this point we allowed only the addition of ordinary ASP-programs $R \in \mathcal{P}_{\langle \mathcal{H}, \mathcal{B} \rangle}$.
			As a preparation for the addition of general \hex-programs, we show now that if programs $P$ and $Q$ are $\langle \mathcal{H}, \mathcal{B} \rangle$-equivalent,
			then sets $\mathcal{B}$ and $\mathcal{H}$ can be extended
			by atoms that do not appear in $P$ and $Q$
			and the programs are still equivalent wrt.~the expanded sets.
			Intuitively, this allows introducing auxiliary atoms without harming their equivalence.
			This possibility is needed for our extension\revam{s} of the results to the case where $R$ can be a general \hex-program.

			\leanparagraph{Expanding Sets $\mathcal{B}$ and $\mathcal{H}$.}
			If programs $P$ and $Q$ are $\langle \mathcal{H}, \mathcal{B} \rangle$-equivalent,
			then they are also $\langle \mathcal{H}', \mathcal{B}' \rangle$-equivalent whenever $\mathcal{H}' \setminus \mathcal{H}$ and $\mathcal{B}' \setminus \mathcal{B}$
			contain only atoms \rev{which}{that} do not appear in $P$ or $Q$. This is intuitively the case because such atoms cannot interfere with atoms \rev{which}{that} are already in the program.
			
			Formally, one can show the following result:

			\addProposition{prop:hbextension}{
				For sets $\mathcal{H}$ and $\mathcal{B}$ of atoms, \hex-programs $P$ and $Q$, and an atom $a$ \rev{which}{that} does not occur in $P$ or $Q$,
				the following holds:
				\begin{enumerate}[(i)]
					\item \label{prop:hextension:i} $P \equiv_{\langle \mathcal{H}, \mathcal{B} \rangle} Q$ iff $P \equiv_{\langle \mathcal{H} \cup \{ a \}, \mathcal{B} \rangle} Q$; and
					\item \label{prop:hextension:ii} $P \equiv_{\langle \mathcal{H}, \mathcal{B} \rangle} Q$ iff $P \equiv_{\langle \mathcal{H}, \mathcal{B} \cup \{ a \} \rangle} Q$.
				\end{enumerate}
			}

			\addProof{prop:hbextension}{
				Property~(\ref{prop:hextension:i}) ($\Rightarrow$)
					We make a proof by contraposition. If $P \equiv_{\langle \mathcal{H} \cup \{ a \}, \mathcal{B} \rangle} Q$ does not hold,
					then either $P \subseteq_{\langle \mathcal{H} \cup \{ a \}, \mathcal{B} \rangle} Q$ or $Q \subseteq_{\langle \mathcal{H} \cup \{ a \}, \mathcal{B} \rangle} P$;
					as the two cases are symmetric it suffices to consider the former.
					If $P \subseteq_{\langle \mathcal{H} \cup \{ a \}, \mathcal{B} \rangle} Q$ does not hold then by Proposition~\ref{prop:containment}
					there is a witness $(X,Y)$ for
					$P \not\subseteq_{\langle \mathcal{H} \cup \{ a \}, \mathcal{B} \rangle} Q$.
					We show that we can also construct a witness for $P \not\subseteq_{\langle \mathcal{H}, \mathcal{B} \rangle} Q$, which implies by
					another application of Proposition~\ref{prop:containment} that $P \subseteq_{\langle \mathcal{H}, \mathcal{B} \rangle} Q$
					and thus $P \equiv_{\langle \mathcal{H}, \mathcal{B} \rangle} Q$ do not hold.

					In particular, $(X \setminus \{ a \}, Y \setminus \{ a \})$ is a witness for $P \not\subseteq_{\langle \mathcal{H}, \mathcal{B} \rangle} Q$.
					We show this separately depending on the type of $(X,Y)$.
					
					\begin{itemize}
						\item If neither $X$ nor $Y$ contains $a$, then $(X,Y)$ itself is also a witness for $P \not\subseteq_{\langle \mathcal{H}, \mathcal{B} \rangle} Q$.
							Property~(i) of Definition~\ref{def:witness} holds because we know that $Y \models P$ and for each $Y' \subsetneq Y$ with $Y' \models f P^Y$
							we have that $Y'|_{\mathcal{H} \cup \{ a \}} \subsetneq Y|_{\mathcal{H} \cup \{ a \}}$; the latter implies $Y'|_{\mathcal{H}} \subsetneq Y|_{\mathcal{H}}$
							since $a \not\in Y$ and thus $Y'$ and $Y$ must differ in an atom from $\mathcal{H}$.
							
							For Property~(ii), if $Y \not\models Q$ we are done. Otherwise we know that $X \subsetneq Y$ and $X \models f Q^Y$
							and that for all $X'$ with $X \leq^{\mathcal{B}}_{\mathcal{H} \cup \{ a \}} X'$ we have $X' \not\models f P^Y$
							(since $(X,Y)$ satisfies Property~(ii) wrt.~$\mathcal{H} \cup \{ a \}$ and $\mathcal{B}$).
							We have to show that $X' \not\models f P^Y$ holds also for all $X'$ with $X \leq^{\mathcal{B}}_{\mathcal{H}} X' \subsetneq Y$.
							However, each $X'$ such that $X \leq^{\mathcal{B}}_{\mathcal{H}} X'$ has to satisfy
							$X'|_{\mathcal{H}} \supseteq X|_{\mathcal{H}}$ and $X'|_{\mathcal{B}} \subseteq X|_{\mathcal{B}}$;
							the former implies $X'|_{\mathcal{H} \cup \{ a \}} \supseteq X|_{\mathcal{H} \cup \{ a \}}$ because $a \not\in Y$, $X \subsetneq Y$ and $X' \subsetneq Y$.
							Then for this $X'$ also $X \leq^{\mathcal{B}}_{\mathcal{H} \cup \{ a \}} X'$ holds, and therefore
							satisfaction of Property~(ii) wrt.~$\mathcal{H} \cup \{ a \}$ and $\mathcal{B}$ implies $X' \not\models f P^Y$.
							Thus Property~(ii) holds also wrt.~$\mathcal{H}$ and $\mathcal{B}$.

						\item If only $Y$ but not $X$ contains $a$, then $(X,Y \setminus \{ a \})$ is also a witness for $P \not\subseteq_{\langle \mathcal{H}, \mathcal{B} \rangle} Q$.
							For Property~(i), $Y \models P$ implies $Y \setminus \{ a \} \models P$ because $a$ does not occur in $P$.
							Now suppose there is a $Y' \subsetneq Y \setminus \{ a \}$ such that $Y' \models f P^Y$ and $Y'|_{\mathcal{H}} = Y|_{\mathcal{H}}$.
							Then $Y'$ and $Y$ differ in an atom other than $a$ and we have that $Y' \cup \{ a \} \subsetneq Y$
							and $Y' \cup \{ a \}|_{\mathcal{H} \cup \{ a \}} = Y|_{\mathcal{H} \cup \{ a \}}$; this contradicts the assumption that Property~(i) holds wrt.~$\mathcal{H} \cup \{ a \}$ and $\mathcal{B}$.
							
							For Property~(ii), if $Y \not\models Q$ then also $Y \setminus \{ a \} \not\models Q$ because $a$ does not occur in $Q$ and we are done.
							Otherwise we know that $X \subsetneq Y$ and $X \models f Q^Y$
							and that for all $X'$ with $X \leq^{\mathcal{B}}_{\mathcal{H} \cup \{ a \}} X'$ we have $X' \not\models f P^Y$
							(since $(X,Y)$ satisfies Property~(ii) wrt.~$\mathcal{H} \cup \{ a \}$ and $\mathcal{B}$).
							In this case, $X$ and $Y$ must in fact differ in more atoms than just $a$: otherwise $Y \models P$ would imply $X \models f P^Y$ (because $a$ does not occur in $P$ and $f P^Y \subseteq P$);
							since $X \leq^{\mathcal{B}}_{\mathcal{H} \cup \{ a \}} X' \subsetneq Y$ for any $X'$ with $X'|_{\mathcal{H} \cup \mathcal{B}} = X|_{\mathcal{H} \cup \mathcal{B}}$
							this would contradict the assumption that Property~(ii) holds wrt.~$\mathcal{H} \cup \{ a \}$ and $\mathcal{B}$.
							But then $X \subsetneq Y \setminus \{ a \}$. Moreover, each $X'$ such that $X \leq^{\mathcal{B}}_{\mathcal{H}} X' \subsetneq Y \setminus \{ a \}$ has to satisfy
							$X'|_{\mathcal{H}} \supseteq X|_{\mathcal{H}}$ and $X'|_{\mathcal{B}} \subseteq X|_{\mathcal{B}}$;
							the former implies $X'|_{\mathcal{H} \cup \{ a \}} \supseteq X|_{\mathcal{H} \cup \{ a \}}$ because $a \not\in Y \setminus \{ a \}$, $X \subsetneq Y$ and $X' \subsetneq Y$.
							Then for this $X'$ also $X \leq^{\mathcal{B}}_{\mathcal{H} \cup \{ a \}} X' \subsetneq Y$ holds, and therefore
							satisfaction of Property~(ii) wrt.~$\mathcal{H} \cup \{ a \}$ and $\mathcal{B}$ implies $X' \not\models f P^Y$.
							Thus Property~(ii) holds also wrt.~$\mathcal{H}$ and $\mathcal{B}$.

						\item If both $X$ and $Y$ contain $a$, then $(X \setminus \{ a \},Y \setminus \{ a \})$ is also a witness for $P \not\subseteq_{\langle \mathcal{H}, \mathcal{B} \rangle} Q$.
							For Property~(i), $Y \models P$ implies $Y \setminus \{ a \} \models P$ because $a$ does not occur in $P$.
							Now suppose there is a $Y' \subsetneq Y \setminus \{ a \}$ such that $Y' \models f P^Y$ and $Y'|_{\mathcal{H}} = (Y \setminus \{ a \})|_{\mathcal{H}}$.
							Then $Y'$ and $Y \setminus \{ a \}$ differ in an atom other than $a$ and we have that $Y' \cup \{ a \} \subsetneq Y$, $Y' \cup \{ a \} \models f P^Y$ (since $a$ does not occur in $P$)
							and $Y' \cup \{ a \}|_{\mathcal{H} \cup \{ a \}} = Y|_{\mathcal{H} \cup \{ a \}}$; this contradicts the assumption that Property~(i) holds wrt.~$\mathcal{H} \cup \{ a \}$ and $\mathcal{B}$.
							
							For Property~(ii), if $Y \not\models Q$ then also $Y \setminus \{ a \} \not\models Q$ because $a$ does not occur in $Q$ and we are done.
							Otherwise we know that $X \subsetneq Y$ (and thus $X \setminus \{ a \} \subsetneq Y \setminus \{ a \}$) and $X \models f Q^Y$
							and that for all $X'$ with $X \leq^{\mathcal{B}}_{\mathcal{H} \cup \{ a \}} X' \subsetneq Y$ we have $X' \not\models f P^Y$
							(since $(X,Y)$ satisfies Property~(ii) wrt.~$\mathcal{H} \cup \{ a \}$ and $\mathcal{B}$).
							We have to show that $X' \not\models f P^Y$ holds also for all $X'$ with $X \setminus \{ a \} \leq^{\mathcal{B}}_{\mathcal{H}} X' \subsetneq Y \setminus \{ a \}$.
							Consider such an $X'$, then
							$X'|_{\mathcal{H}} \supseteq (X \setminus \{ a \})|_{\mathcal{H}}$, $X'|_{\mathcal{B}} \subseteq (X \setminus \{ a \})|_{\mathcal{B}}$, and $X' \subsetneq Y \setminus \{ a \}$.
							Now let $X'' = X' \cup \{ a \}$. 
							Then $X''|_{\mathcal{H} \cup \{ a \}} \supseteq X|_{\mathcal{H} \cup \{ a \}}$ because $a$ is added to $X''$
							and the superset relation is already known to hold for all other atoms from $\mathcal{H}$.
							Moreover, $X''|_{\mathcal{B}} \subseteq X|_{\mathcal{B}}$ still holds
							because $X'|_{\mathcal{B}} \subseteq X|_{\mathcal{B}}$ and the only element $a$ 
							added to $X''$ is also in $X$.
							Moreover, we still have $X'' \subsetneq Y$ because $a \in Y$ and $X'$ and $Y$ differ in at least one atom other than $a$ due to $X' \subsetneq Y \setminus \{ a \}$.
							These conditions together imply $X \leq^{\mathcal{B}}_{\mathcal{H} \cup \{ a \}} X'' \subsetneq Y$,
							and thus
							satisfaction of Property~(ii) wrt.~$\mathcal{H} \cup \{ a \}$ and $\mathcal{B}$ implies $X'' \not\models f P^Y$.
							Since $X''$ and $X'$ differ only in $a$, \rev{which}{that} does not appear in $f P^Y$, this further implies $X' \not\models f P^Y$.
							Hence Property~(ii) holds also wrt.~$\mathcal{H}$ and $\mathcal{B}$.
					\end{itemize}
					
				Property~(\ref{prop:hextension:i}) ($\Leftarrow$)
					Trivial because
					$P \equiv_{\langle \mathcal{H} \cup \{ a \}, \mathcal{B} \rangle} Q$
					is a stronger condition than
					$P \equiv_{\langle \mathcal{H}, \mathcal{B} \rangle} Q$
					since it allows a larger class of programs to be added.

			\bigskip
				Property~(\ref{prop:hextension:ii}) ($\Rightarrow$)
					We make a proof by contraposition. If $P \equiv_{\langle \mathcal{H}, \mathcal{B} \cup \{ a \} \rangle} Q$ does not hold,
					then either $P \subseteq_{\langle \mathcal{H}, \mathcal{B} \cup \{ a \} \rangle} Q$ or $Q \subseteq_{\langle \mathcal{H}, \mathcal{B} \cup \{ a \} \rangle} P$;
					as the two cases are symmetric it suffices to consider the former.
					If $P \subseteq_{\langle \mathcal{H}, \mathcal{B} \cup \{ a \} \rangle} Q$ does not hold then by Proposition~\ref{prop:containment}
					there is a witness $(X,Y)$ for
					$P \not\subseteq_{\langle \mathcal{H}, \mathcal{B} \cup \{ a \} \rangle} Q$.
					We show that we can also construct a witness for $P \not\subseteq_{\langle \mathcal{H}, \mathcal{B} \rangle} Q$, which implies by
					another application of Proposition~\ref{prop:containment} that $P \subseteq_{\langle \mathcal{H}, \mathcal{B} \rangle} Q$
					and thus $P \equiv_{\langle \mathcal{H}, \mathcal{B} \rangle} Q$ does not hold.
					
					We show in particular that $(X,Y)$ is also a witness for $P \not\subseteq_{\langle \mathcal{H}, \mathcal{B} \rangle} Q$.
					Property~(i) of Definition~\ref{def:witness} is also satisfied wrt.~$\mathcal{H}$ and $\mathcal{B}$ (instead of $\mathcal{H}$ and $\mathcal{B} \cup \{ a \}$)
					as this condition is independent of $\mathcal{B}$.

					If $Y \not\models Q$ then Property~(ii) is also satisfied and we are done.
					Otherwise we know, that $X \subsetneq Y$, $X \models f Q^Y$ and for all $X'$ with $X \leq^{\mathcal{B} \cup \{ a \}}_{\mathcal{H}} X' \subsetneq Y$ we have $X' \not\models f P^Y$.
					We have to show that $X' \not\models f P^Y$ holds also for all $X'$ with $X \leq^{\mathcal{B}}_{\mathcal{H}} X' \subsetneq Y$.
					Consider such an $X'$, then $X'|_{\mathcal{H}} \supseteq X|_{\mathcal{H}}$, $X'|_{\mathcal{B}} \subseteq X|_{\mathcal{B}}$ and $X' \subsetneq Y$.
					Now let $X'' = X' \setminus \{ a \}$ if $a \in X'$ and $a \not\in X$, and $X'' = X'$ otherwise.
					We have then $X''|_{\mathcal{B} \cup \{ a \}} \subseteq X|_{\mathcal{B} \cup \{ a \}}$ because $a$ is removed from $X''$ whenever it is not in $X$,
					and the subset relation is known for all other atoms from $\mathcal{B}$.
					Moreover, $X''|_{\mathcal{H}} \supseteq X|_{\mathcal{H}}$ still holds because $X'|_{\mathcal{H}} \supseteq X|_{\mathcal{H}}$ and the only element $a$ which might be missing in $X''$ compared to $X$
					is only removed if it is not in $X$ anyway.
					These conditions together imply $X \leq^{\mathcal{B} \cup \{ a \}}_{\mathcal{H}} X' \subsetneq Y$,
					and thus
					satisfaction of Property~(ii) wrt.~$\mathcal{H}$ and $\mathcal{B} \cup \{ a \}$ implies $X'' \not\models f P^Y$.
					Since $X''$ and $X'$ may only differ in $a$, which does not appear in $f P^Y$, this implies $X' \not\models f P^Y$.
					Hence Property~(ii) holds also wrt.~$\mathcal{H}$ and $\mathcal{B}$.
					
				Property~(\ref{prop:hextension:ii}) ($\Leftarrow$)
					Trivial because
					$P \equiv_{\langle \mathcal{H}, \mathcal{B} \cup \{ a \} \rangle} Q$
					is a stronger condition than
					$P \equiv_{\langle \mathcal{H}, \mathcal{B} \rangle} Q$
					since it allows a larger class of programs to be added.
			}

			\reva{The proof is done by contraposition. The main idea of the $(\Rightarrow)$-direction of (\ref{prop:hextension:i}) is to
			assume wlog.~that $P \not\subseteq_{\langle \mathcal{H} \cup \{ a \}, \mathcal{B} \rangle} Q$ and start with a witness thereof.
			One can then construct also a witness for $P \not\subseteq_{\langle \mathcal{H}, \mathcal{B} \rangle} Q$.
			The $(\Leftarrow)$-direction is trivial because $P \equiv_{\langle \mathcal{H} \cup \{ a \}, \mathcal{B} \rangle} Q$
			is a stronger condition than $P \equiv_{\langle \mathcal{H}, \mathcal{B} \rangle} Q$.
			The proof for (\ref{prop:hextension:ii}) is analogous.}

			By iterative applications of this result we get the desired result:
			
			\addCorollary{cor:extension}{
				\rev{
				For sets $\mathcal{H}$ and $\mathcal{B}$ of atoms, programs $P$ and $Q$, and an sets of atoms $\mathcal{H}'$ and $\mathcal{B}'$ which does not occur in $P$ or $Q$,
				}{
				Let $\mathcal{H}$, $\mathcal{B}$, $\mathcal{H}'$ and $\mathcal{B}'$ be sets of atoms and let $P$ and $Q$ be programs
				such that the atoms in $\mathcal{H}' \cup \mathcal{B}'$ do not occur in $P$ or $Q$. Then
				}we have $P \equiv_{\langle \mathcal{H}, \mathcal{B} \rangle} Q$ iff $P \equiv_{\langle \mathcal{H} \cup \mathcal{H}', \mathcal{B} \cup \mathcal{B}' \rangle} Q$.
			}
			
			\addProof{cor:extension}{
				The claim follows immediately by applying Proposition~\ref{prop:hbextension} iteratively to each element in $\mathcal{H}'$ resp.~$\mathcal{B}'$.
			}

			\leanparagraph{Addition of General \hex-Programs.}
			In the following,
			for sets $\mathcal{H}$, $\mathcal{B}$ of atoms we define the set
			$$\mathcal{P}^e_{\langle \mathcal{H}, \mathcal{B} \rangle} = \left\{ \text{\hex-program } P \bigm\vert \begin{array}{@{}l@{}}
																													H(P) \subseteq \mathcal{H}, B^{+}(P) \cup B^{-}(P) \subseteq \mathcal{B}, \\
																													\text{only } \mathcal{B} \text{ are input to external atoms }
																												\end{array} \right\}$$
			of general \hex-programs whose head atoms come only from $\mathcal{H}$ and whose body atoms and input atoms to external atoms come only from $\mathcal{B}$.\footnote{
			Input atoms to external atoms must also be in $\mathcal{B}$ as they appear in bodies of our rewriting by~Lemma~\ref{lem:inputReplacement} below.}
			We then extend Definition~\ref{def:hbequivalence} as follows.
			
			\begin{definition}
				\label{def:hbeequivalence}
				\hex-programs $P$ and $Q$ are \emph{e-equivalent wrt.~a pair $\langle \mathcal{H}, \mathcal{B} \rangle$ of sets of atoms},
				or $\langle \mathcal{H}, \mathcal{B} \rangle^e$-equivalent, denoted $P \equiv^e_{\langle \mathcal{H}, \mathcal{B} \rangle} Q$,
				if $\mathcal{AS}(P \cup R) = \mathcal{AS}(Q \cup R)$ for all $R \in \mathcal{P}^e_{\langle \mathcal{H}, \mathcal{B} \rangle}$.
			\end{definition}
			
			Towards a characterization of $\langle \mathcal{H}, \mathcal{B} \rangle^e$-equivalence,
			we make use of external atom inlining as by Definition~\ref{def:inlining}
			without changing the answer sets of a program, cf.~Proposition~\ref{prop:externalAtomInlining}.
			
			We start with a technical result which allows for renaming a predicate input parameter $p_i \in \vec{p}$
			of an external atom $e = \ext{g}{\vec{p}}{\vec{c}}$ in a program $P$ to a new predicate $q$ \rev{which}{that} does not occur in $P$.
			This allows us to rename predicates such that inlining does not introduce rules \rev{which}{that} derive atoms other than auxiliaries,
			which is advantageous in the following.

			The idea of the renaming is to add auxiliary rules \rev{which}{that} define $q$ such that its extension represents exactly
			the former atoms over $p_i$, i.e., each atom $p_i(\vec{d})$ is represented by $q(p_i, \vec{d})$.
			Then, external predicate $\amp{g}$ is replaced by a new $\amp{g}'$
			whose semantics is adopted to this encoding of the input atoms.

			For the formalization of this idea, let $\vec{p}|_{p_i \rightarrow q}$ be vector $\vec{p}$ after replacement of its $i$-th element $p_i$ by $q$.
			Moreover, for an assignment $Y$ let $Y^q = Y \cup  \{ p_i(\vec{d}) \mid q(p_i, \vec{d}) \in Y \}$
			be the extended assignment which `extracts' from each atom $q(p_i, \vec{d}) \in Y$ the original atom $p_i(\vec{d})$.
			\rev{Then one can show:}
			{One can then show that for any program $P$,
			renaming input predicates of an external atom does not change the semantics of $P$ (modulo auxiliary atoms):}
			
			\addLemma{lem:inputReplacement}{
				For an external atom $e = \ext{g}{\vec{p}}{\vec{c}}$ in program~$P$, $p_i \in \vec{p}$, a new predicate $q$,
				let $e' = \ext{g'}{\vec{p}|_{p_i \rightarrow q}}{\vec{c}}$ s.t.
				$\extsem{g'}{Y}{\vec{p}|_{p_i \rightarrow q}}{\vec{c}} = \extsem{g}{Y^q }{\vec{p}}{\vec{c}}$ for all assignments $Y$.
				
				For
				$P' = P|_{e \rightarrow e'} \cup \{ q(p_i, \vec{d}) \leftarrow p_i(\vec{d}) \mid p_i(\vec{d}) \in A(P) \}$,
				$\mathcal{AS}(P)$ and $\mathcal{AS}(P')$ coincide, modulo atoms $q(\cdot)$.
			}
			
			\addProof{lem:inputReplacement}{	
				($\Rightarrow$)
					For an answer set $Y$ of $P$ we show that $Y' = Y \cup \{ q(p_i, \vec{d}) \mid p_i(\vec{d}) \in Y \}$ is an answer set of $P'$.
					
					Since input parameter $q$ in $e'$ behaves like $p_i$ in $e$, $Y' \models q(p_i, \vec{d})$ iff $Y \models p_i(\vec{d})$ for all $p_i(\vec{d}) \in A(P)$ by construction,
					and $Y'$ satisfies all rules $r \in \{ q(p_i, \vec{d}) \leftarrow p_i(\vec{d}) \mid p_i(\vec{d}) \in A(P) \}$
					by construction, we have that $Y'$ is a model of $P'$.
					
					Now suppose towards a contradiction that there is a smaller model $Y_{<}' \subsetneq Y'$ of $f P'^{Y'}$ and let this model be subset-minimal.
					Then $Y_{<}' \setminus Y'$ must contain at least one atom other than over $q$
					because switching an atom $q(p_i, \vec{d})$ to false is only possible if the respective atom $p_i(\vec{d})$ is also switched to false, otherwise a rule
					$r \in \{ q(p_i, \vec{d}) \leftarrow p_i(\vec{d}) \mid p_i(\vec{d}) \in A(P) \}$ (which is contained in the reduct $f P'^{Y'}$ because $Y' \models B(r)$) would remain unsatisfied.
					But then for $Y_{<} = Y_{<}' \cap A(P)$ we have that $Y_{<} \subsetneq Y$.
					Now consider some $r \in f P^Y$: then there is a respective $r' \in f P'^{Y'}$ with $e'$ in place of $e$ and such that $Y_{<}' \models r'$.
					Observe that $p_i(\vec{d}) \in Y_{<}$ implies $q(p_i, \vec{d}) \in Y_{<}'$ (otherwise a rule in $f P'^{Y'}$ remains unsatisfied under $Y_{<}'$)
					and that $q(p_i, \vec{d}) \in Y_{<}'$ implies $p_i(\vec{d}) \in Y_{<}$ due to assumed subset-minimality of $Y_{<}'$ (there is no reason to set $q(p_i, \vec{d})$ to true if $p_i(\vec{d})$ is false).
					This gives in summary that $q(p_i, \vec{d}) \in Y_{<}'$ iff $p_i(\vec{d}) \in Y_{<}$ for all atoms $p_i(\vec{d}) \in A(P)$.
					But then we have also $Y_{<} \models r$ because the only possible difference between $r$ and $r'$ is that $r$ might contain $e$ while $r'$ contains $e'$,
					but since $q(p_i, \vec{d}) \in Y_{<}'$ iff $p_i(\vec{d})$ for all atoms $p_i(\vec{d}) \in A(P)$, we have that $Y_{<}' \models r'$ implies $Y_{<} \models r$.
					That is, $Y_{<} \subsetneq Y$ is a smaller model of $f P^Y$, which contradicts the assumption that $Y$ is an answer set.

				($\Leftarrow$)
					For an answer set $Y'$ of $P'$ we show that $Y = Y' \cap A(P)$ is an answer set of $P$.
					First observe that for any $p_i(\vec{d}) \in A(P)$ we have that $q(p_i, \vec{d}) \in Y'$ iff $p_i(\vec{d}) \in Y$:
					the if-direction follows from satisfaction of the rules in $P'$ under $Y'$, the only-if direction follows from subset-minimality of $Y'$.

					Then the external atoms $e$ in $P$ behave under $Y$ like the respective $e'$ in $P'$ under $Y'$, which implies that $Y \models P$.
					
					Now suppose towards a contradiction that there is a smaller model $Y_{<} \subsetneq Y$ of $f P^Y$.
					We show that then for $Y_{<}' = Y_{<} \cup \{ q(p_i, \vec{d}) \mid p_i(\vec{d}) \in Y_{<} \}$ we have $Y_{<}' \models f P'^{Y'}$.
					But this follows from the observation that $f P'^{Y'}$ consists only of (i) rules that correspond to rules in $f P^Y$ but with $e'$ in place of $e$,
					and (ii) the rule $q(p_i, \vec{d}) \leftarrow p_i(\vec{d})$ for all $p_i(\vec{d}) \in Y'$.
					Satisfaction of (i) follows from the fact that $Y \models e$ iff $Y' \models e'$, satisfaction of (ii) is given by construction of $Y_{<}'$.
					Moreover, we have that $Y_{<}' \subsetneq Y'$: we have $Y_{<} \subsetneq Y \subseteq Y'$ and all atoms $q(p_i, \vec{d})$ added to $Y_{<}$ are also in $Y'$
					because it satisfies the rule $q(p_i, \vec{d}) \leftarrow p_i(\vec{d}) \in P'$; properness of the subset-relation follows from $Y_{<} \subsetneq Y$.
					Therefore we have $Y_{<}' \subsetneq Y'$ and $Y_{<}' \models f P'^{Y'}$, which contradicts the assumption that $Y'$ is an answer set of $P'$.
			}

			We now come to the actual inlining.
			Observe that Definitions~\ref{def:inlining} and~\ref{def:neginlining} are \emph{modular}
			in the sense that inlining external atoms $E$ in a program $P$
			affects only the rules of $P$ containing some \revtwo{of}{external atom from} $E$
			and adds additional rules, but does not change the remaining rules
			\rev{}{(i.e., our transformation performs only changes that are `local' to rules that contain \revtwo{$R$}{some external atom from $E$})}.
			\revam{If $X \triangle Y = (X \setminus Y) \cup (Y \setminus X)$ denotes the symmetric difference between sets $X$ and $Y$,}
			\rev{o}{O}ne can formally show:
			
			\addLemma{lem:inliningmodularity}{
				For a \hex-program $P$ and a set of (positive or negative) external atoms $E$ in $P$,
				we have $P \rev{\triangle}{\cap} P_{[E]} = \{ r \in P \mid \text{none of } E \text{ occur in } r \}$.
			}
			
			\addProof{lem:inliningmodularity}{
				For a single external atom $e \in E$
				observe that all rules $r \in P_{[e]}$, which were constructed by (\ref{def:inlining:1})-(\ref{def:inlining:3}) in Definition~\ref{def:inlining},
				contain at least one atom \rev{which}{that} does not appear in $P$. Thus these rules can only be in $P_{[e]}$ but not in $P$ and thus not in $P \rev{\triangle}{\cap} P_{[e]}$.
				For the rules $r \in P_{[e]}$ constructed by (\ref{def:inlining:4}) in Definition~\ref{def:inlining},
				note that $r \in P$ \rev{iif}{iff} $e$ does not appear in $r$.
				This is further the case iff $r \in P \rev{\triangle}{\cap} P_{[e]}$.
				In summary, $P \rev{\triangle}{\cap} P_{[e]}$ contains all and only the rules from $P$ \rev{which}{that} do not contain $e$.
				
				By iteration of the argument, one gets the same result for the set $E$ of external atoms.
			}

			This equips us to turn to our main goal of characterizing equivalence of \hex-programs.
			If programs $P$ and $Q$ are $\langle \mathcal{H}, \mathcal{B} \rangle$-equivalent,
			then $P \cup R$ and $Q \cup R$ have the same answer sets for all ordinary ASP-programs $R \in \mathcal{P}_{\langle \mathcal{H}, \mathcal{B} \rangle}$.
			We will show that equivalence holds in fact even for
			\hex-programs $R \in \mathcal{P}^e_{\langle \mathcal{H}, \mathcal{B} \rangle}$.
			To this end, assume that $P$ and $Q$ are $\langle \mathcal{H}, \mathcal{B} \rangle$-equivalent for some $\mathcal{H}$ and $\mathcal{B}$
			and let $R \in \mathcal{P}^e_{\langle \mathcal{H}, \mathcal{B} \rangle}$.
			
			We \rev{apply a transformation to}{want to inline} all (positive or negative)
			\reva{occurrences of}
			external atoms \reva{from} $E$ in $P \cup R$ and $Q \cup R$ \rev{which}{that} appear in the $R$ part, \emph{but not \reva{the occurrences} in the $P$ part or $Q$ part}.
			\rev{To this end, observe that if the same external atom $e$ is shared between $P$ and $R$ resp.~$Q$ and $R$,
			the restriction of the inlining to the $R$ part is still possible by standardizing external atoms in $R$ apart from those in $P$ and $Q$, e.g.,
			by introducing a copy of the external predicate; we assume in the following that external atoms have been standardized apart as needed.}
			{However, since the application of the transformation as by Definition~\ref{def:inlining} to $P \cup R$ resp.~$Q \cup R$ would inline \emph{all} occurrences of $E$,
			we first have to standardize occurrences in $R$ apart from those in $P$ resp.~$Q$.
			This can be \revtwo{done done}{done} by introducing a copy of the external predicate; we assume in the following that external atoms have been standardized apart as needed,
			i.e., the external atoms $E$ appear only in $R$ but not in $P$ and $Q$.}
			\reva{Note that although \revtwo{}{external atoms from} $E$ appear only in program part $R$, the transformation is formally still applied to $P \cup R$ and $Q \cup R$ and not just to $R$.}
			The \reva{overall} transformation is then given as follows:
			\begin{enumerate}[(1)]
				\item rename their input parameters using Lemma~\ref{lem:inputReplacement}; and
				\item subsequently inline them by applying Definition~\ref{def:inlining} \reva{to $P \cup R$ and $Q \cup R$}.
			\end{enumerate}

			Note that neither of the two steps modifies \reva{the program parts} $P$ or $Q$:
			for (1) this is by construction of the modified program in Lemma~\ref{lem:inputReplacement},
			for (2) this follows from Lemma~\ref{lem:inliningmodularity}.
			\reva{
			Hence, what we get are programs of form $P \cup R'$ and $Q \cup R'$, where $R'$ consists of modified rules from $R$ and some auxiliary rules.
			}
			As observable from Lemma~\ref{lem:inputReplacement} and Definition~\ref{def:inlining}, head atoms $H(R')$ in $R'$ come either from $H(R)$ or are newly introduced auxiliary atoms;
			the renaming as by Lemma~\ref{lem:inputReplacement} prohibits that $H(R')$ contains input atoms to external atoms in $R$.
			Body atoms $B(R')$ in $R'$ come either from $B(R)$, from input atoms to external atoms in $R$ (see rules~(\ref{def:inlining:2})),
			or are newly introduced auxiliary atoms.
			Since $R \in \mathcal{P}^e_{\langle \mathcal{H}, \mathcal{B} \rangle}$, this implies that $H(R') \subseteq \mathcal{H} \cup \mathcal{H}'$ and $B(R') \subseteq \mathcal{B} \cup \mathcal{B}'$,
			where $\mathcal{H}'$ and $\mathcal{B}'$ are newly introduced auxiliary atoms.
			Since the auxiliary atoms do not occur in $P$ and $Q$, by Corollary~\ref{cor:extension} they do not harm equivalence, i.e., $\langle \mathcal{H}, \mathcal{B} \rangle$-equivalence
			implies $\langle \mathcal{H} \cup \mathcal{H}', \mathcal{B} \cup \mathcal{B}' \rangle$-equivalence.
			Thus, $\langle \mathcal{H}, \mathcal{B} \rangle$-equivalence of $P$ and $Q$ implies that $P \cup R'$ and $Q \cup R'$ have the same answer sets.			

			The claim follows then from the observation that, due to Lemma~\ref{lem:inputReplacement} and soundness and completeness of inlining (cf.~Proposition~\ref{prop:externalAtomInlining}),
			$P \cup R$ and $Q \cup R$ have the same answer sets whenever
			$P \cup R'$ and $Q \cup R'$ have the same answer sets.

			\begin{example}
				Consider the programs
				\begin{align*}
					P =& \{ a \leftarrow \ext{\mathit{neg}}{b}{}; \ b \leftarrow \ext{\mathit{neg}}{a}{}; \ a \leftarrow b \} \\
					Q =& \{ a \vee b \reva{\leftarrow}; \ a \leftarrow b \}
				\end{align*}
				and let $\mathcal{H} = \{a,c\}$ and $\mathcal{B} = \{b\}$.
				Note that $P \equiv_{\langle \mathcal{H}, \mathcal{B} \rangle} Q$.
				To observe this result, recall that we know $P \equiv_{\langle \{a,b\}, \{b\} \rangle} Q$ from Example~\ref{ex:hbmodel}, which implies $P \equiv_{\langle \{a\}, \{b\} \rangle} Q$.
				As $c \not\in A(P)$, $c \not\in A(Q)$, Proposition~\ref{prop:hbextension} further implies that $P \equiv_{\langle \{a,c\}, \{b\} \rangle} Q$.
				
				\medskip
				\noindent Let $R = \{ c \leftarrow \ext{\mathit{neg}}{b}{} \} \in \mathcal{P}^e_{\langle \mathcal{H}, \mathcal{B} \rangle}$.
				Renaming the input predicate of $\ext{\mathit{neg}}{b}{}$ by step (1) yields the program $\{ q(b) \leftarrow b; \ c \leftarrow \ext{\mathit{neg}'}{q}{} \}$.
				After step (2) we have:
				\begin{align*}
					R' = \{
							& q(b) \leftarrow b; \ c \leftarrow x_e; \ x_e \leftarrow \overline{q(b)}; \ \overline{x_e} \leftarrow \naf x_e \\
							& \overline{q(b)} \leftarrow  \naf q(b); \ \overline{q(b)} \leftarrow x_e; \ q(b) \vee \overline{q(b)} \leftarrow x_e \}
				\end{align*}
				Here, rule $q(b) \leftarrow b$ comes from step (1), $c \leftarrow x_e$ represents the rule in $R$, and the remaining rules from inlining in step (2).
				Except for new auxiliary atoms, we have that $H(R')$ use only atoms from $\mathcal{H}$ and $B(R')$ only atoms from $B(R')$.
				One can check that $P \cup R'$ and $Q \cup R'$ have the same (unique) answer set $\{ a, c, \rev{\overline{x_e}}{x_e}, \overline{q(b)} \}$,
				which corresponds to the (same) unique answer set $\{ a, c \}$ of $P \cup R$ and $Q \cup R$, respectively.
			\end{example}

			One can then show that 
			equivalence wrt.~program extensions that contain external atoms is characterized by the same criterion as extensions with ordinary ASP-programs only.
			
			\addProposition{prop:semodelshex}{
				For sets $\mathcal{H}$ and $\mathcal{B}$ of atoms and \hex-programs $P$ and $Q$,
				we have $P \equiv^e_{\langle \mathcal{H}, \mathcal{B} \rangle} Q$ iff $\sigma_{\langle \mathcal{H}, \mathcal{B} \rangle}(P) = \sigma_{\langle \mathcal{H}, \mathcal{B} \rangle}(Q)$.
			}

			\addProof{prop:semodelshex}{
				$(\Rightarrow)$
					If $\mathcal{AS}(P \cup R) = \mathcal{AS}(Q \cup R)$ for all $R \in \mathcal{P}^e_{\langle \mathcal{H}, \mathcal{B} \rangle}$,
					then this holds in particular for all programs $R \in \mathcal{P}_{\langle \mathcal{H}, \mathcal{B} \rangle}$ without external atoms.
					Then by Proposition~\ref{prop:semodels} we have $\sigma_{\langle \mathcal{H}, \mathcal{B} \rangle}(P) = \sigma_{\langle \mathcal{H}, \mathcal{B} \rangle}(Q)$.
				
				$(\Leftarrow)$
					Suppose $\sigma_{\langle \mathcal{H}, \mathcal{B} \rangle}(P) = \sigma_{\langle \mathcal{H}, \mathcal{B} \rangle}(Q)$, then
					by Proposition~\ref{prop:semodels} we have $P \equiv_{\langle \mathcal{H}, \mathcal{B} \rangle} Q$
					and by Corollary~\ref{cor:extension} we have $P \equiv_{\langle \mathcal{H} \cup \mathcal{H}', \mathcal{B} \cup \mathcal{B}' \rangle} Q$
					for all sets $\mathcal{H}'$, $\mathcal{B}'$ of atoms \rev{which}{that} do not occur in $P$ or $Q$.
					Now consider $R \in \mathcal{P}^e_{\langle \mathcal{H}, \mathcal{B} \rangle}$.
					We have to show that $\mathcal{AS}(P \cup R) = \mathcal{AS}(Q \cup R)$.
					
					Let $R'$ be the ordinary ASP-program after standardizing input atoms to external atoms apart from the atoms in $P$ and $Q$ (using Lemma~\ref{lem:inputReplacement})
					and subsequent inlining all external atoms in $R$ using
					Definition~\ref{def:inlining}. Note that $R'$ uses only atoms from $\mathcal{H}$ in its heads, atoms from $\mathcal{B}$ in its bodies, and \revam{and }newly introduced atoms $A(R') \setminus A(R)$;
					the latter are selected such that they do not occur in $P$ or $Q$.
					We further have that $R'$ is free of external atoms, thus $R' \in \mathcal{P}_{\langle \mathcal{H} \cup \mathcal{H}', \mathcal{B} \cup \mathcal{B}' \rangle}$
					for $\mathcal{H}' = \mathcal{B}' = A(R') \setminus A(R)$.
					
					We then have
					$P \equiv_{\langle \mathcal{H} \cup \mathcal{H}', \mathcal{B} \cup \mathcal{B}' \rangle} Q$
					(by Corollary~\ref{cor:extension}, as discussed above).
					By definition of $\equiv_{\langle \mathcal{H} \cup \mathcal{H}', \mathcal{B} \cup \mathcal{B}' \rangle}$
					this gives $\mathcal{AS}(P \cup R') = \mathcal{AS}(Q \cup R')$.
					Then by Proposition~\ref{prop:externalAtomInlining} we have that $\mathcal{AS}(P \cup R) = \mathcal{AS}(Q \cup R)$.
			}

			\reva{The idea of the proof is to reduce the problem to the case where $R$ is free of external atoms and apply Proposition~\ref{prop:semodels}.
			To this end, we inline the external atoms in $R$.
			This reduction is possible thanks to the fact that inlining introduces only auxiliary atoms that to not appear in $P$ and $Q$, which
			do not affect equivalence as stated by Corollary~\ref{cor:extension}.}
			
			\rev{Finally, f}{F}or the Herbrand base $\mathit{HB}_{\mathcal{C}}(P)$ of all atoms constructible from the predicates in $P$ and the constants $\mathcal{C}$,
			strong equivalence~\cite{Lifschitz:2001:SEL:383779.383783} corresponds to the special case of $\langle \mathit{HB}_{\mathcal{C}}(P), \mathit{HB}_{\mathcal{C}}(P) \rangle$-equivalence,
			and uniform equivalence~\cite{DBLP:conf/iclp/EiterF03} corresponds to $\langle \mathit{HB}_{\mathcal{C}}(P), \emptyset \rangle$-equivalence;
			this follows directly from definition of strong resp.~uniform equivalence.

	\section{Inconsistency of \hex-Programs}
	\label{sec:inconsistency}

		We turn now to inconsistency of \hex-programs.
		Similarly to equivalence, we want to characterize inconsistency wrt.~program extensions.
		\reva{Inconsistent programs are programs without answer sets.
		Observe that due to nonmonotonicity, inconsistent \hex-program can become consistent under program extensions.

		\begin{example}
			Consider the program $P = \{ p \leftarrow \ext{\mathit{neg}}{p}{} \}$,
			which resembles $P' = \{ p \leftarrow \naf p \}$ in ordinary ASP.
			The program is inconsistent because $Y_1 = \emptyset$ violates the (only) rule of the program,
			while $Y_2 = \{ p \}$ is not a minimal model of the reduct $f P^{Y_2} = \emptyset$.
			However, the extended program $P \cup \{ p \leftarrow \}$ has the answer set $Y_2$.
		\end{example}
		
		Some program extensions preserve inconsistency of a program,
		and it is a natural question \revtwo{is under}{under} which program extensions this is the case.
		}
		Akin to equivalence, sets $\mathcal{H}$ and $\mathcal{B}$
		constrain the atoms that may \revam{be }occur in rule heads, rule bodies and input atoms to external atoms of the added program, respectively.
		In contrast to equivalence, the criterion naturally concerns only a single program.
		However, we are still able to derive the criterion from the above results.
		
		\leanparagraph{Deriving a Criterion for Inconsistency.}
		We formalize our envisaged notion of inconsistency \reva{from above} as follows:
		
		\begin{definition}
			A \hex-program $P$ is called \emph{persistently inconsistent} wrt.~sets of atoms $\mathcal{H}$ and $\mathcal{B}$,
			if $P \cup R$ is inconsistent for all $R \in \mathcal{P}^e_{\langle \mathcal{H}, \mathcal{B} \rangle}$.
		\end{definition}
		
		\begin{example}
			\label{ex:inc1}
			The program $P = \{ p \leftarrow \ext{\mathit{neg}}{p}{} \}$
			is persistently inconsistent wrt.~all $\mathcal{H}$ and $\mathcal{B}$ such that $p \not\in \mathcal{H}$.
			This is because any model $Y$ of $P$, and thus of $P \cup R$ for some $R \in \mathcal{P}^e_{\langle \mathcal{H}, \mathcal{B} \rangle}$,
			must set $p$ to true due to the rule $p \leftarrow \ext{\mathit{neg}}{p}{}$.
			However, $Y \setminus \{ p \}$ is a model of $f (P \cup R)^Y$ if no rule in $R$ derives $p$, hence $Y$ is not a subset-minimal model of $f (P \cup R)^Y$.
		\end{example}
		
		\reva{We now want to characterize persistent inconsistency of a program wrt.~sets of atoms $\mathcal{H}$ and $\mathcal{B}$ in terms of a formal criterion.}
		We start deriving \rev{a}{the} criterion by observing that a program $P_{\bot}$ is
		persistently inconsistent wrt.~any $\mathcal{H}$ and $\mathcal{B}$ whenever it is classically inconsistent.
		Then $P_{\bot} \cup R$ does not even have classical models for any $R \in \mathcal{P}^e_{\langle \mathcal{H}, \mathcal{B} \rangle}$,
		and thus it cannot have answer sets.
		For such a $P_{\bot}$, another program $P$ is persistently inconsistent wrt.~$\mathcal{H}$ and $\mathcal{B}$
		iff it is $\langle \mathcal{H}, \mathcal{B} \rangle^e$-equivalent to $P_{\bot}$; the latter can by Proposition~\ref{prop:semodels} be checked by comparing their $\langle \mathcal{H}, \mathcal{B} \rangle$-models.
		This allows us to derive the desired criterion in fact as a special case of the one for equivalence.

		Classically inconsistent programs do not have $\langle \mathcal{H}, \mathcal{B} \rangle$-models due to violation of Property~(i) of Definition~\ref{def:hbmodel}.
		Therefore, checking for persistent inconsistency works by checking \rev{if}{whether} $P$ does not have $\langle \mathcal{H}, \mathcal{B} \rangle$-models either.
		To this end, it is necessary that each classical model $Y$ of $P$ violates Property~(i) of Definition~\ref{def:hbmodel}, otherwise $(Y,Y)$ (and possibly $(X,Y)$ for some $X \subsetneq Y$)
		would be $\langle \mathcal{H}, \mathcal{B} \rangle$-models of $P$. Formally:

		\addProposition{prop:inconsistency}{
			A \hex-program $P$ is persistently inconsistent wrt.~sets of atoms $\mathcal{H}$ and $\mathcal{B}$
			iff for each classical model $Y$ of $P$ there is an $Y' \subsetneq Y$ such that $Y' \models f P^Y$ and $Y'|_{\mathcal{H}} = Y|_{\mathcal{H}}$.
		}

		\addProof{prop:inconsistency}{
			Let $P_{\bot}$ be a program without classical models (e.g., $\{ a \leftarrow; \ \leftarrow a \}$).
			Then, by monotonicity of classical logic, $P_{\bot} \cup R$ is inconsistent (wrt.~the \hex-semantics) for all $R \in \mathcal{P}^e_{\langle \mathcal{H}, \mathcal{B} \rangle}$, i.e.,
			we have that $\mathcal{AS}(P_{\bot} \cup R) = \emptyset$ for all $R \in \mathcal{P}^e_{\langle \mathcal{H}, \mathcal{B} \rangle}$.
			
			We have to show that $\mathcal{AS}(P \cup R) = \emptyset$ for all $R \in \mathcal{P}^e_{\langle \mathcal{H}, \mathcal{B} \rangle}$
			iff for each model $Y$ of $P$ there is an $Y' \subsetneq Y$ such that $Y' \models f P^Y$ and $Y'|_{\mathcal{H}} = Y|_{\mathcal{H}}$.
			Due to Proposition~\ref{prop:externalAtomInlining}, each program with external atoms may be replaced by an ordinary ASP-program such that the answer sets correspond to each other one-by-one;
			therefore the former statement holds iff $\mathcal{AS}(P \cup R) = \emptyset$ for all $R \in \mathcal{P}_{\langle \mathcal{H}, \mathcal{B} \rangle}$,
			i.e., it suffices to consider ordinary ASP-programs $R$.
			The claim is proven if we can show that
			$\mathcal{AS}(P \cup R) \subseteq \mathcal{AS}(P_{\bot} \cup R)$ for all $R \in \mathcal{P}_{\langle \mathcal{H}, \mathcal{B} \rangle}$.
			
			This corresponds to deciding $P \subseteq_{\langle \mathcal{H}, \mathcal{B} \rangle} P_{\bot}$.		
			By Proposition~\ref{prop:containment}, $P \subseteq_{\langle \mathcal{H}, \mathcal{B} \rangle} P_{\bot}$
			is the case iff \emph{no} witness for $P \not\subseteq_{\langle \mathcal{H}, \mathcal{B} \rangle} P_{\bot}$ exists.
			Since $P_{\bot}$ does not have any classical models, each pair $(X,Y)$ of assignments trivially satisfies Condition~(ii) because $Y \not\models P_{\bot}$,
			thus a pair $(X,Y)$ is not a witness iff it violates Property~(i).
			This condition is violated by $(X,Y)$ iff $Y \not\models P$ or there exists a $Y' \subsetneq Y$ such that $Y' \models f P^Y$ and $Y'|_{\mathcal{H}} = Y|_{\mathcal{H}}$;
			this is exactly the stated condition.
		}

		\begin{example}[cont'd]
			\label{ex:inc2}
			For the program $P$ from Example~\ref{ex:inc1}
			we have \rev{for each classical model $Y \supseteq \{ p \}$ holds}{that $Y \supseteq \{ p \}$ holds for each classical model $Y$ of $P$}.
			However, for each such $Y$ we have that $Y' = Y \setminus \{ p \}$ is a model of $f P^Y$, $Y' \subsetneq Y$ and $Y|_{\mathcal{H}} = Y'|_{\mathcal{H}}$,
			which proves that $P \cup R$ is inconsistent for all $R \in \mathcal{P}_{\langle \mathcal{H}, \mathcal{B} \rangle}$.
		\end{example}
		
		\begin{example}
			\label{ex:inc3}
			Consider the program $P = \{ a \leftarrow \ext{\mathit{aOrNotB}}{a,b}{}; \ \leftarrow a \}$.
			It is persistently inconsistent wrt.~all $\mathcal{H}$ and $\mathcal{B}$ such that $b \not\in \mathcal{H}$.
			This is the case because the rule $a \leftarrow \ext{\mathit{aOrNotB}}{a,b}{}$ derives $a$ whenever $b$ is false, which violates the constraint $\leftarrow a$.
			Formally, one can observe that we have $a \not\in Y$ and $b \in Y$ for each classical model $Y$ of $P$.
			But then $Y' = Y \setminus \{ b \}$ is a model of $f P^Y$, $Y' \subsetneq Y$ and $Y|_{\mathcal{H}} = Y'|_{\mathcal{H}}$.
		\end{example}

		The criterion for inconsistency follows therefore as a special case from the criterion for program equivalence.
		
		\leanparagraph{Applying the Criterion using Unfounded Sets.}
		Proposition~\ref{prop:inconsistency} formalizes a condition for deciding persistent inconsistency based on models of the program's reduct.
		However, practical implementations usually do not explicitly generate the reduct, but are often based on \emph{unfounded sets}~\cite{faber2005-lpnmr}.
		For a model $Y$ of a program $P$, smaller models $Y' \subsetneq Y$ of the reduct $f P^Y$ and unfounded sets of $P$ wrt.~$Y$ correspond to each other one-by-one.
		This allows us to transform the above decision criterion such that it can be directly checked using unfounded sets.
		
		We use unfounded sets for logic programs as introduced 
		by~\citeN{faber2005-lpnmr}
		for programs with arbitrary aggregates.

		\begin{definition}[Unfounded Set]
			\label{def:unfoundedset}
			Given a program $P$ and an assignment $Y$, let $U$
			be any set of atoms appearing in $P$. Then\rev{,}{} 
			$U$ is an \emph{unfounded set for $P$ wrt.~$Y$} if, for each $r \in P$ with $H(r) \cap U \not= \emptyset$,
			at least one of the following holds:
			\begin{enumerate}[(i)]
				\item\label{def:unfoundedset:i} some literal of $B(r)$ is false wrt.~$Y$; or
				\item\label{def:unfoundedset:ii} some literal of $B(r)$ is false wrt.~$Y \setminus U$; or
				\item\label{def:unfoundedset:iii} some atom of $H(r) \setminus U$ is true wrt.~$Y$.
			\end{enumerate}
		\end{definition}
		
		
		\addLemma{lem:ufsSmallerModelCorrespondence}{
			For a \hex-program $P$ and a model $Y$ of $P$,
			a set of atoms $U$ is an unfounded set of $P$ wrt.~$Y$ iff $Y \setminus U \models f P^Y$.
		}
		
		\addProof{lem:ufsSmallerModelCorrespondence}{
			($\Rightarrow$)
				We have to show that any rule $r \in f P^Y$ is satisfied under $Y \setminus U$.
				First observe that $Y \models H(r)$ because otherwise we also had $Y \not\models B(r)$ (since $Y$ is a model of $P$) and thus $r \not\in f P^Y$.
				If $Y \setminus U \models H(r)$ we are done ($Y \setminus U \models r$). Otherwise we have $H(r) \cap U \not=\emptyset$ and thus one of the conditions of Definition~\ref{def:unfoundedset} holds for $r$.
				This cannot be Condition~(i) because otherwise we had $r \not\in f P^Y$. If it is Condition~(ii) then $Y \setminus U \not\models B(r)$ and thus $Y \setminus U \models r$.
				If it is Condition~(iii) then $Y \setminus U \models H(r)$ and thus $Y \setminus U \models r$.
				
			($\Leftarrow$)
				Let $Y' \subseteq Y$ be a model of $f P^Y$.
				We have to show that $U = Y \setminus Y'$ is a unfounded set of $P$ wrt.~$Y$.
				To this end we need to show that for all $r \in P$ with $H(r) \cap U \not= \emptyset$ one of the conditions of Definition~\ref{def:unfoundedset} holds.
				If $r \not\in f P^Y$ then $Y \not\models B(r)$ and thus Condition~(i) holds.
				If $r \in f P^Y$ then we either have $Y' \not\models B(r)$ or $Y' \models H(r)$.
				If $Y' \not\models B(r)$ then $Y \setminus U \not\models B(r)$ because $Y \setminus U = Y'$, i.e., Condition~(ii) holds.
				If $Y' \models H(r)$ then there is an $h \in H(r)$ s.t.~$h \in Y$ and $h \in Y'$ and thus $h \not\in U$. Then we have $h \in Y \setminus U$ and thus $Y \models h$, i.e., Condition~(iii) holds.
		}
		
		\reva{The lemma is shown for all rules of the program ony-by-one.}
		By contraposition, the lemma implies that for a model $Y$ of $P$ and a model $Y' \subseteq Y$ of $f P^Y$ we have that $Y \setminus Y'$ is an unfounded set of $P$ wrt.~$Y$.
		This allows us to restate our decision criterion as follows:
		
		\addCorollary{cor:ufsInconsistency}{
			A \hex-program $P$ is persistently inconsistent wrt.~sets of atoms $\mathcal{H}$ and $\mathcal{B}$ 
			iff for each classical model $Y$ of $P$ there is a nonempty unfounded set $U$ of $P$ wrt.~$Y$ s.t.~$U \cap Y \not= \emptyset$ and $U \cap \mathcal{H} = \emptyset$.
		}
		
		\addProof{cor:ufsInconsistency}{
			By Proposition~\ref{prop:inconsistency} we know that
			$P \cup R$ is inconsistent for all $R \in \mathcal{P}_{\langle \mathcal{H}, \mathcal{B} \rangle}$
			iff for each model $Y$ of $P$ there is an $Y' \subsetneq Y$ such that $Y' \models f P^Y$ and $Y'|_{\mathcal{H}} = Y|_{\mathcal{H}}$.
			Each such model $Y'$ corresponds one-by-one to a nonempty unfounded set $U = Y \setminus Y'$ of $P$ wrt.~$Y$,
			for \rev{which}{that} we obviously have $U \cap Y \not= \emptyset$ and $Y'|_{\mathcal{H}} = Y|_{\mathcal{H}}$ iff $U \cap \mathcal{H} = \emptyset$.
		}

		\begin{example}[cont'd]
			\label{ex:inc4}
			For the program $P$ from Example~\ref{ex:inc3}
			we have that $U = \{ b \}$ is an unfounded set of $P$ wrt.~any classical model $Y$ of $P$;
			by assumption $b \not\in \mathcal{H}$ we have $U \cap \mathcal{H} = \emptyset$.
		\end{example}

		\leanparagraph{Application.}
		We now want to discuss a specific use-case of the decision criterion for program inconsistency.
		However, we stress that this section focuses on the study of the criterion, which is interesting by itself,
		while a detailed realization of the application is beyond its scope and discussed in more detail by~\citeN{r2017-ijcai}.

		The state-of-the-art evaluation approach for \hex-programs makes use of \emph{program splitting} for handling programs with variables.
		That is, the overall program is partitioned into components \rev{which}{that}
		are arranged in an \emph{acyclic graph}.
		Then, beginning from the components without predecessors, each component is separately grounded and solved,
		and each answer set is one-by-one added as facts to the successor components. The process is repeated in a recursive manner
		such that eventually the leaf components will yield the final answer sets, cf.~\citeN{efikrs2015-tplp}.

		\rev{The main reason for program splitting is that \emph{value invention}, i.e., the introduction of constants by external sources which do not occur in the input program,
		may lead to a grounding bottleneck if evaluated as monolithic program.}
		{The main reason for program splitting is \emph{value invention}, which is supported by non-ground \hex-programs, i.e.,
		the introduction of constants by external sources that do not occur in the input program.
		\revtwo{Since determining for a given program the set of relevant constants is computationally expensive in general,
		this may lead to a grounding bottleneck if evaluated as monolithic program.}{In general, determining
		the set of relevant constants is computationally expensive. This may lead to a grounding bottleneck if evaluated as monolithic program.}}
		This is because the grounder needs to evaluate external atoms under all possible inputs in order to ensure that
		all possible outputs are respected in the grounding, as demonstrated by the following example.

		\begin{example}
			\label{ex:graph}
			Consider the program
			\begin{align*}
				P = \{ &r_1\colon \mathit{in}(X) \vee \mathit{out}(X) \leftarrow \mathit{node}(X) \\
						&r_2\colon  \leftarrow \mathit{in}(X), \mathit{in}(Y), \mathit{edge}(X,Y) \\
						&r_3\colon \mathit{size}(S) \leftarrow \ext{\mathit{count}}{\mathit{in}}{S} \\
						&r_4\colon \reva{\leftarrow} \mathit{size}(S), S \rev{>}{<} \mathit{limit} \}
			\end{align*}
			where facts over $\mathit{node}(\cdot)$ and $\mathit{edge}(\cdot)$ define a graph.
			Then $r_1$ and $r_2$ guess an independent set and $r_3$ computes its size,
			\rev{which}{that} is limited to a certain \rev{maximum}{minimum} size $\mathit{limit}$ in $r_4$.
			The grounder must evaluate $\amp{\mathit{count}}$ under all exponentially many possible extensions of $\mathit{in}$
			in order to instantiate rule $r_3$ for all relevant values of variable $S$.
		\end{example}

		In this example, program splitting allows for avoiding unnecessary evaluations.
		To this end, the program might be split into $P_1 = \{ r_1, r_2 \}$ and $P_2 = \{ r_3, r_4 \}$ as illustrated in Figure~\ref{fig:splitting}.
		Then the state-of-the-art algorithm grounds and solves $P_1$, which computes all independent sets,
		and for each of them $P_2$ is grounded and solved.
		
		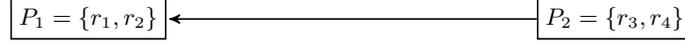
\begin{figure}
			\centering
			\beginpgfgraphicnamed{independentset}
			\begin{tikzpicture}[->,>=stealth',shorten >=1pt,auto,node distance=7cm,semithick]
				\node         (u1) [draw]                   {$P_1=\{r_1, r_2\}$};
				\node         (u2) [draw,right of=u1]       {$P_2=\{r_3, r_4\}$};
				\path (u2) edge [->] node {} (u1);
			\end{tikzpicture}
			\endpgfgraphicnamed
			\caption{Evaluation of $P$ from Example~\ref{ex:graph} based on program splitting}
			\label{fig:splitting}
		\end{figure}
		
		Since the number of independent sets can be exponentially
		smaller than the set of all node selections, the grounding bottleneck can be avoided.
		However, program splitting has the disadvantage that nogoods learned from conflict-driven algorithms~\cite{gks2012-aij}
		cannot be effectively propagated through the whole program, but only within a component.

		The results from Section~\ref{sec:inconsistency} can be used
		to identify a program component as persistently inconsistent wrt.~possible input facts from the predecessor component.
		This information might be used to construct a constraint \rev{which}{that} describes the reason $R$ for this inconsistency in terms of the input facts,
		which can be added as constraint $c_R$ to predecessor
		components in order to eliminate assignments earlier, \rev{which}{that} would make a successor component inconsistent anyway.
		The idea is visualized in Figure~\ref{fig:splittingpruning}.
		
		For details about the computation of inconsistency reasons, exploiting them for the evaluation and experiments we refer to~\citeN{r2017-ijcai}.

		\begin{figure}
			\centering
			\beginpgfgraphicnamed{independentset}
			\begin{tikzpicture}[->,>=stealth',shorten >=1pt,auto,node distance=7cm,semithick]
				\node         (u1) [draw]                   {$P_1=\{r_1, r_2\}$};
				\node         (u2) [draw,right of=u1]       {$P_2=\{r_3, r_4\}$};
				\path (u2) edge [->] node {} (u1);

				\path (u1) edge [->,dashed,bend left] node {add answer set as input atoms $I$} (u2);
				\node         (inc) [below=2cm of u2, align=left]       {detect persistent inconsistency \\ wrt.~$\mathcal{H} = I$ and $\mathcal{B} = \emptyset$};
				\path (u2) edge [->,dashed] node {} (inc);
				\node         (ir) [draw,dashed,left of=inc,align=left]       {inconsistency reason $R$ \\ in terms of input facts $I$};
				\path (inc) edge [->,dashed] node {compute} (ir);
				
				\path (ir) edge [->,dashed] node {add as constraint $c_R$} (u1);
			\end{tikzpicture}
			\endpgfgraphicnamed
			\caption{Exploiting persistent inconsistency for search space pruning}
			\label{fig:splittingpruning}
		\end{figure}
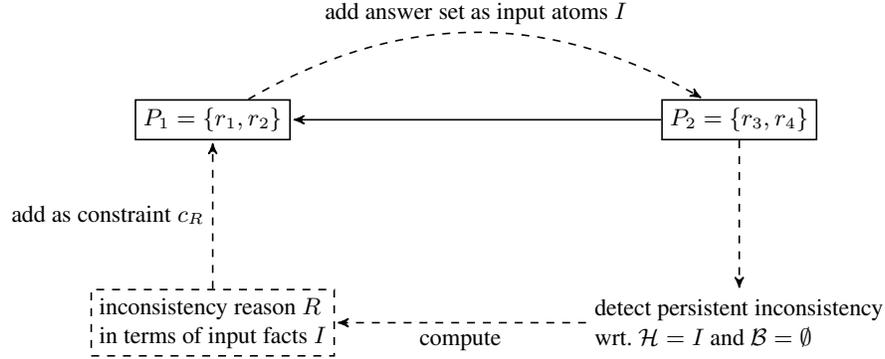

	\section{Discussion and Conclusion}
	\label{sec:conclusion}

		\reva{
		\leanparagraph{Applying the Results to Special Cases of \hex{}.}
		The results presented in this paper carry over to special cases of \hex{}, which, however, often use a specialized syntax.
		\revtwo{Using}{Considering} the example of constraint ASP we briefly sketch how the results can still be applied using another rewriting.

		Constraint ASP allows for using \emph{constraint atoms} in place of ordinary atoms,
		which are of kind $a_1 \circ a_2$, where $a_1$ and $a_2$ are arithmetic expressions
		over (constraint) variables and constants,
		and $\circ$ is a comparison operator. A concrete example is $\mathit{work}(\mathit{lea}) \$+ \mathit{work}(\mathit{john}) \$> 10$, which
		expresses that the sum of the working hours of $\mathit{lea}$ and $\mathit{john}$, represented by constraint variables
		$\mathit{work}(\mathit{lea})$ and $\mathit{work}(\mathit{john})$, is greater than $10$.
		
		Consider the program
		\begin{align*}
		P = \{ &\mathit{project1} \vee \mathit{project2} \leftarrow \\
			&\mathit{work}(\mathit{lea}) \$+ \mathit{work}(\mathit{john}) \$> 10 \leftarrow \mathit{project1} \\
			&\mathit{work}(\mathit{lea}) \$+ \mathit{work}(\mathit{john}) \$> 15 \leftarrow \mathit{project2} \\
			&\leftarrow \mathit{work}(\mathit{lea}) \$> 6\\
			&\leftarrow \mathit{work}(\mathit{john}) \$> 6 \}\text{,}
		\end{align*}
		which represents that either $\mathit{project1}$ or $\mathit{project2}$ is to be realized.
		If $\mathit{project1}$ is chosen, then $\mathit{lea}$ and $\mathit{john}$ together have to spend more than $10$ hours working on the project,
		for $\mathit{project2}$ they have to work more than $15$ hours. However, neither of them wants to spend more than $6$ hours on the project.
		
		Here, the ASP solver assigns truth values to the ordinary and to the constraint atoms, while a constraint solver at the backend
		ensures that these truth values are consistent with the semantics of the constraint theory,
		i.e., that there is an assignment of integers to all constraint variables that witness the truth values of the constraint atoms \revtwo{assignment}{assigned} by the ASP solver.
		For instance,
		the ASP solver may assign $\mathit{project1}$ and $\mathit{work}(\mathit{lea}) \$+ \mathit{work}(\mathit{john}) \$> 10$ to true, and
		$\mathit{work}(\mathit{lea}) \$+ \mathit{work}(\mathit{john}) \$> 15$,
		$\mathit{work}(\mathit{lea}) \$> 6$ and $\mathit{work}(\mathit{john}) \$> 6$ to false in order to satisfy all rules of the program.
		This assignment is consistent with the constraint solver since assigning both $\mathit{work}(\mathit{lea})$
		and $\mathit{work}(\mathit{john})$ to $6$ is consistent with the truth values of the constraint atoms.
		In contrast,
		if the ASP solver assigns $\mathit{project2}$ and $\mathit{work}(\mathit{lea}) \$+ \mathit{work}(\mathit{john}) \$> 15$ to true and
		both $\mathit{work}(\mathit{lea}) \$> 6$ and $\mathit{work}(\mathit{john}) \$> 6$ to false,
		then one cannot assign integers to $\mathit{work}(\mathit{lea})$
		and $\mathit{work}(\mathit{john})$ that are each smaller or equal to $6$ but whose sum is greater than $15$.
		Thus, as expected, the only solution is to realize $\mathit{project1}$.
		
		Although the syntax is tailored and different from \hex, constraint ASP is in fact a special case
		and can be rewritten to a standard \hex-program. To this end, one may introduce a guessing rule of kind
		$\mathit{ctrue}(``\mathit{work}(\mathit{lea}) \$> 6") \vee \mathit{cfalse}(``\mathit{work}(\mathit{lea}) \$> 6") \leftarrow$ for each constraint atom
		and feed the guesses as input to a special external atom of kind $\ext{\mathit{constraintSolverOk}}{\mathit{ctrue}, \mathit{cfalse}}{}$, which interfaces the constraint solver.
		\revtwo{Under the assumption that}{We assume that}
		$\ext{\mathit{constraintSolverOk}}{\mathit{ctrue}, \mathit{cfalse}}{}$ evaluates to true iff the guess is consistent with the constraint solver
		and to false otherwise\revtwo{,}{. Then} an ASP-constraint of form $\leftarrow \naf \ext{\mathit{constraintSolverOk}}{\mathit{ctrue}, \mathit{cfalse}}{}$ in the \hex-program can check the guesses.
		For details of this rewriting we refer to~\citeN{derr2015-aspocp}.
		
		One way to apply the results in this paper to special cases of \hex{} is therefore to first translate dedicated syntax to standard \hex-syntax
		using a rewriting whose correctness was shown.
		Conversely, using such a rewriting as a starting point, one may also translate the results of this paper to the language of special cases of \hex.
		
		\revtwo{}{Applying the results of this paper to special cases of \hex{} allows for making use of the inlining technique also when evaluating programs
			or when checking equivalence of programs that belong to such special cases.
			For instance, one can use the inlining technique for evaluating programs with constraint theories or check equivalence of DL-programs.}
		}
		
		\leanparagraph{Related Work.}
		Our external source inlining approach is related to \reva{inlining-based} evaluation approaches for DL-programs~\cite{eilst2008-aij}, i.e.,~programs with ontologies,
		cf.~\citeN{DBLP:conf/ecai/HeymansEX10}, \citeN{DBLP:conf/rr/XiaoE11} and~\citeN{DBLP:conf/dlog/BajraktariOS17}, but it is more general.
		The former approaches are specific for embedding (certain types of) description logic ontologies.
		In contrast, ours is generic and can handle arbitrary external sources as long as they are decidable and have finite output for each input (cf.~Section~\ref{sec:preliminaries}).
		\revtwo{}{Note that DL-programs can be seen as \hex-programs with a tailored syntax, cf.~\citeN{eiks2008} for formal rewritings of DL-programs to \hex{}.
		When abstracting from these syntactic differences, one can say that our rewriting is correct for a larger class of input programs compared to existing rewritings.}
		
		Our rewriting uses the saturation technique\revtwo{ and is in this respect also related to}{, similar to}
		the one by~\rev{\citeN{afg2015}}{\citeN{afg2015}} (cf.~also \citeN{DBLP:journals/fuin/Alviano16}), who translated nonmonotonic (cyclic) aggregates to disjunctions.
		However, an important difference to our approach \rev{are}{is} that they support only a fixed set of traditional aggregates (such as minimum, maximum, etc)
		whose semantics is directly exploited in a hard-coded fashion in their rewriting, while our approach is generic and thus more flexible.
		\reva{\revtwo{In this respect, our}{Our} approach can be seen as a generalization of previous approaches for specialized formalisms to an integration of ASP with arbitrary sources.}
		\revtwo{Moreover, their rewriting still uses}{Another important difference is that existing rewritings still use} simplified (monotonic) aggregates in the resulting rewritten program
		while we go a step further and eliminate external atoms altogether.
		\revtwo{}{Hence, our rewriting not only supports a larger class of input programs, but also rewrites this larger class to a program from a narrower class.
			}
		This allows the resulting program to be directly forwarded to an ordinary ASP solver,
		while support for aggregates of any kind or additional compatibility checks of guesses are not required.
		
		Based on this inlining approach, we further provided a characterization of equivalence of \hex-programs.
		The criteria generalize previous results for ordinary ASP by \citeN{DBLP:journals/tplp/Woltran08}.%
		\revam{This is a convenient result, but it is not immediate due to possibly nonmonotonic external atoms and the use of the FLP- instead of the GL-reduct.}
		Strong~\cite{Lifschitz:2001:SEL:383779.383783} and uniform equivalence~\cite{DBLP:conf/iclp/EiterF03}
		are \rev{well-knwon}{well-known} and important special cases thereof and carry over as well.
		
		\citeN{DBLP:conf/jelia/Woltran04} also discussed the special cases of
		\emph{head-relativized} equivalence ($\mathcal{H} = \mathit{HB}_{\mathcal{C}}(P)$ while $\mathcal{B}$ can be freely chosen),
		and
		\emph{body-relativized} equivalence ($\mathcal{B} = \mathit{HB}_{\mathcal{C}}(P)$ while $\mathcal{H}$ can be freely chosen).
		Also the cases where $\mathcal{B} \subseteq \mathcal{H}$ and $\mathcal{H} \subseteq \mathcal{B}$ were analyzed.
		Corollaries have been derived \rev{which}{that} simplify the conditions to check for these special cases.
		They all follow directly from an analogous version of Proposition~\ref{prop:semodelshex} for plain ASP by substituting $\mathcal{H}$ or $\mathcal{B}$ by a fixed value.
		Since we established by Proposition~\ref{prop:semodelshex} that the requirements hold also for \hex-programs,
		their corollaries, as summarized in Section~5 by~\citeN{DBLP:journals/tplp/Woltran08}, hold analogously.

		The work is also related to the one by \citeN{TRUSZCZYNSKI20101285},
		who extended strong equivalence to propositional theories under FLP-semantics.
		However, the relationship concerns only the use of the FLP-semantics, while the notion of equivalence and the formalism for which the \rev{equivalencei}{equivalence} is shown are different.
		In particular, $\langle \mathcal{H}, \mathcal{B} \rangle$-equivalence and external sources were not considered.
		
		\leanparagraph{Conclusion and Outlook.}
		We presented an approach for external source inlining based on support sets.
		Due to nonmonotonicity of external atoms, the encoding is not trivial and requires a saturation encoding.
		We note that the results are interesting beyond \hex-programs
		since well-known ASP extensions, such as programs with aggregates~\cite{flp2011-ai}
		or with specific external atoms such as constraint atoms~\cite{geossc09a},
		\rev{amount to}{are} special cases of \hex, and thus the results are applicable in such cases.
		
		One application of the technique can be found in an alternative evaluation approach,
		which is intended to be used for external sources \rev{which}{that} have a compact representation as support sets.
		Previous approaches had to guess the truth values of external atoms and verify the guesses
		either by explicit evaluation (as in the traditional approach) or by matching guesses against support sets (as in the approach by~\citeN{eiterFR014}).
		Instead, the new inlining-based approach compiles external atoms away altogether such that
		the program can be entirely evaluated by an ordinary ASP solver.
		For the considered class of external sources,
		our experiments show a clear and significant improvement over the previous support-set-based approach by~\citeN{eiterFR014},
		which is explained by the fact that the slightly higher initialization costs are exceeded by
		the significant benefits of avoiding external calls altogether,
		and for the considered types of external sources also over the traditional approach.
		
		Another application is found in the extension of previous \rev{notions}{characterizations} of program equivalence
		from ordinary ASP\reva{-} to \hex-programs. We generalizes such \rev{notions}{characterizations} from ordinary ASP to \hex-programs.
		Since this is a theoretical result, compact representation of external sources is not an issue here.
		From the criterion for program equivalence we derive further criteria for program inconsistency wrt.~program extensions,
		which have applications in context of evaluation algorithms for \hex-programs.
		
		Potential future work includes refinements of the rewriting.
		Currently, a new auxiliary variable $\overline{a}$ is introduced for all input atoms $a$ of all external atoms.
		Thus, a quadratic number of auxiliary atoms is required.
		While the reuse of the auxiliary variables is not always possible, the identification of cases were auxiliary variables
		can be shared among multiple inlined external atoms is interesting.
		For the equivalence \rev{criteria}{criterion}, future work may also include the extension of the results to non-ground programs, cf.~\citeN{DBLP:conf/aaai/EiterFTW05}.
		
		Moreover, currently we do not distinguish between body atoms and input atoms to external atoms when we define which programs are allows to be added.
		A more fine-grained approach \rev{which}{that} supports this distinction may allow for identifying programs as equivalent \rev{which}{that}
		are not equivalent wrt.~to the current notion.
		Also allowing only external atoms with specific properties, such as monotonicity, may lead to more fine-grained criteria.

		Furthermore, a recent alternative notion of equivalence is \emph{rule equivalence}~\cite{DBLP:conf/foiks/BliemW16}.
		Here, not the set of atoms \rev{which}{that} can occur in the added program is constrained, but the type of the rules.
		In particular, proper rules may be added, while the addition of facts is limited to certain atoms;
		generalizing this notion to \hex-programs is a possible starting point for future work.

	\ifinlineref

	\else
		\iftlp
			\bibliographystyle{acmtrans}
		\else
			\bibliographystyle{elsarticle-harv}
		\fi
		\bibliography{inlining}

\begin{thebibliography}{}

\bibitem[\protect\citeauthoryear{Alviano}{Alviano}{2016}]{DBLP:journals/fuin/Alviano16}
{\sc Alviano, M.} 2016.
\newblock Evaluating answer set programming with non-convex recursive
  aggregates.
\newblock {\em Fundamentae Informaticae\/}~{\em 149,\/}~1-2, 1--34.

\bibitem[\protect\citeauthoryear{Alviano, Faber, and Gebser}{Alviano
  et~al\mbox{.}}{2015}]{afg2015}
{\sc Alviano, M.}, {\sc Faber, W.}, {\sc and} {\sc Gebser, M.} 2015.
\newblock Rewriting recursive aggregates in answer set programming: back to
  monotonicity.
\newblock {\em {TPLP}\/}~{\em 15,\/}~4-5, 559--573.

\bibitem[\protect\citeauthoryear{Bajraktari, Ortiz, and Simkus}{Bajraktari
  et~al\mbox{.}}{2017}]{DBLP:conf/dlog/BajraktariOS17}
{\sc Bajraktari, L.}, {\sc Ortiz, M.}, {\sc and} {\sc Simkus, M.} 2017.
\newblock Clopen knowledge bases: Combining description logics and answer set
  programming.
\newblock In {\em Proceedings of the 30th International Workshop on Description
  Logics, Montpellier, France, July 18-21, 2017.}, {A.~Artale}, {B.~Glimm},
  {and} {R.~Kontchakov}, Eds. {CEUR} Workshop Proceedings, vol. 1879.
  CEUR-WS.org.

\bibitem[\protect\citeauthoryear{Baumann, Dvor{\'{a}}k, Linsbichler, and
  Woltran}{Baumann et~al\mbox{.}}{2017}]{DBLP:conf/ijcai/BaumannDLW17}
{\sc Baumann, R.}, {\sc Dvor{\'{a}}k, W.}, {\sc Linsbichler, T.}, {\sc and}
  {\sc Woltran, S.} 2017.
\newblock A general notion of equivalence for abstract argumentation.
\newblock In {\em Proceedings of the Twenty-Sixth International Joint
  Conference on Artificial Intelligence, {IJCAI} 2017, Melbourne, Australia,
  August 19-25, 2017}, {C.~Sierra}, Ed. ijcai.org, 800--806.

\bibitem[\protect\citeauthoryear{Bliem and Woltran}{Bliem and
  Woltran}{2016}]{DBLP:conf/foiks/BliemW16}
{\sc Bliem, B.} {\sc and} {\sc Woltran, S.} 2016.
\newblock Equivalence between answer-set programs under (partially) fixed
  input.
\newblock In {\em FoIKS}. Lecture Notes in Computer Science, vol. 9616.
  Springer, 95--111.

\bibitem[\protect\citeauthoryear{Calvanese, Lembo, Lenzerini, and
  Rosati}{Calvanese et~al\mbox{.}}{2007}]{clmr2007}
{\sc Calvanese, D.}, {\sc Lembo, D.}, {\sc Lenzerini, M.}, {\sc and} {\sc
  Rosati, R.} 2007.
\newblock Tractable reasoning and efficient query answering in description
  logics: The {DL}-{L}ite family.
\newblock {\em Journal of Automated Reasoning\/}~{\em 39,\/}~3 (October),
  385--429.

\bibitem[\protect\citeauthoryear{Darwiche and Marquis}{Darwiche and
  Marquis}{2002}]{DBLP:journals/corr/abs-1106-1819}
{\sc Darwiche, A.} {\sc and} {\sc Marquis, P.} 2002.
\newblock A knowledge compilation map.
\newblock {\em J. Artif. Intell. Res. {(JAIR)}\/}~{\em 17}, 229--264.

\bibitem[\protect\citeauthoryear{{De Rosis}, Eiter, Redl, and Ricca}{{De Rosis}
  et~al\mbox{.}}{2015}]{derr2015-aspocp}
{\sc {De Rosis}, A.}, {\sc Eiter, T.}, {\sc Redl, C.}, {\sc and} {\sc Ricca,
  F.} 2015.
\newblock Constraint answer set programming based on {HEX}-programs.
\newblock In {\em Eighth Workshop on Answer Set Programming and Other Computing
  Paradigms ({ASPOCP} 2015), August 31, 2015, Cork, Ireland} (August 31, 2015).

\bibitem[\protect\citeauthoryear{Drescher and Walsh}{Drescher and
  Walsh}{2012}]{DBLP:conf/iclp/DrescherW12}
{\sc Drescher, C.} {\sc and} {\sc Walsh, T.} 2012.
\newblock Answer set solving with lazy nogood generation.
\newblock In {\em Technical Communications of the 28th International Conference
  on Logic Programming, {ICLP} 2012, September 4-8, 2012, Budapest, Hungary},
  {A.~Dovier} {and} {V.~S. Costa}, Eds. LIPIcs, vol.~17. Schloss Dagstuhl -
  Leibniz-Zentrum fuer Informatik, 188--200.

\bibitem[\protect\citeauthoryear{Eiter and Fink}{Eiter and
  Fink}{2003}]{DBLP:conf/iclp/EiterF03}
{\sc Eiter, T.} {\sc and} {\sc Fink, M.} 2003.
\newblock Uniform equivalence of logic programs under the stable model
  semantics.
\newblock In {\em Logic Programming, 19th International Conference, {ICLP}
  2003, Mumbai, India, December 9-13, 2003, Proceedings}, {C.~Palamidessi}, Ed.
  Lecture Notes in Computer Science, vol. 2916. Springer, 224--238.

\bibitem[\protect\citeauthoryear{Eiter, Fink, Ianni, Krennwallner, Redl, and
  Sch{\"{u}}ller}{Eiter et~al\mbox{.}}{2016}]{efikrs2015-tplp}
{\sc Eiter, T.}, {\sc Fink, M.}, {\sc Ianni, G.}, {\sc Krennwallner, T.}, {\sc
  Redl, C.}, {\sc and} {\sc Sch{\"{u}}ller, P.} 2016.
\newblock A model building framework for answer set programming with external
  computations.
\newblock {\em {TPLP}\/}~{\em 16,\/}~4, 418--464.

\bibitem[\protect\citeauthoryear{Eiter, Fink, Krennwallner, and Redl}{Eiter
  et~al\mbox{.}}{2012}]{efkr2012-tplp}
{\sc Eiter, T.}, {\sc Fink, M.}, {\sc Krennwallner, T.}, {\sc and} {\sc Redl,
  C.} 2012.
\newblock Conflict-driven {ASP} solving with external sources.
\newblock {\em TPLP\/}~{\em 12,\/}~4-5, 659--679.

\bibitem[\protect\citeauthoryear{Eiter, Fink, Krennwallner, and Redl}{Eiter
  et~al\mbox{.}}{2016}]{efkr2016-aij}
{\sc Eiter, T.}, {\sc Fink, M.}, {\sc Krennwallner, T.}, {\sc and} {\sc Redl,
  C.} 2016.
\newblock Domain expansion for {ASP}-programs with external sources.
\newblock {\em Artificial Intelligence\/}~{\em 233}, 84--121.

\bibitem[\protect\citeauthoryear{Eiter, Fink, Krennwallner, Redl, and
  Sch\"{u}ller}{Eiter et~al\mbox{.}}{2014}]{efkrs2014-jair}
{\sc Eiter, T.}, {\sc Fink, M.}, {\sc Krennwallner, T.}, {\sc Redl, C.}, {\sc
  and} {\sc Sch\"{u}ller, P.} 2014.
\newblock Efficient {HEX}-program evaluation based on unfounded sets.
\newblock {\em Journal of Artificial Intelligence Research\/}~{\em 49},
  269--321.

\bibitem[\protect\citeauthoryear{Eiter, Fink, Redl, and Stepanova}{Eiter
  et~al\mbox{.}}{2014}]{eiterFR014}
{\sc Eiter, T.}, {\sc Fink, M.}, {\sc Redl, C.}, {\sc and} {\sc Stepanova, D.}
  2014.
\newblock Exploiting support sets for answer set programs with external
  evaluations.
\newblock In {\em Proceedings of the Twenty-Eighth {AAAI} Conference on
  Artificial Intelligence, July 27 -31, 2014, Qu{\'{e}}bec City, Qu{\'{e}}bec,
  Canada.}, {C.~E. Brodley} {and} {P.~Stone}, Eds. {AAAI} Press, 1041--1048.

\bibitem[\protect\citeauthoryear{Eiter, Fink, and Stepanova}{Eiter
  et~al\mbox{.}}{2014}]{Eiter:2014:TPD:3006652.3006701}
{\sc Eiter, T.}, {\sc Fink, M.}, {\sc and} {\sc Stepanova, D.} 2014.
\newblock Towards practical deletion repair of inconsistent dl-programs.
\newblock In {\em Proceedings of the Twenty-first European Conference on
  Artificial Intelligence}. ECAI'14. IOS Press, Amsterdam, The Netherlands, The
  Netherlands, 285--290.

\bibitem[\protect\citeauthoryear{Eiter, Fink, Tompits, and Woltran}{Eiter
  et~al\mbox{.}}{2005}]{DBLP:conf/aaai/EiterFTW05}
{\sc Eiter, T.}, {\sc Fink, M.}, {\sc Tompits, H.}, {\sc and} {\sc Woltran, S.}
  2005.
\newblock Strong and uniform equivalence in answer-set programming:
  Characterizations and complexity results for the non-ground case.
\newblock In {\em Proceedings, The Twentieth National Conference on Artificial
  Intelligence and the Seventeenth Innovative Applications of Artificial
  Intelligence Conference, July 9-13, 2005, Pittsburgh, Pennsylvania, {USA}},
  {M.~M. Veloso} {and} {S.~Kambhampati}, Eds. {AAAI} Press / The {MIT} Press,
  695--700.

\bibitem[\protect\citeauthoryear{Eiter, Ianni, and Krennwallner}{Eiter
  et~al\mbox{.}}{2009}]{eik2009-rw}
{\sc Eiter, T.}, {\sc Ianni, G.}, {\sc and} {\sc Krennwallner, T.} 2009.
\newblock {Answer Set Programming: A Primer}.
\newblock In {\em 5th International Reasoning Web Summer School (RW 2009),
  Brixen/Bressanone, Italy, August 30--September 4, 2009}, {S.~Tessaris},
  {E.~Franconi}, {T.~Eiter}, {C.~Gutierrez}, {S.~Handschuh}, {M.-C. Rousset},
  {and} {R.~A. Schmidt}, Eds. LNCS, vol. 5689. Springer, 40--110.

\bibitem[\protect\citeauthoryear{Eiter, Ianni, Krennwallner, and
  Schindlauer}{Eiter et~al\mbox{.}}{2008}]{eiks2008}
{\sc Eiter, T.}, {\sc Ianni, G.}, {\sc Krennwallner, T.}, {\sc and} {\sc
  Schindlauer, R.} 2008.
\newblock {Exploiting Conjunctive Queries in Description Logic Programs}.
\newblock Tech. Rep. INFSYS RR-1843-08-02, Institut f{\"u}r
  Informationssysteme, TU Wien, Favoritenstra{\ss}e 9-11, A-1040 Vienna. Mar.

\bibitem[\protect\citeauthoryear{Eiter, Ianni, Lukasiewicz, Schindlauer, and
  Tompits}{Eiter et~al\mbox{.}}{2008}]{eilst2008-aij}
{\sc Eiter, T.}, {\sc Ianni, G.}, {\sc Lukasiewicz, T.}, {\sc Schindlauer, R.},
  {\sc and} {\sc Tompits, H.} 2008.
\newblock Combining answer set programming with description logics for the
  semantic web.
\newblock {\em Artif. Intell.\/}~{\em 172,\/}~12-13, 1495--1539.

\bibitem[\protect\citeauthoryear{Faber}{Faber}{2005}]{faber2005-lpnmr}
{\sc Faber, W.} 2005.
\newblock Unfounded sets for disjunctive logic programs with arbitrary
  aggregates.
\newblock In {\em Proceedings of the Eighth International Conference on Logic
  Programming and Nonmonotonic Reasoning (LPNMR 2005), Diamante, Italy,
  September 5-8, 2005}. Vol. 3662. Springer, 40--52.

\bibitem[\protect\citeauthoryear{Faber, Leone, and Pfeifer}{Faber
  et~al\mbox{.}}{2011}]{flp2011-ai}
{\sc Faber, W.}, {\sc Leone, N.}, {\sc and} {\sc Pfeifer, G.} 2011.
\newblock Semantics and complexity of recursive aggregates in answer set
  programming.
\newblock {\em Artificial Intelligence\/}~{\em 175,\/}~1 (January), 278--298.

\bibitem[\protect\citeauthoryear{Franco and Martin}{Franco and
  Martin}{2009}]{FM09HBSAT}
{\sc Franco, J.} {\sc and} {\sc Martin, J.} 2009.
\newblock {\em A History of Satisfiability}. Frontiers in Artificial
  Intelligence and Applications, vol. 185.
\newblock IOS Press, Chapter~1, 3--74.

\bibitem[\protect\citeauthoryear{Gebser, Kaufmann, and Schaub}{Gebser
  et~al\mbox{.}}{2012}]{gks2012-aij}
{\sc Gebser, M.}, {\sc Kaufmann, B.}, {\sc and} {\sc Schaub, T.} 2012.
\newblock Conflict-driven answer set solving: From theory to practice.
\newblock {\em Artificial Intelligence\/}~{\em 187-188}, 52--89.

\bibitem[\protect\citeauthoryear{Gebser, Ostrowski, and Schaub}{Gebser
  et~al\mbox{.}}{2009}]{geossc09a}
{\sc Gebser, M.}, {\sc Ostrowski, M.}, {\sc and} {\sc Schaub, T.} 2009.
\newblock Constraint answer set solving.
\newblock In {\em Proceedings of the Twenty-fifth International Conference on
  Logic Programming (ICLP'09)}, {P.~Hill} {and} {D.~Warren}, Eds. Lecture Notes
  in Computer Science, vol. 5649. Springer-Verlag, 235--249.

\bibitem[\protect\citeauthoryear{Gelfond and Lifschitz}{Gelfond and
  Lifschitz}{1988}]{gelf-lifs-88}
{\sc Gelfond, M.} {\sc and} {\sc Lifschitz, V.} 1988.
\newblock {The Stable Model Semantics for Logic Programming}.
\newblock In {\em {Logic Programming: Proceedings of the 5th International
  Conference and Symposium}}, {R.~Kowalski} {and} {K.~Bowen}, Eds. {MIT Press},
  1070--1080.

\bibitem[\protect\citeauthoryear{Gelfond and Lifschitz}{Gelfond and
  Lifschitz}{1991}]{gelf-lifs-91}
{\sc Gelfond, M.} {\sc and} {\sc Lifschitz, V.} 1991.
\newblock {Classical Negation in Logic Programs and Disjunctive Databases}.
\newblock {\em {New Generation Computing}\/}~{\em 9,\/}~3--4, 365--386.

\bibitem[\protect\citeauthoryear{Heymans, Eiter, and Xiao}{Heymans
  et~al\mbox{.}}{2010}]{DBLP:conf/ecai/HeymansEX10}
{\sc Heymans, S.}, {\sc Eiter, T.}, {\sc and} {\sc Xiao, G.} 2010.
\newblock Tractable reasoning with dl-programs over datalog-rewritable
  description logics.
\newblock In {\em {ECAI} 2010 - 19th European Conference on Artificial
  Intelligence, Lisbon, Portugal, August 16-20, 2010, Proceedings},
  {H.~Coelho}, {R.~Studer}, {and} {M.~Wooldridge}, Eds. Frontiers in Artificial
  Intelligence and Applications, vol. 215. {IOS} Press, 35--40.

\bibitem[\protect\citeauthoryear{Lembo, Lenzerini, Rosati, Ruzzi, and
  Savo}{Lembo et~al\mbox{.}}{2011}]{Lembo2011}
{\sc Lembo, D.}, {\sc Lenzerini, M.}, {\sc Rosati, R.}, {\sc Ruzzi, M.}, {\sc
  and} {\sc Savo, D.~F.} 2011.
\newblock {\em Query Rewriting for Inconsistent DL-Lite Ontologies}.
\newblock Springer Berlin Heidelberg, Berlin, Heidelberg, 155--169.

\bibitem[\protect\citeauthoryear{Lifschitz, Pearce, and Valverde}{Lifschitz
  et~al\mbox{.}}{2001}]{Lifschitz:2001:SEL:383779.383783}
{\sc Lifschitz, V.}, {\sc Pearce, D.}, {\sc and} {\sc Valverde, A.} 2001.
\newblock Strongly equivalent logic programs.
\newblock {\em ACM Trans. Comput. Logic\/}~{\em 2,\/}~4 (Oct.), 526--541.

\bibitem[\protect\citeauthoryear{Mayer, Stumptner, Bettex, and Falkner}{Mayer
  et~al\mbox{.}}{2009}]{mbsf2009}
{\sc Mayer, W.}, {\sc Stumptner, M.}, {\sc Bettex, M.}, {\sc and} {\sc Falkner,
  A.} 2009.
\newblock On solving complex rack configuration problems using csp methods.
\newblock In {\em Proceedings of the Workshop on Configuration at the 21st
  International Conference on Artificial Intelligence}, {M.~Stumptner} {and}
  {P.~Albert}, Eds. Pasadena, CA, USA, 53--60.

\bibitem[\protect\citeauthoryear{Nieuwenhuis and Oliveras}{Nieuwenhuis and
  Oliveras}{2005}]{Nieuwenhuis05theorypropagation}
{\sc Nieuwenhuis, R.} {\sc and} {\sc Oliveras, A.} 2005.
\newblock {DPLL(T)} with exhaustive theory propagation and its application to
  difference logic.
\newblock In {\em In CAV'05 LNCS 3576}. Springer, 321--334.

\bibitem[\protect\citeauthoryear{Ohrimenko, Stuckey, and Codish}{Ohrimenko
  et~al\mbox{.}}{2009}]{Ohrimenko:2009:PVL:1553323.1553342}
{\sc Ohrimenko, O.}, {\sc Stuckey, P.~J.}, {\sc and} {\sc Codish, M.} 2009.
\newblock Propagation via lazy clause generation.
\newblock {\em Constraints\/}~{\em 14,\/}~3 (Sept.), 357--391.

\bibitem[\protect\citeauthoryear{Ostrowski and Schaub}{Ostrowski and
  Schaub}{2012}]{os2012-tplp}
{\sc Ostrowski, M.} {\sc and} {\sc Schaub, T.} 2012.
\newblock {ASP} modulo {CSP:} the clingcon system.
\newblock {\em {TPLP}\/}~{\em 12,\/}~4-5, 485--503.

\bibitem[\protect\citeauthoryear{Redl}{Redl}{2017a}]{r2017-ijcai}
{\sc Redl, C.} 2017a.
\newblock Conflict-driven {ASP} solving with external sources and program
  splits.
\newblock In {\em Proceedings of the Twenty-Sixth International Joint
  Conference on Artificial Intelligence (IJCAI 2017), August 19--25, 2017,
  Melbourne, Australia} (August 19--25, 2017). AAAI Press, 1239--1246.

\bibitem[\protect\citeauthoryear{Redl}{Redl}{2017b}]{r2017a-aaai}
{\sc Redl, C.} 2017b.
\newblock Efficient evaluation of answer set programs with external sources
  based on external source inlining.
\newblock In {\em Proceedings of the Thirty-First AAAI Conference (AAAI 2017),
  February 4--9, 2016, San Francisco, California, USA} (February 4--9, 2016).
  AAAI Press.

\bibitem[\protect\citeauthoryear{Redl}{Redl}{2017c}]{r2017b-aaai}
{\sc Redl, C.} 2017c.
\newblock On equivalence and inconsistency of answer set programs with external
  sources.
\newblock In {\em Proceedings of the Thirty-First AAAI Conference (AAAI 2017),
  February 4--9, 2016, San Francisco, California, USA} (February 4--9, 2016).
  AAAI Press.

\bibitem[\protect\citeauthoryear{Truszczy{\'n}ski}{Truszczy{\'n}ski}{2010}]{TRUSZCZYNSKI20101285}
{\sc Truszczy{\'n}ski, M.} 2010.
\newblock Reducts of propositional theories, satisfiability relations, and
  generalizations of semantics of logic programs.
\newblock {\em Artificial Intelligence\/}~{\em 174,\/}~16, 1285 -- 1306.

\bibitem[\protect\citeauthoryear{Woltran}{Woltran}{2004}]{DBLP:conf/jelia/Woltran04}
{\sc Woltran, S.} 2004.
\newblock Characterizations for relativized notions of equivalence in answer
  set programming.
\newblock In {\em Logics in Artificial Intelligence, 9th European Conference,
  {JELIA} 2004, Lisbon, Portugal, September 27-30, 2004, Proceedings}, {J.~J.
  Alferes} {and} {J.~A. Leite}, Eds. Lecture Notes in Computer Science, vol.
  3229. Springer, 161--173.

\bibitem[\protect\citeauthoryear{Woltran}{Woltran}{2008}]{DBLP:journals/tplp/Woltran08}
{\sc Woltran, S.} 2008.
\newblock A common view on strong, uniform, and other notions of equivalence in
  answer-set programming.
\newblock {\em {TPLP}\/}~{\em 8,\/}~2, 217--234.

\bibitem[\protect\citeauthoryear{Xiao and Eiter}{Xiao and
  Eiter}{2011}]{DBLP:conf/rr/XiaoE11}
{\sc Xiao, G.} {\sc and} {\sc Eiter, T.} 2011.
\newblock Inline evaluation of hybrid knowledge bases - {PhD} description.
\newblock In {\em Web Reasoning and Rule Systems - 5th International
  Conference, {RR} 2011, Galway, Ireland, August 29-30, 2011. Proceedings},
  {S.~Rudolph} {and} {C.~Gutierrez}, Eds. Lecture Notes in Computer Science,
  vol. 6902. Springer, 300--305.

\end{thebibliography}
	\fi

	\clearpage
	\newpage
\begin{appendix}

\section{Proofs}
\label{sec:proofs}

\proofs{}

\end{appendix}


\end{document}